%% file: template.tex
\documentclass{article}

\usepackage{arxiv}
\usepackage{tikz}
\usepackage[utf8]{inputenc} 
\usepackage[T1]{fontenc}    
\usepackage{hyperref}       
\usepackage{url}            
\usepackage{booktabs}       
\usepackage{amsfonts}       
\usepackage{nicefrac}       
\usepackage{microtype}      
\usepackage{lipsum}
\usepackage{cite}

\usepackage{amssymb}
\usepackage{lineno,hyperref}
\usepackage{amsmath}
\usepackage{bm}
\usepackage{subcaption}
\usepackage{graphicx}
\usepackage{svg}
\usepackage{rotating}
\usepackage{lscape}
\usepackage{array,multirow}
\usepackage[title]{appendix}
\usepackage{float}
\usepackage{stmaryrd}
\usepackage{textcomp}
\usepackage{algorithm}
\usepackage{algpseudocode}
\usepackage{amsthm}
\usepackage{listings}
\usepackage{xcolor}
\usepackage{siunitx}
\usepackage{symbols}
\usepackage{pifont}
\usepackage{lineno}

\usepackage[resetlabels]{multibib}
\newcites{M}{References}
\newcites{SM}{References}

\newcommand{\beginsupplement}{%
     \setcounter{section}{0}
        \renewcommand{\thesection}{S\arabic{section}}%
     }
 \newcommand{\beginsupplementsub}{%
        \setcounter{equation}{0}
        \renewcommand{\theequation}{\thesection.\arabic{equation}}%
        \setcounter{table}{0}
        \renewcommand{\thetable}{\thesection.\arabic{table}}%
        \setcounter{figure}{0}
        \renewcommand{\thefigure}{\thesection.\arabic{figure}}%
     }    

\usepackage{import}
\subimport{layers/}{init}

\usetikzlibrary{positioning}
\usetikzlibrary{3d} 

\def\ConvColor{rgb:yellow,5;red,2.5;white,5}
\def\ConvReluColor{rgb:yellow,5;red,5;white,5}
\def\PoolColor{rgb:red,1;black,0.3}
\def\UnpoolColor{rgb:blue,2;green,1;black,0.3}

\def\SoftmaxColor{rgb:magenta,5;black,7}   

\def\checkmark{\tikz\fill[scale=0.4](0,.35) -- (.25,0) -- (1,.7) -- (.25,.15) -- cycle;}

\DeclareMathOperator{\tr}{tr}
\newtheorem{remark}{Remark}

\newcommand{\copymidarrow}{\tikz \draw[-Stealth,line width=0.8mm,draw={rgb:blue,4;red,1;green,1;black,3}] (-0.3,0) -- ++(0.3,0);}

\title{A framework for data-driven solution and parameter estimation of PDEs using conditional generative adversarial networks}

\author{
  Teeratorn Kadeethum \\
  Sibley School of Mechanical and Aerospace Engineering \\
  Cornell University, 
   New York, USA \\
  \texttt{tk658@cornell.edu} \\
  
  \And
 Daniel O'Malley \\
  Los Alamos National Laboratory \\
   New Mexico, USA \\
  \texttt{omalled@lanl.gov} \\
  
  \And
    Jan Niklas Fuhg \\
  Sibley School of Mechanical and Aerospace Engineering \\
  Cornell University, 
   New York, USA \\
  \texttt{jf853@cornell.edu} \\
  
   \And
 Youngsoo Choi \\
  Lawrence Livermore National Laboratory \\
   California, USA \\
  \texttt{choi15@llnl.gov} \\
  
  \And
  Jonghyun Lee \\
  Civil \& Environmental Engineering/Water Resources Research Center\\
University of Hawai'i at Manoa \\
   Hawai, USA \\
  \texttt{jonghyun.harry.lee@hawaii.edu} \\
  
  \And
 Hari S. Viswanathan \\
  Los Alamos National Laboratory \\
   New Mexico, USA \\
  \texttt{viswana@lanl.gov} \\
  
   \And
 Nikolaos Bouklas \\
  Sibley School of Mechanical and Aerospace Engineering and\\
  Center for Applied Mathematics \\
  Cornell University,
   New York, USA \\
  \texttt{nb589@cornell.edu} \\
}

\renewcommand{\shorttitle}{Data-driven solution and parameter estimation of PDEs using cGAN}

\hypersetup{
pdftitle={A template for the arxiv style},
pdfsubject={q-bio.NC, q-bio.QM},
pdfauthor={David S.~Hippocampus, Elias D.~Striatum},
pdfkeywords={First keyword, Second keyword, More},
}

\begin{document}
\maketitle

\begin{abstract}
	This work is the first to employ and adapt the image-to-image translation concept based on conditional generative adversarial networks (cGAN) towards learning a forward and an inverse solution operator of partial differential equations (PDEs). Even though the proposed framework could be applied as a surrogate model for the solution of any PDEs, here we focus on steady-state solutions of coupled hydro-mechanical processes in heterogeneous porous media. Strongly heterogeneous material properties, which translate to the heterogeneity of coefficients of the PDEs and discontinuous features in the solutions, require specialized techniques for the forward and inverse solution of these problems. Additionally, parametrization of the spatially heterogeneous coefficients is excessively difficult by using standard reduced order modeling techniques. In this work, we overcome these challenges by employing the image-to-image translation concept to learn the forward and inverse solution operators and utilize a U-Net generator and a patch-based discriminator. We use heterogeneous permeability fields as an input parameter and pressure and displacement fields as outputs for the forward model, enabling calculation at least 2000 times faster than a linear finite element solver. For inverse modeling, the framework allows for estimating output in terms of the permeability field, given input of pressure and/or displacement fields, where input data is incomplete and contaminated by noise. The framework provides a speed-up of 120000 times compared to a Gaussian prior-based inverse modeling approach while also delivering more accurate inverse results. Our results show that the proposed data-driven reduced order model has competitive predictive performance capabilities in accuracy and computational efficiency as well as training time requirements compared to state-of-the-art data-driven methods for both forward and inverse problems.
\end{abstract}


\section{Introduction}




Many engineering applications involving porous media ranging from groundwater and contaminant hydrology to tissue engineering and drug delivery rely on physical simulations
\cite{middleton2015shale,choo2018cracking,yu2020poroelastic,vinje2018fluid,kadeethum2019investigation,bouklas2012swelling}. These simulations are usually governed by partial differential equations (PDEs). For instance, coupled hydro-mechanical (HM) processes in porous media, in which fluid flow and solid deformation tightly interact, could be described by Biot's formulation of linear poroelasticity \cite{biot1941general}. These PDEs could be solved analytically for simple geometries and boundary conditions \cite{liu2009lie,kadeethum2020well} or approximated numerically using different techniques such as finite volume or finite element methods \cite{wheeler2014coupling,deng2017locally,liu2018lowest,devarajan2020thermal}, which we will further term as full order model (FOM). However, the latter requires substantial computational resources, especially when discontinuities arise in the solution  \cite{hansen2010discrete,hesthaven2016certified}. Hence, the FOM is not directly suitable to handle large-scale inverse problems, optimization, or even concurrent multiscale calculations in which an extensive set of simulations are required to be explored \cite{ballarin2019pod,hesthaven2016certified}.  \par

Development of a reduced order model (ROM) \cite{schilders2008model,choi2020gradient, mcbane2021component, choi2019accelerating, amsallem2015design} which aims to produce a low-dimensional representation of FOM could be an alternative towards handling field-scale inverse problems, optimization, or control. ROMs are generally composed of an offline and an online stage \cite{kim2020efficient, kim2020fast, hoang2020domain, carlberg2018conservative}. The offline stage begins with the initialization of uncertain input parameters, referred to as a training set. Then, the FOM is solved successively using each member of the training set as inputs, and we will refer to the corresponding solutions as snapshots from now on. Data compression techniques are then used to compress the FOM snapshots to produce basis functions that span a reduced space of very low dimensionality (compared to the original dimension) but still guarantee accurate reproduction of the snapshots \cite{decaria2020artificial,cleary1984data,choi2021spacetime}. ROMs can generate forward solutions during the online stage for any new value of parameters by seeking an approximated solution in the reduced space. The ROM methodology relies on a parametrized problem (i.e., repeated evaluations of a problem depending on parameters, which could correspond to physical properties, geometric characteristics, or boundary conditions \cite{ballarin2019pod,venturi2019weighted,hesthaven2016certified}). However, it is difficult to parameterize spatial fields of PDE coefficients (e.g., corresponding to heterogeneous material properties) by a small number of parameters. For instance, the problem of interest in HM processes commonly involves complex subsurface structures \cite{matthai2009upscaling,flemisch2018benchmarks,jia2017comprehensive} where the corresponding spatially distributed parameters can span several orders of magnitude and include discontinuous features. As a result, the traditional parametrized ROM might not be suitable for this type of problem because a “global” proper orthogonal decomposition (POD) approach might require a high dimensional reduced basis (see \ref{sec:sup_pod} for more information). Moreover, to handle inverse problems, FOM or ROM are usually used in conjunction with optimization algorithms or adjoint state methods that are not straightforward to implement and still suffer from the curse of dimensionality \cite{o2019learning,lin2016computationally,villa2018hippylib}. \par  

\begin{figure}[!ht]
   \centering
         \includegraphics[keepaspectratio, height=5.5cm]{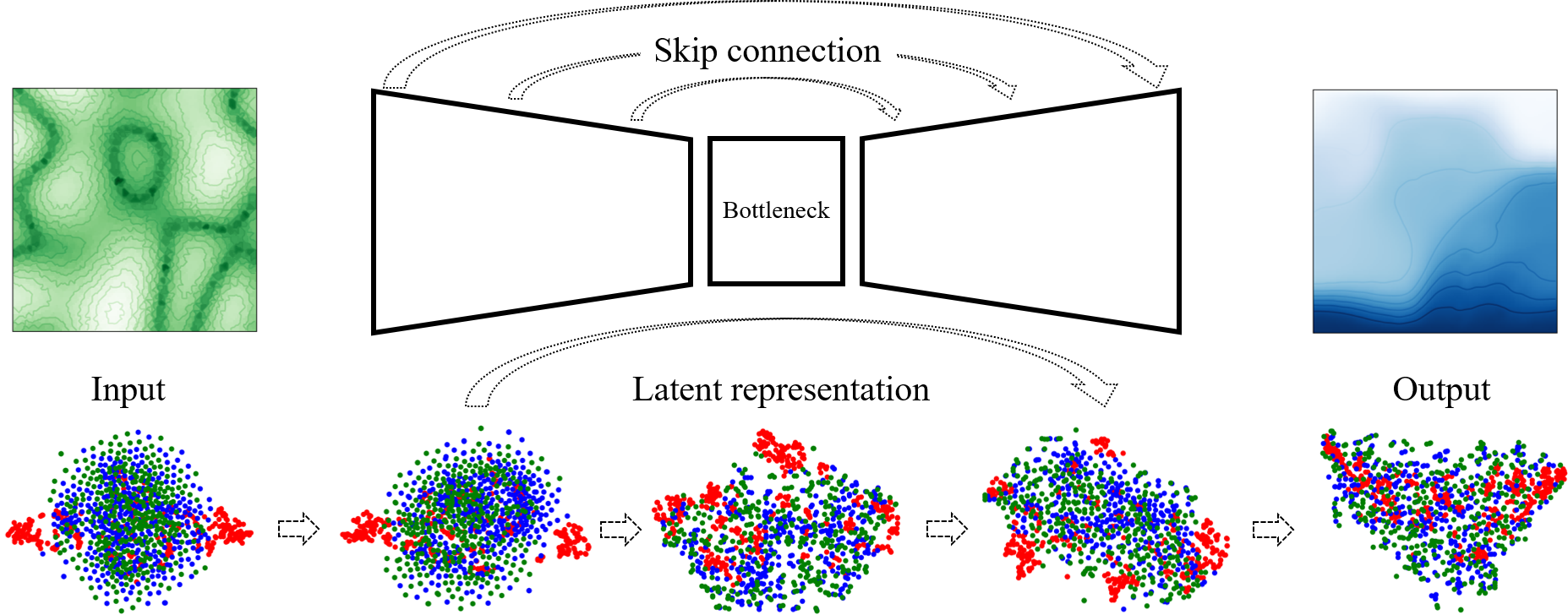}
\caption{Illustration of strongly heterogeneous input and output fields. Our framework aims to handle continuous and/or discontinuous input and output fields through nonlinear data compression and skip connections. Red, blue, and green dots represent t-Distributed Stochastic Neighbor Embedding (t-SNE) under three different statistical distributions. We illustrate that red, blue, and green dots distribute differently for each stage of nonlinear compression. In the latent representation, all colors mix with each other. Multiscale features are captured progressively through skip connections, so that decoded input samples become uncorrelated and evenly distributed in the latent space. Therefore, without skip connection (multiscale data transfer), the output field might lose some essential features (see \ref{sec:no_unet}). }
\label{fig:intro_pic}
\end{figure}

Data-driven and physics-informed modeling enabled by machine learning techniques has recently gained significant attention in scientific computing as an alternative to the aforementioned reduced order modeling techniques. Directly incorporating physical laws, Physics-informed neural networks (PINN) were proposed to solve PDEs in forward and inverse settings without the need for labeled data \cite{raissi2019physics, kadeethum2020physics, fuhg2021mixed}. Their limitation lies in the fact that they can not efficiently handle problems with heterogeneous coefficients. Extended-PINN (xPINN) attempts to address this issue using domain decomposition \cite{jagtap2020extended}. However, it still cannot handle a high degree of spatial heterogeneity of the PDE coefficients as an input (see Figure \ref{fig:intro_pic}) since it is not practical to subdivide these types of fields. Even though PINN and its variants can be used for parameter estimation problems, they cannot efficiently provide multiple inquiries for the forward problem.  Data-driven learning of solution operators of PDEs has recently been proposed in DeepONet \cite{li2020neural} and Neural Operator \cite{li2020neural}, which have been shown to approximate the solution of PDEs in a forward solution setting. These methods, however, are limited to an approximation of continuous functions, which might not be applicable to problems that deal with discontinuous input and output functional spaces, as shown in Figure \ref{fig:intro_pic}. We note that even though the Neural Operator model \cite{li2020neural} could generalize between spatially heterogeneous input fields, discontinuities must be smoothed prior to training. Additionally, we have found the training time required for the latter approach is exceedingly high for standard engineering applications (see \ref{sec:com_no}). \par

This study focuses on a data-driven reduced order modeling approach 
geared towards HM processes in porous media with heterogeneous material properties, where resolving discontinuities is critical for the fidelity of the simulation. The data-driven approach is attractive because it does not require any modifications of FOM source codes, and subsequently, provides more flexibility when coupling to existing FOM platforms \cite{mignolet2013review,hesthaven2018non,xiao2015non2}. Along with the benefits of the data-driven approach, concerns arise regarding a lack of strict enforcement of physical laws. Navigating between the data-driven and physics-informed paradigms is a challenging task, often governed by the trade-off between speed-up and accuracy as well as the availability of data. \par

Central to the developments in this work is an idea of generative adversarial networks (GAN). Since the introduction of GAN \cite{goodfellow2014generative} in 2014, this architecture has gained popularity because of its generalization ability and has been widely used in image/video analysis \cite{goodfellow2014generative,liu2016coupled}. GAN is based on two networks, (1) a generator and (2) a discriminator, which compete with each other. The generator's goal is to produce realistic output that resembles the real data (belonging to the training set). The discriminator's job is to differentiate between the output of the generator and the real data. The training phase is complete when the discriminator cannot distinguish between real data belonging to the generator's training set and data produced by the generator. The concept of deep convolutional neural networks (CNN) has been employed for both the generator and the discriminator of GAN to capture highly heterogeneous spatial features \cite{odena2016deconvolution,radford2015unsupervised,yu2017unsupervised} and their evolution in time \cite{yoon2019time,naruse2019generative}. Additionally, an augmentation of discrete labeled data has been used to control the generator's output, termed conditional GAN or cGAN \cite{shen2020interpreting,mirza2014conditional,chen2016infogan,lee2021fast}. Still, all the aforementioned frameworks cannot be used directly to address the data-driven solution of PDEs since they are limited to only categorical conditioning, which is not sufficient for the parametrization of spatially heterogeneous input fields. \par



To approximate primary variables with given heterogeneous coefficients of the PDEs and to estimate the spatially heterogeneous coefficients given the primary variables, we employ the idea of the image-to-image translation framework \cite{isola2017image}. Different from a classical cGAN in which is conditioned by labeled data (e.g., classes of object the image is representing), the image-to-image translation framework relies on conditional fields, which could be edges, masks, points, or segmentation maps \cite{wang2018high,ledig2017photo,xu2018attngan,zhu2017unpaired,liu2017unsupervised,huang2018multimodal}. This idea has been extended to high-resolution images \cite{wang2018high,ledig2017photo}, unsupervised translation \cite{zhu2017unpaired,liu2017unsupervised,huang2018multimodal}, controllable facial attribute editing \cite{xu2018attngan}, and controllable segmentation mapping \cite{park2019semantic}. We rely on the image-to-image translation concept because we want to construct mappings of gradients, discontinuities, and multiscale features between input and output fields. We achieve these goals through (1) nonlinear data compression (latent representation) and (2) multiscale feature mapping (skip connections), as illustrated in Figure \ref{fig:intro_pic}. The work presented here is novel in two main ways


\begin{enumerate}

    \item We are the first to propose and adapt the image-to-image translation concept to learn forward and inverse solution operators of PDEs. This data-driven ROM framework can efficiently parametrize highly complex input fields and generalize mappings between those input fields and their corresponding output fields. We show that the framework can handle discontinuities in both input and output fields, which is different than the current state-of-the-art studies of data-driven learning of solution operators \cite{lu2021learning,li2020neural}. 
    
    \item We illustrate that using a U-Net generator \cite{ronneberger2015u} instead of the deep CNN (see \ref{sec:no_unet} for more detail and performance comparison), and a  patch-based discriminator \cite{demir2018patch} along with spectral normalization \cite{miyato2018spectral} or Wasserstein loss with gradient penalty \cite{arjovsky2017wasserstein,gulrajani2017improved} can improve the framework's accuracy and training stability without adding costly training time. In the forward setting, this framework provides a speed-up of at least 2000-fold while maintaining a high level of accuracy. For the inverse modeling in which input data is sparsely available and noisy, the framework delivers a speed-up of 120000 times compared to a Gaussian prior-based inverse modeling approach, while its accuracy is significantly improved. 
    



\end{enumerate}

\section{Results}

We present two possible applications of our proposed model; (1) forward modeling: with given permeability fields, the model approximates the field of the primary variables (i.e., pressure and displacement fields), and (2) inverse modeling: with given a subset of primary variable fields, the model estimates the permeability field. We utilize a Discontinuous Galerkin finite element model of linear poroelasticity developed in \cite{kadeethum2020enriched, kadeethum2020finite} to generate training and test sets. For input parameters to the FOM, we utilize three types of highly heterogeneous permeability fields: (1) a permeability field given as a Gaussian distribution (\ref{sec:forward_gs}), (2) a permeability field defined by a bimodal transformation (\ref{sec:forward_bm}), and (3) a permeability field from a Zinn \& Harvey transformation (\ref{sec:forward_zh}) \cite{zinn2003good}. We here only present results of the W model (corresponding to one out of three variants of our ROM), which is developed based on Wasserstein loss with gradient penalty (see \ref{sec:cgan_w}) as it provides the best accuracy and the most training stability. We compare the performance of three models, (1) base model (\ref{sec:cgan_base}), (2) SN model (\ref{sec:cgan_sn}), and (3) W model (\ref{sec:cgan_w}) in detail for both training and test dynamics in \ref{sec:forward_gs}, \ref{sec:forward_bm}, and \ref{sec:forward_zh}. 

\subsection{Forward modeling}\label{sec:forward}

\begin{figure}[!ht]
   \centering

         \includegraphics[keepaspectratio, height=3.5cm]{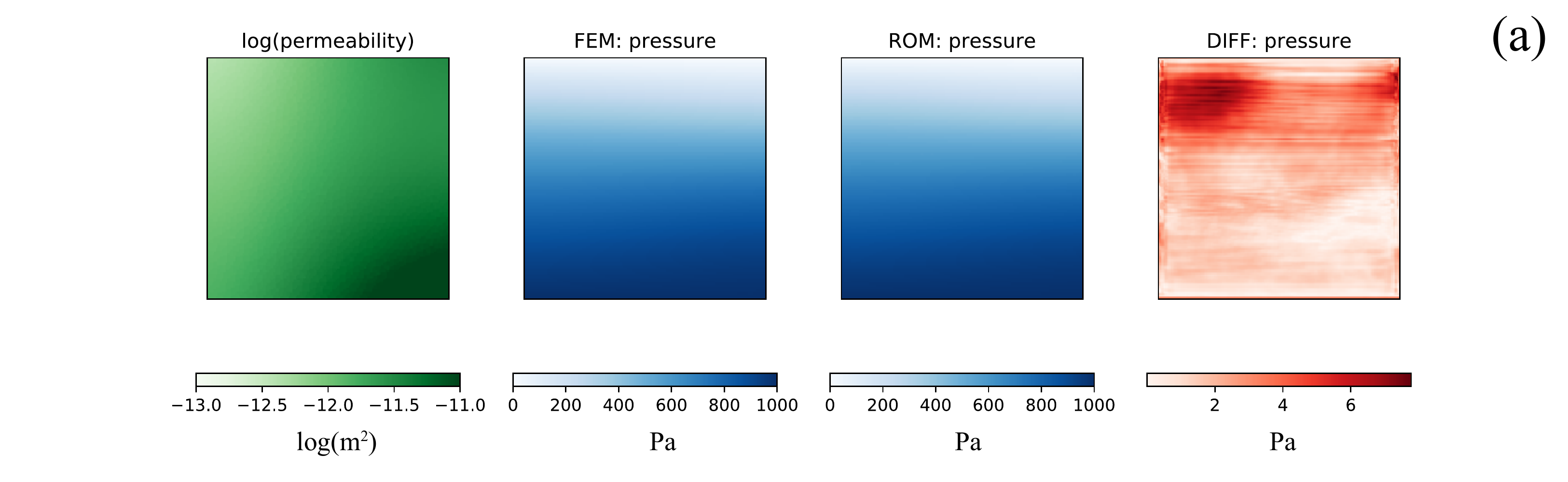}
         \includegraphics[keepaspectratio, height=3.5cm]{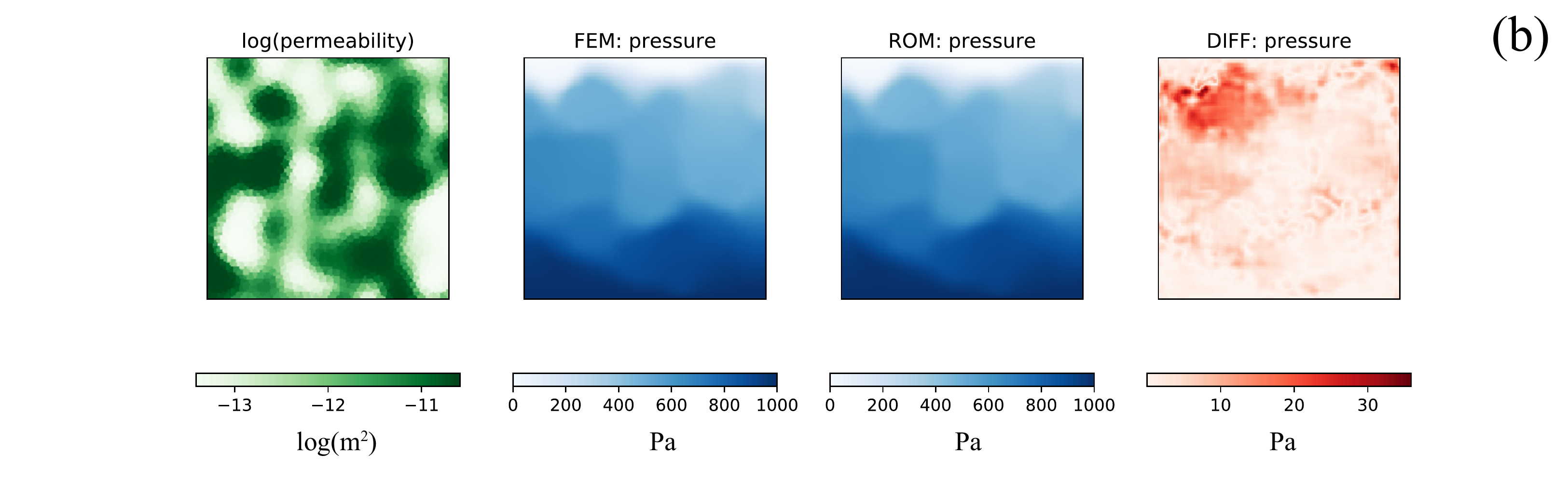}
         \includegraphics[keepaspectratio, height=3.5cm]{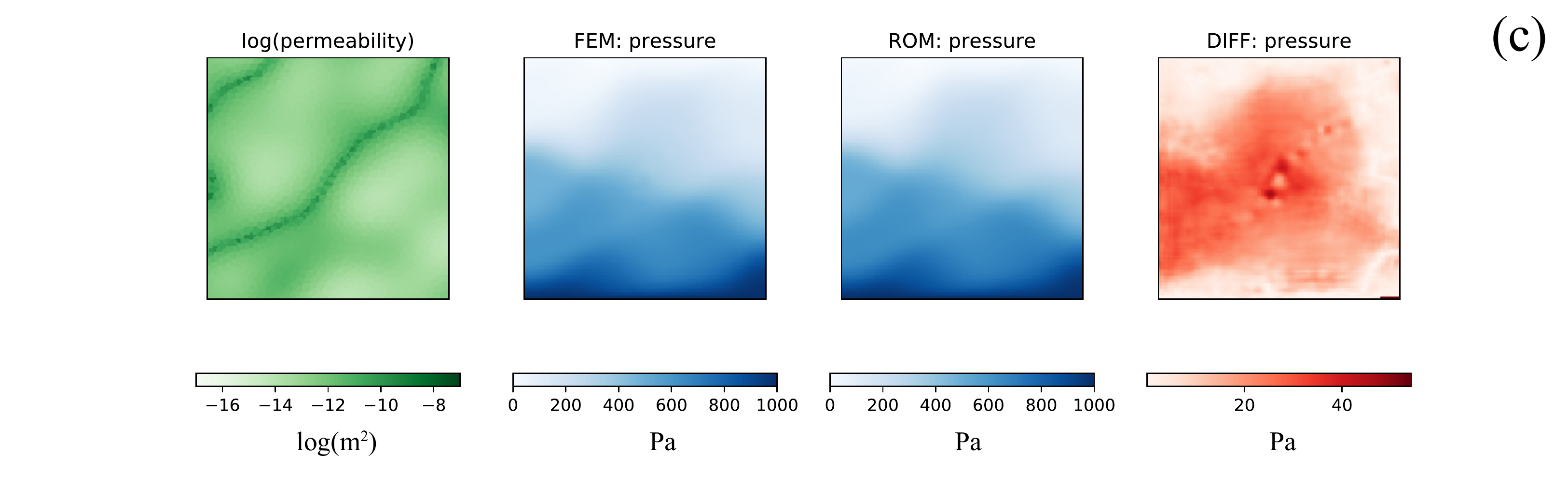}
   \caption{The results of the test case of the W model trained with 30000 examples (10000 for each Gaussian distribution, bimodal transformation, and Zinn \& Harvey transformation). The model then is tested using 3000 examples (1000 for each Gaussian distribution, bimodal transformation, and Zinn \& Harvey transformation). These three cases shown here are randomly picked from 3000 test examples. Note that the ranges (minimum and maximum values) of each permeability field are different. However, our model can still predict the pressure field with acceptable accuracy.}
   \label{fig:ex4_pic}

\end{figure}

Focusing on the steady-state solution of poroelasticity equations (as described in \ref{sec:governing_equations}) while considering highly heterogeneous material properties, we train and test the ROM based on three types of highly heterogeneous permeability fields simultaneously. The input to the ROM is three types of permeability field, and the output is pressure and displacement fields. Details on how we generate each permeability field, set model parameters, training, and testing of the ROM can be found in \ref{sec:forward_gs} for the Gaussian distribution, \ref{sec:forward_bm} for the bimodal transformation, and \ref{sec:forward_zh} for the Zinn \& Harvey transformation.  We illustrate three test cases (out of 3000 test examples, which are excluded from the training set) in Figure \ref{fig:ex4_pic}. The size of the training set will be discussed in the following paragraphs. From this figure, we note that our model can provide reasonable approximations of the FEM results. The difference between solutions produced by the FOM and ROM (further referred to as DIFF) is calculated by

\begin{equation}\label{eq:diff}
\operatorname{DIFF}(\bm{X})= \left|\bm{X}_h - \widehat{\bm{X}_h}\right|.
\end{equation}

\noindent
$\bm{X}_h$ is a finite-dimensional approximation of the set of primary variables, corresponding to displacement and pressure fields in this study, as calculated by the FOM. $\widehat{\bm{X}_h}$ is an approximation of $\bm{X}_h$ produced by the ROM. An extensive discussion can be found in the \textbf{problem description and methods} section. \par

The maximum DIFF values lie in a range from 0.01\% to 0.1\% (depending on the type of permeability of the test cases). When the ROM is trained and tested separately for each type of permeability field (see Figs. \ref{fig:ex1_pic}, \ref{fig:ex2_pic}, and \ref{fig:ex3_pic}), the maximum DIFF values remain in the same order of magnitude. This behavior indicates that the proposed ROM can generalize efficiently among three types of microstructures. As expected, the permeability field from the Zinn \& Harvey transformation case has the highest DIFF value since the permeability field exhibits extensive discontinuities over the domain and the highest contrast. The permeability field as bimodal transformation generally has greater DIFF values than the Gaussian distribution case. \par

We then investigate the effect of the number of training samples ($N$) on the model dynamics (i.e., accuracy and its trend). We have five cases with a different number of training examples; (1) $N=1500$ training examples, (2) 3000 training examples, (3) 7500 training examples, (4) 15000 training examples, and (5) 30000 training examples. As previously mentioned, we train and test our model using three types of permeability fields; hence, we divide the portion of each field equally. For instance, assuming we have 30000 training examples, we use 10000 samples from each permeability field. In line with this, for testing, we use $N=3000$ testing examples (1000 examples from each field) for all five cases.  \par

\begin{figure}[!ht]
   \centering
         \includegraphics[keepaspectratio, height=3.5cm]{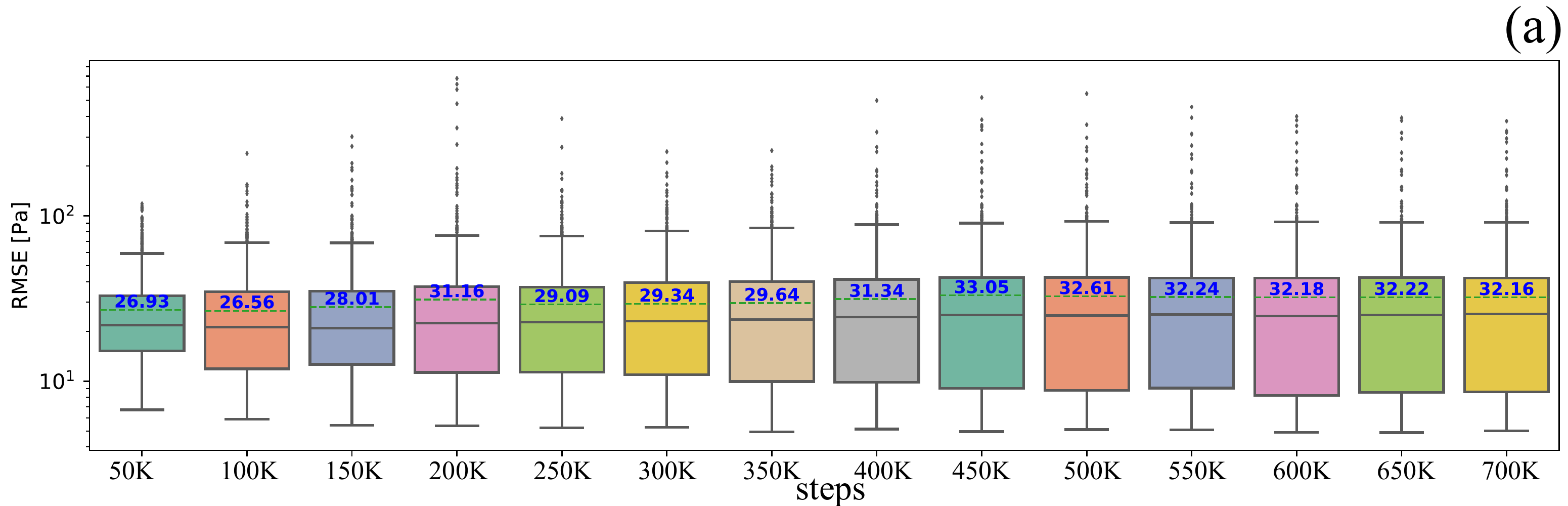}
         \includegraphics[keepaspectratio, height=3.5cm]{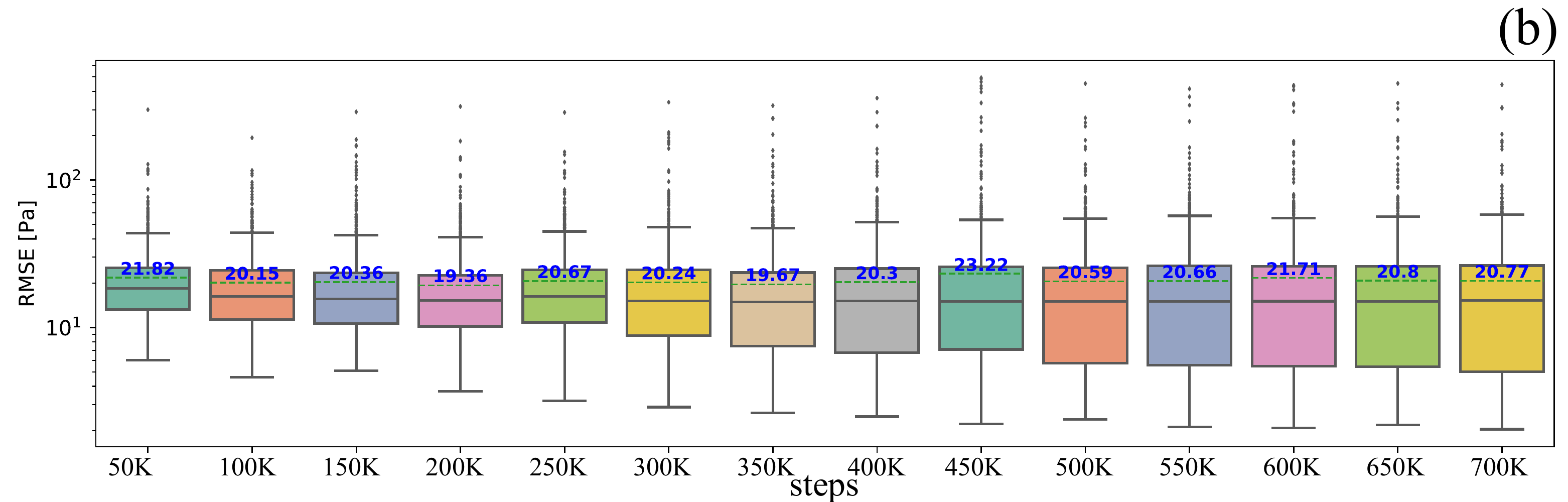}
         \includegraphics[keepaspectratio, height=3.5cm]{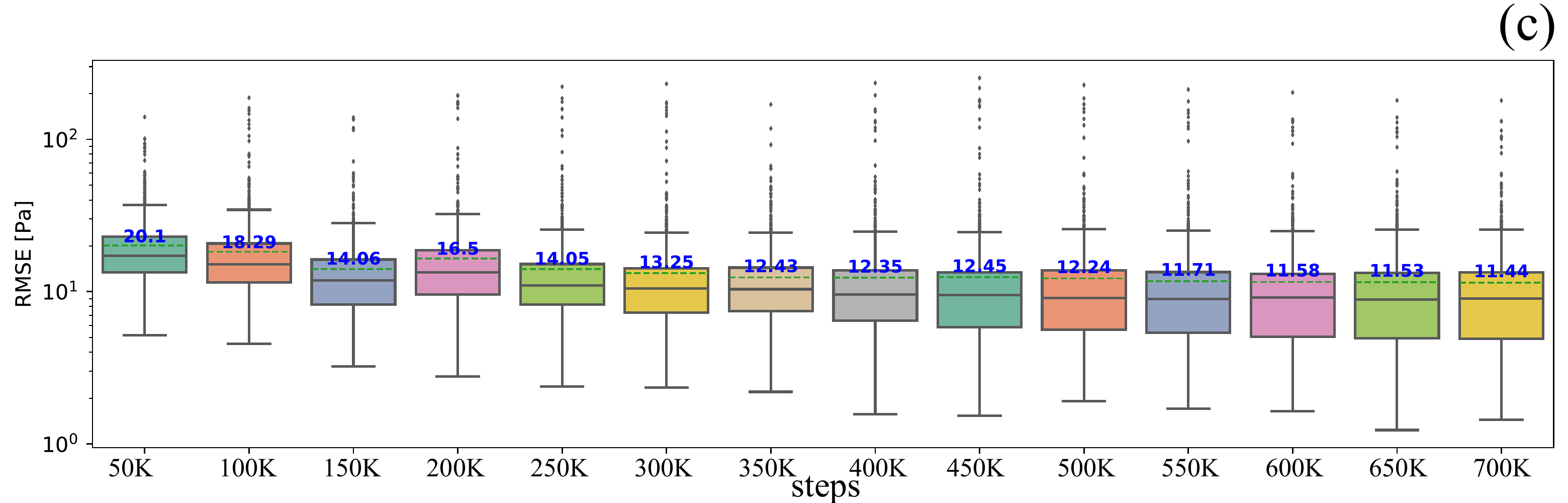}
         \includegraphics[keepaspectratio, height=3.5cm]{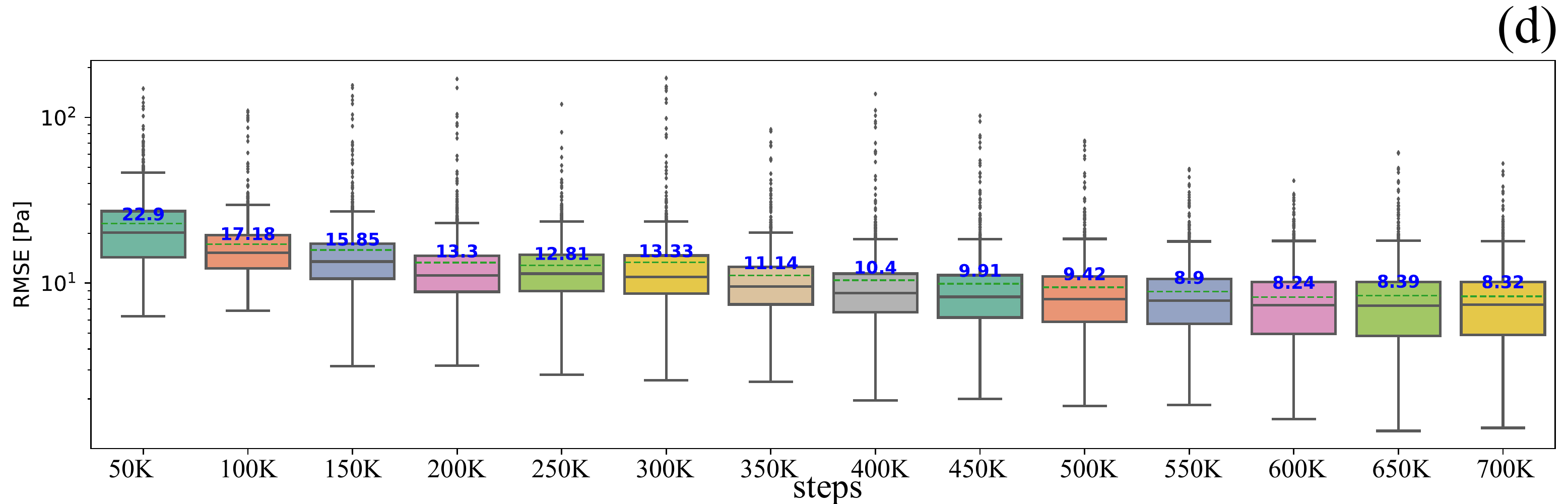}
         \includegraphics[keepaspectratio, height=3.5cm]{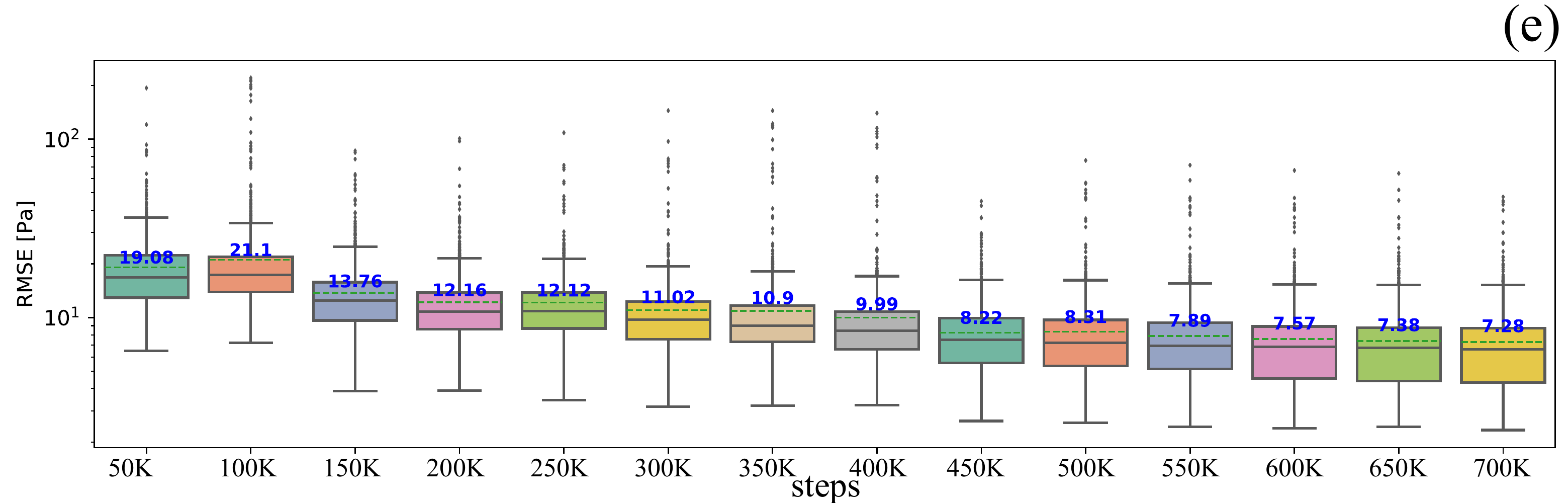}
   \caption{Root Mean Square Error (RMSE) over the training steps of the W model using (a) 1500 training samples, (b) 3000 training samples, (c) 7500 training samples, (d) 15000 training samples, and (e) 30000 training samples. We use an equal number of training examples for each permeability field. For instance, in (e), we use 10000 for each Gaussian distribution, bimodal transformation, and Zinn \& Harvey transformation. Each step refers to each time we perform back-propagation, including updating both generator and discriminator's parameters.}
   \label{fig:ex4_test}

\end{figure}

The root mean square error (RMSE) of the test cases, defined as

\begin{equation}\label{eq:rmse}
\mathrm{RMSE}=\sqrt{\frac{\sum_{i=1}^{N}\left(x_{i}-\hat{x}_{i}\right)^{2}}{N}}, \quad x_i \in \bm{X}_h \: \mathrm{and} \: \hat{x}_{i} \in \widehat{\bm{X}_h},
\end{equation} 

\noindent
is evaluated and presented in Figure \ref{fig:ex4_test}. We note that $x_{i}$ and $\hat{x}_{i}$ are the ground truth (FOM result) and approximated values (ROM result), respectively. We observe that as the number of training examples increases, the RMSE values and their trends become more accurate, see Figs. \ref{fig:ex4_test}a-e. As we decrease the number of training samples (see \ref{fig:ex4_test}a-c), we can observe more overfitting \cite{tetko1995neural}. These behaviors are discussed in detail in \ref{sec:inverse_random_vs_uniform} and \ref{sec:inverse_random_input_size}. Our numerical experiment shows that if we keep the number of training samples greater than 7500 (2500 examples for each field), our ROM framework performs well. \par

So far, we have seen that the ROM accuracy is acceptable when the W model is employed. Additionally, this model exhibits a reliable training behavior. We now investigate the W model's efficacy by comparing the gained speed-up and the lost accuracy using the ROM. We apply the multifidelity Monte Carlo (MFMC) idea and criteria used in \cite{o2018efficient}. The accuracy loss is quantified through a correlation between FEM and ROM using

\begin{equation}\label{eq:pc}
\rho_{{\bm{X}_h}, \widehat{\bm{X}_h}}=\frac{\operatorname{COV}\left({\bm{X}_h}, \widehat{\bm{X}_h}\right)}{\operatorname{SD}\left(\bm{X}_h\right) \operatorname{SD}\left(\widehat{\bm{X}_h}\right)},
\end{equation}

\noindent
where $\rho_{{\bm{X}_h}, \widehat{\bm{X}_h}}$ is the Pearson correlation coefficient of ${\bm{X}_h}$ with $\widehat{\bm{X}_h}$, $\operatorname{COV}(\cdot)$ is a co-variance, and $\operatorname{SD}(\cdot)$ is standard deviations. To quantify a trade-off between speed-up and accuracy, we use the following criterion \cite{peherstorfer2016optimal, o2018efficient}

\begin{equation}\label{eq:w1w2_pc}
\frac{w_{{\bm{X}_h}}}{w_{\widehat{\bm{X}_h}}}>\frac{1-\rho_{{\bm{X}_h},\widehat{\bm{X}_h}}^{2}}{\rho_{{\bm{X}_h},\widehat{\bm{X}_h}}^{2}}.
\end{equation}

\begin{figure}[!htt]
   \centering
        \includegraphics[keepaspectratio, height=5.0cm]{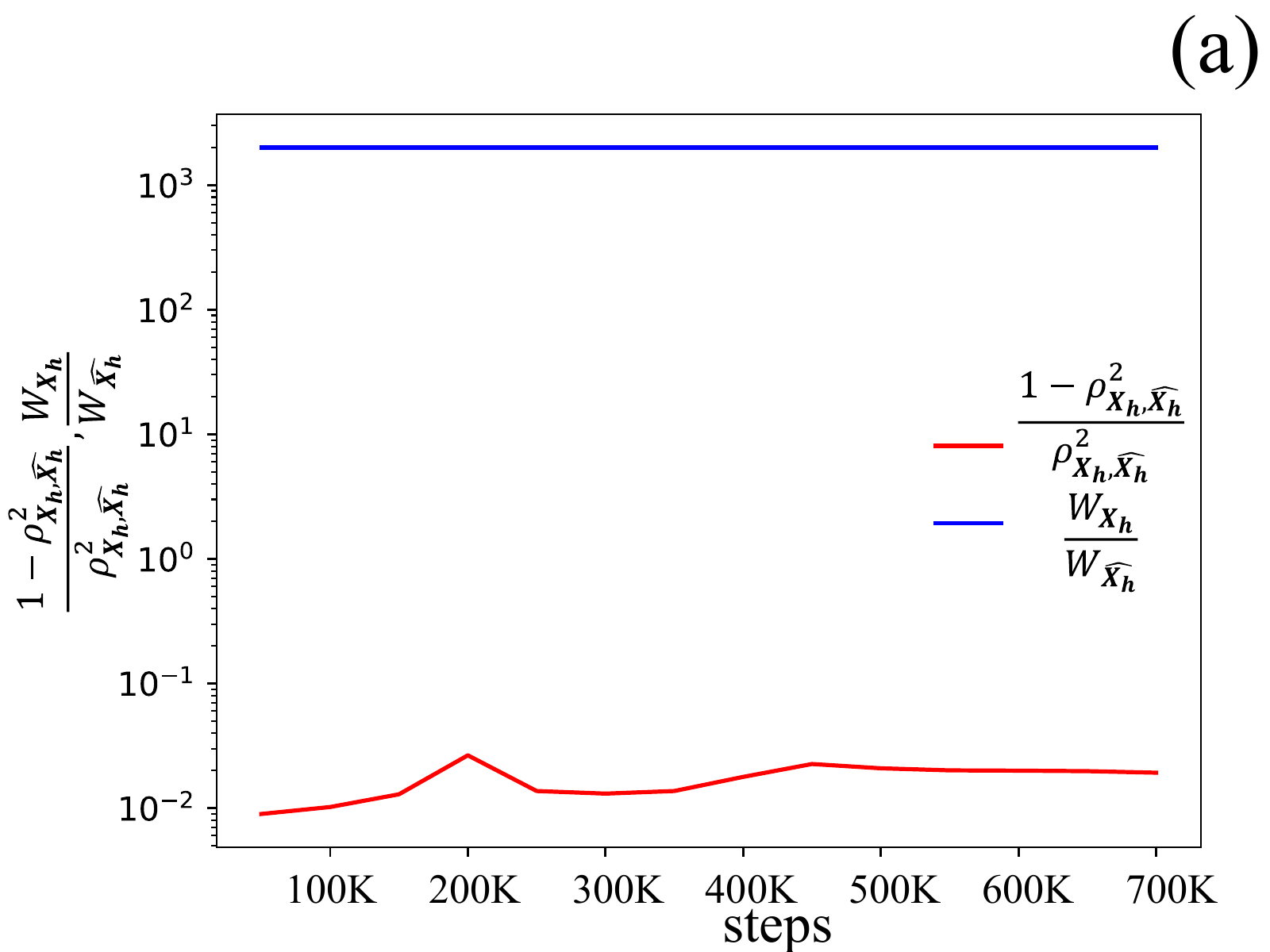}
         \includegraphics[keepaspectratio, height=5.0cm]{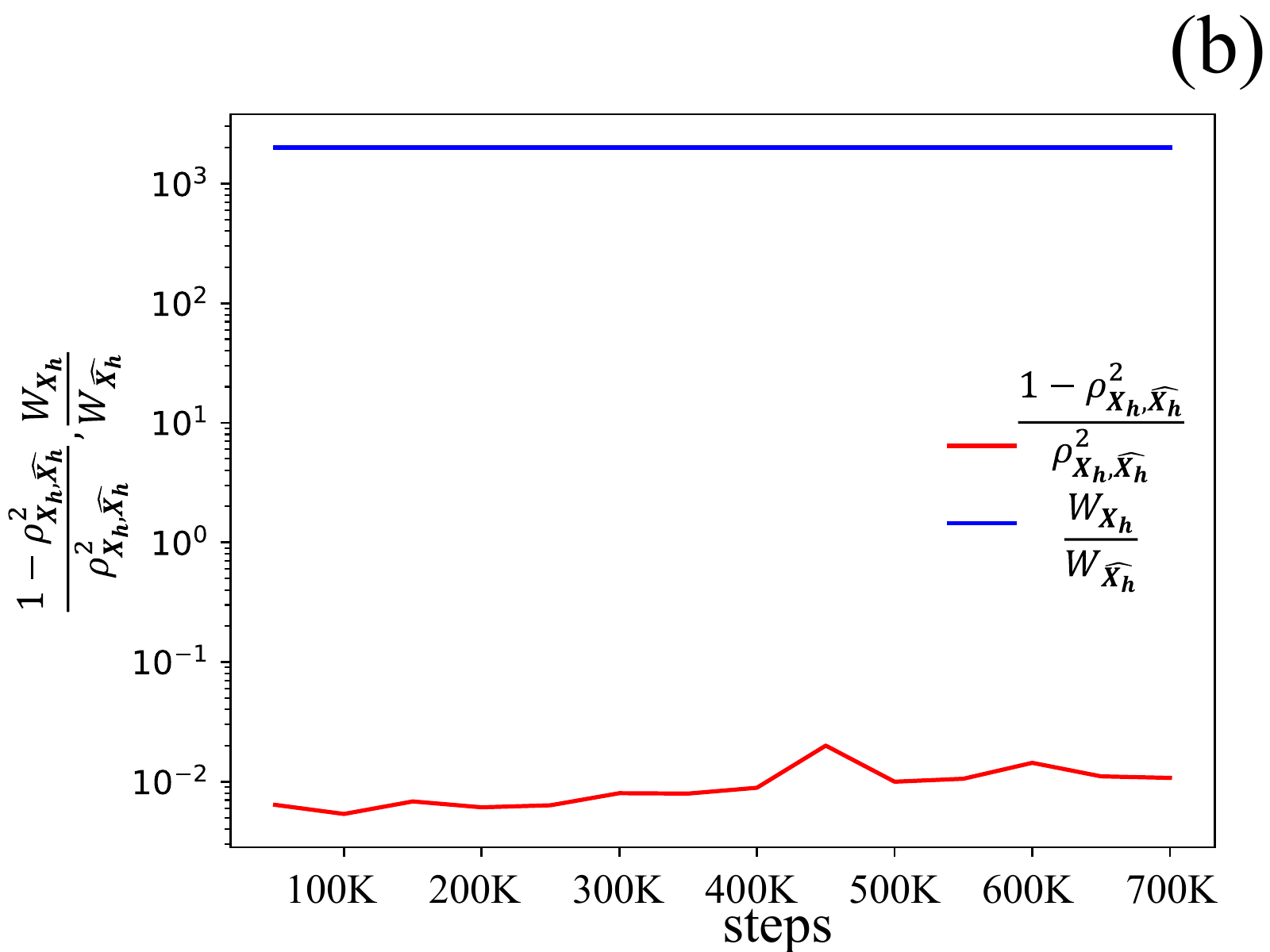}
         \includegraphics[keepaspectratio, height=5.0cm]{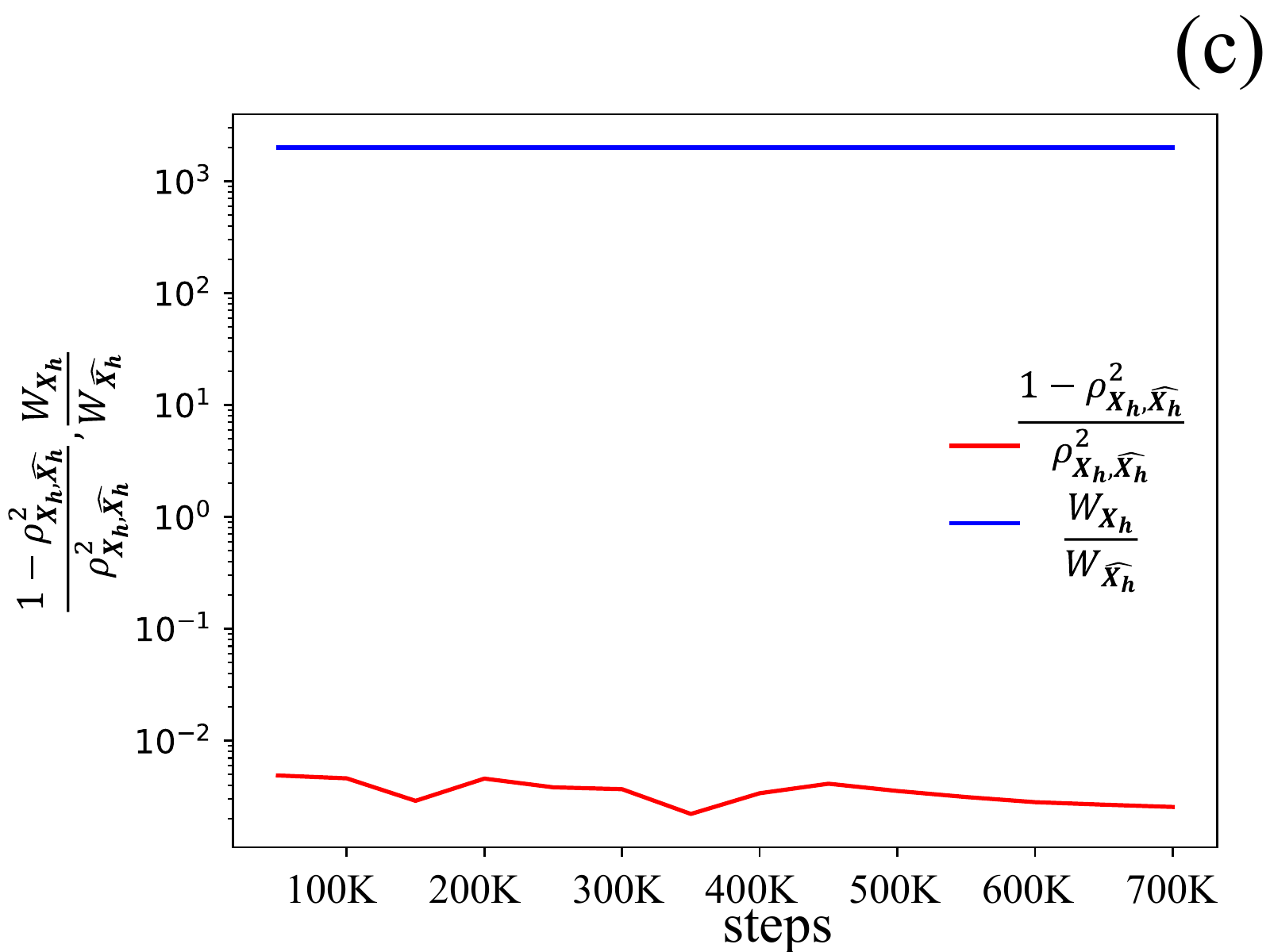}
         \includegraphics[keepaspectratio, height=5.0cm]{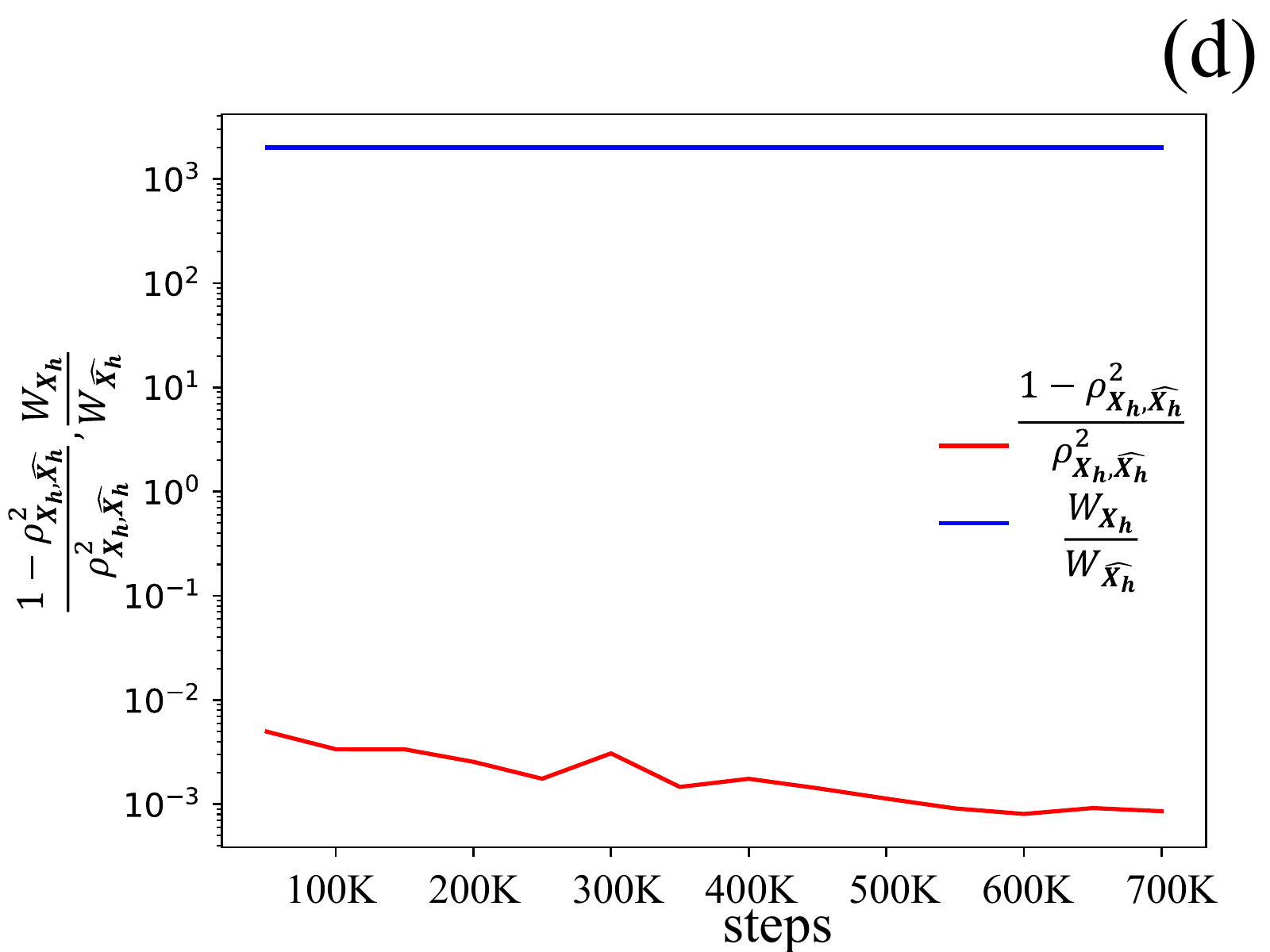}
         \includegraphics[keepaspectratio, height=5.0cm]{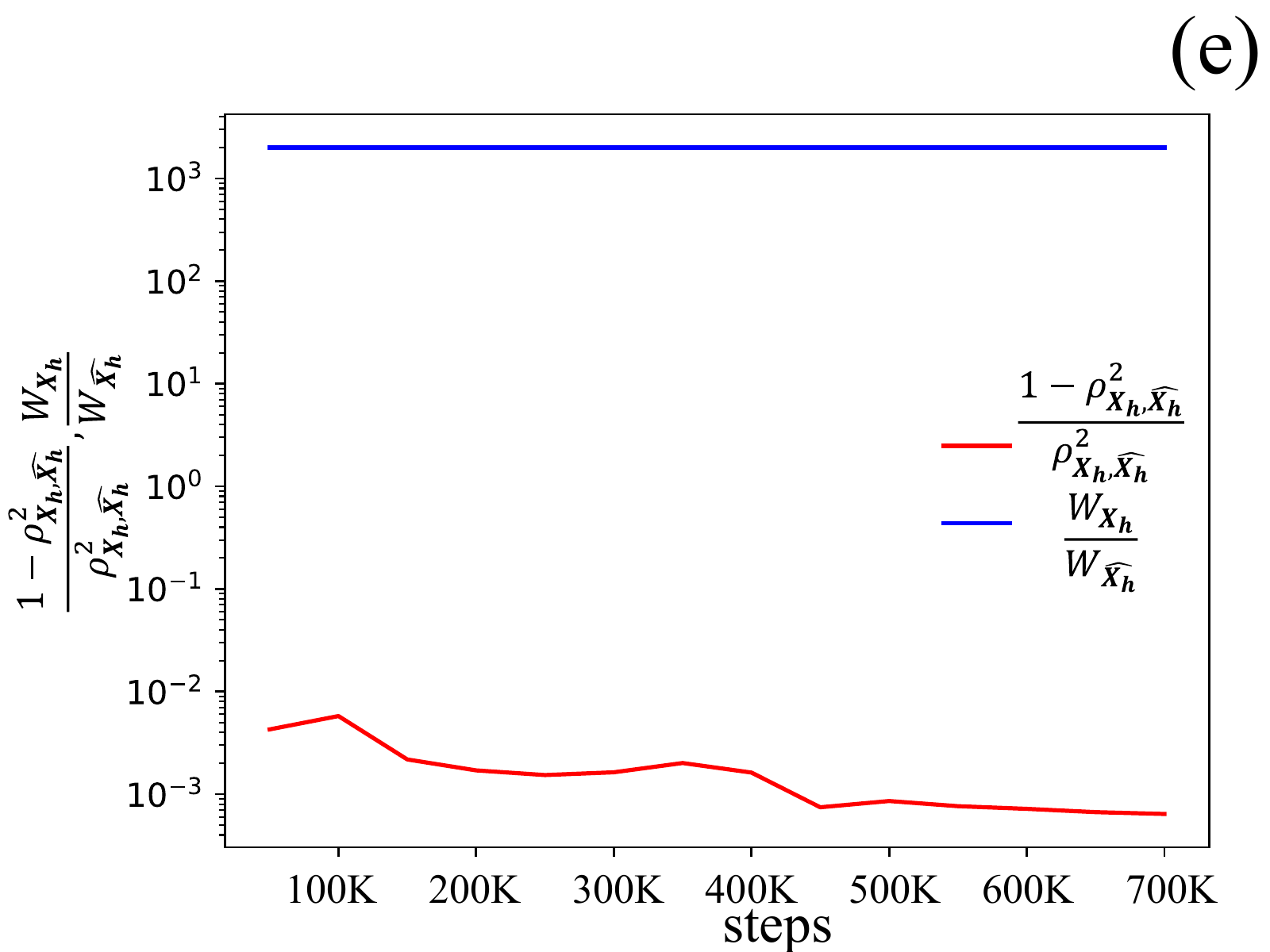}

   \caption{Results of \eqref{eq:w1w2_pc} for W model using (a) 1500 training examples, (b) 3000 training examples, (c) 7500 training examples, (d) 15000 training examples, and (e) 30000 training examples. We use an equal number of training examples for each permeability field. For instance, in (e), we use 10000 for each Gaussian distribution, bimodal transformation, and Zinn \& Harvey transformation. Each step refers to each time we perform back-propagation, including updating both generator and discriminator's parameters.}
   \label{fig:ex4_3F_pc}
\end{figure}

\noindent
Here, $w_{{\bm{X}_h}}$ and $w_{\widehat{\bm{X}_h}}$ denote the computational cost in wall times for computing ${\bm{X}_h}$ and $\widehat{\bm{X}_h}$, respectively. $w_{{\bm{X}_h}}$ and $w_{\widehat{\bm{X}_h}}$ are approximately 4 and 0.002 seconds resulting in a speed-up of 2000. As we solved a time-dependent set of equations to obtain the steady-state response, we utilized ten steps of the time-integration scheme, resulting in 40 seconds for the FOM steady-state solution indicating a speed-up of 20000. Here, we compare to the time of just one solution step of the FOM corresponding to a linear problem for a fair comparison. Any time-dependent or nonlinear problem we lead to a greater speed-up.  We present the results of \eqref{eq:w1w2_pc} for (1) 1500 training samples, (2) 3000 training samples, (3) 7500 training samples, (4) 15000 training samples, and (5) 30000 training samples in Figure \ref{fig:ex4_3F_pc}. From this figure, we observe that all five cases satisfy \eqref{eq:w1w2_pc} (i.e., the blue line is always above the red line). Moreover, as the number of training examples increase the differences between FEM and ROM results decrease (see $\frac{1-\rho_{{\bm{X}_h},\widehat{\bm{X}_h}}^{2}}{\rho_{{\bm{X}_h},\widehat{\bm{X}_h}}^{2}}$ value). \par

Note that \eqref{eq:w1w2_pc} does not account for the time used to train the ROM framework. On average, it took us 12 hours to prepare the base and SN models and 14 hours to train the W model with a Quadro RTX 6000 and 24GB RAM. We train all of our models for all cases for 700000 steps. Again, a step refers to a back-propagation of the loss, including updating both generator and discriminator's parameters throughout this section. The benefit of the proposed framework is that it is a one-time cost for the offline training phase. Then the framework could be inquired many times with much cheaper cost while providing sufficient accuracy (refer to Figure \ref{fig:ex4_3F_pc}). \par

We propose the following thought experiment; a small oil \& gas company needs to make decisions daily about operating ten wells. Their goal is to maximize the expected net present value (NPV). One way to compute the expected NPV is to run 1000 FEM simulations for each of the ten wells every day. This results in 10000 daily simulations. This operation might potentially be too expensive for this small company. An alternative is to use MFMC with the FEM and ROM frameworks. In this case, the company has a one-time cost of, say, 10000 model runs to train the ROM framework. On a daily basis, the company could perform 10 FEM runs, and 10000 ROM runs for each of their ten wells and get better accuracy than 1000 FEM runs per well. In this circumstance, a considerable amount of time and computing resources is saved compared to the 10000 daily FEM simulations that a Monte Carlo (MC) simulation would require. The savings become even more pronounced if the company considers multiple ways to operate the wells each day (e.g., increase the production rate, decrease the production rate, keep it the same). We note that the framework proposed here could also handle different boundary conditions by adjusting the input field to the generator and discriminator.  \par

\subsection{Inverse modeling}\label{sec:inverse}

In contrast to the forward modeling setup where the coefficients were the input and the solution field in terms of the primary variables was the output, here we use a subset of pressure and displacement fields at the steady-state solutions of the FOM as an input, i.e., measurements, to the ROM framework. The model's output is the reconstructed permeability field discussed in Section \ref{sec:forward_zh}. 
The model's input could have one, two, or three channels (i.e., pressure, displacement in the x-direction, and displacement in the y-direction). Throughout this section, we only show two cases; (1) input has one channel - pressure field, and (2) input has three channels - fields of pressure, displacement in the x-direction, and displacement in the y-direction. We note that one could also utilize the ROM framework as a forward approximator (as we did in the \textbf{forward modeling} section) combined with an optimization algorithm to solve the inverse problem. \par


We test our framework in an inverse setting against the Principal Component Geostatistical Approach (PCGA) \cite{lee2014large,lee2016scalable}, which is a scalable Hierarchical Bayesian model with Gaussian prior. PCGA utilizes the low dimensional space of the prior covariance, i.e., principal components, to approximate the maximum a posteriori estimate as to the inverse solution with its linearized uncertainty and is considered as one of the state-of-the-art inverse modeling methods in geoscience \cite{ghorbanidehno2020recent}. It only requires a few hundreds of the forward model runs in total without any intrusive implementation and accelerates dense matrix operations through the linearly scalable fast linear algebra methods \cite{wang2021pbbfmm3d}. We use PCGA in conjunction with our FOM solver \cite{kadeethum2020enriched, kadeethum2020finite} in this study. We reduce the available data points to $3\%$ of the input in a finite-dimensional setting, and we also illustrate the performance of the framework using noisy data. Note that we represent no measurements data using a flag of -1 (e.g., -1000 in a real value for pressure field). \par

\subsubsection{Training and testing data is not corrupted by noise} \label{sec:8_1}

The details of cGAN architecture, parameters, and loss functions could be found in \ref{sec:cgan}. We note that we only use the W model (\ref{sec:cgan_w}) in the current comparison. Moreover, the input field of both training and testing data is not corrupted by noise. We note that this practice is rarely done in inverse modeling literature since data acquisition is always subject to errors. However, this problem is used to test our framework robustness before we progress to the noisy data in the following section. The PCGA parameters are set as follows; the number of principal components is 100, the prior distribution of log permeability follows a Gaussian distribution with a prior standard deviation of $2$ log(\si{m^2}) and exponential covariance with a scale length of $[0.1, 0.1]$. The variance of the measurement error is set to $10^2$ \si{Pa^2}.  \par 

\begin{figure}[!ht]
   \centering
          \includegraphics[keepaspectratio, height=3.5cm]{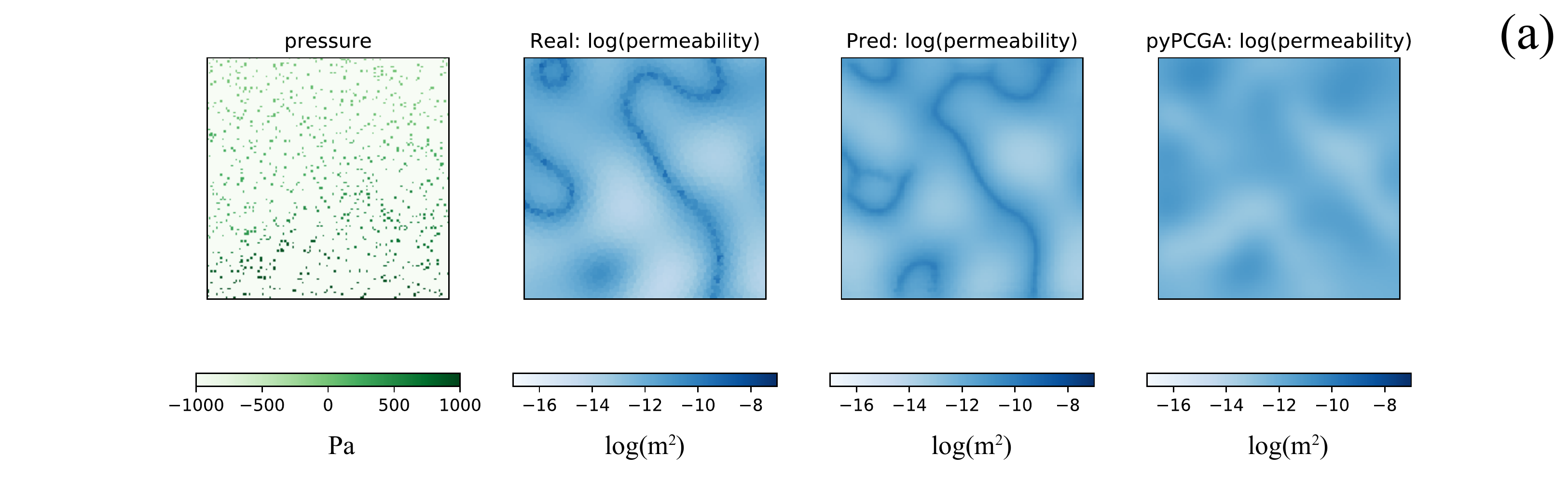}
        \includegraphics[keepaspectratio, height=3.5cm]{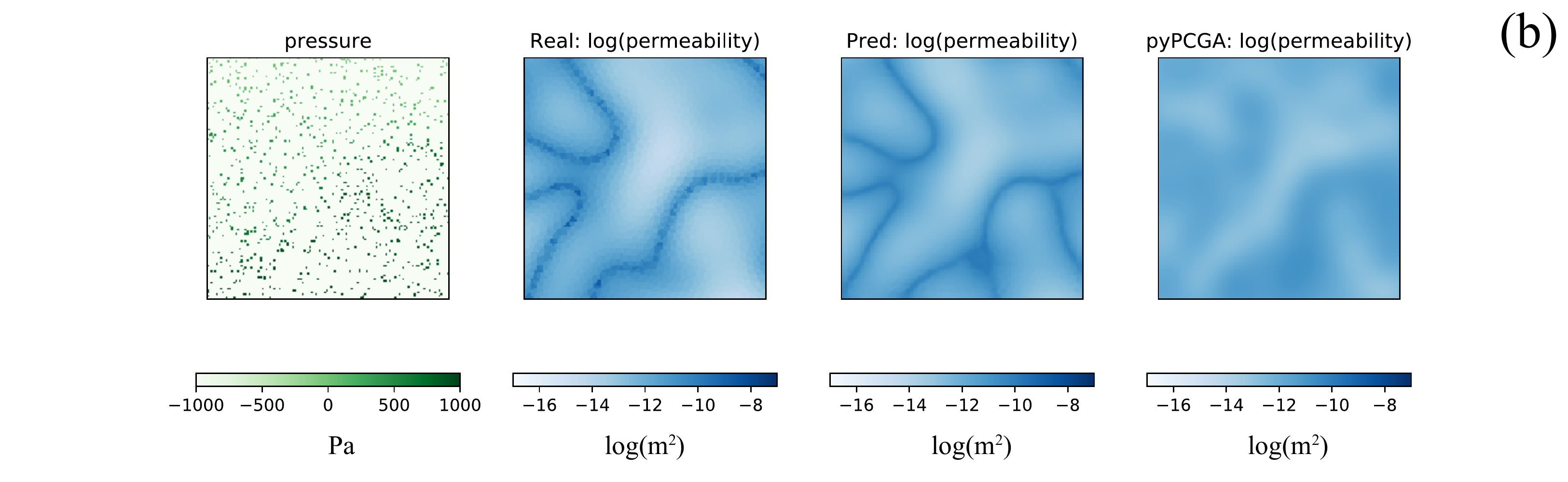}
        \includegraphics[keepaspectratio, height=3.5cm]{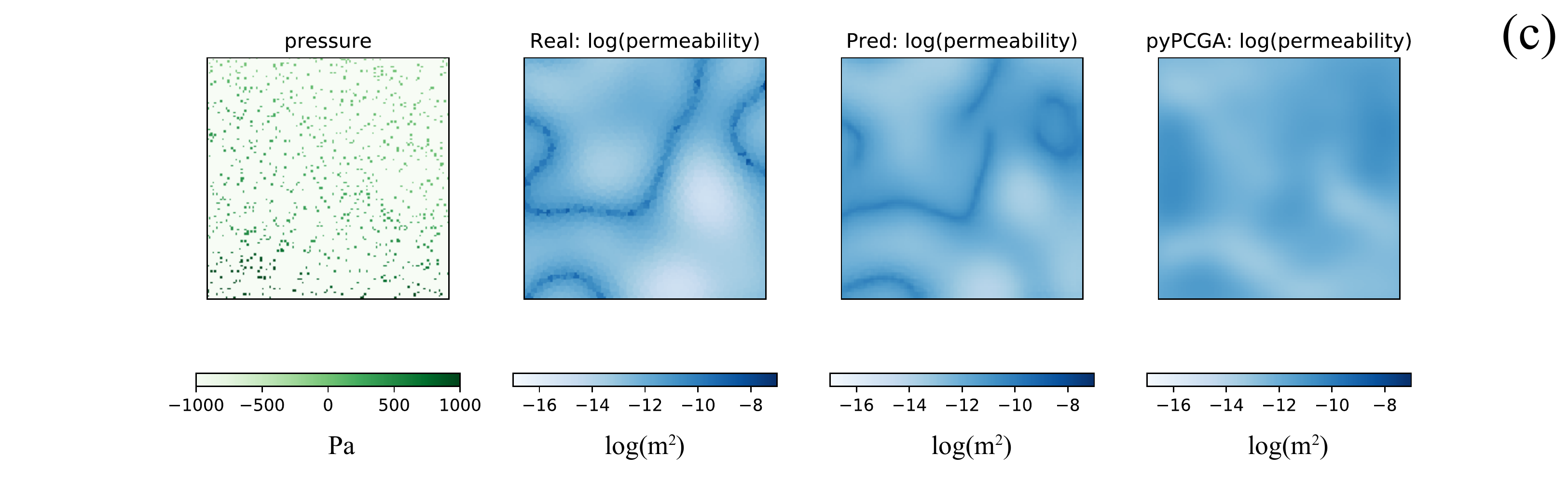}
   \caption{Three test cases' results (a), (b), and (c) of the W model and pyPCGA using 3\% of input fields. For the W model, we use only pressure as its input, and the number of training examples is 10000.}
   \label{fig:ex8_1_test}
\end{figure}

The comparison between the W model and PCGA is shown in Figure \ref{fig:ex8_1_test} for three test cases. We note that we only use pressure as an input field for both models. Since we consider the coupled hydro-mechanical problem as in \ref{sec:governing_equations}, the pressure data is related to the displacement through the coupled PDEs so that the pressure data alone may characterize spatially distributed permeability field better than those obtained from the pressure data in the single-phase flow configuration\cite{kang2017improved,lee2018fast}.  Moreover, we use the result from the checkpoint at 50000 steps for the W model. From Figure \ref{fig:ex8_1_test}, we observe that the W model could capture most of the structures and details with an acceptable error. On the other hand, the PCGA's inversion with spare pressure measurements does not provide accurate results as expected since the estimated permeability fields are smoothed due to Gaussian prior assumption and lack of the displacement information that the pressure data alone cannot capture. The RMSE of the W model is from 0.17 to 0.33 log(\si{m^2}) while the RMSE of the PCGA is from 0.58 to 0.63 log(\si{m^2}). \par

We run the PCGA model using 36 cores (AMD Ryzen Threadripper 3970X) using the python package pyPCGA \url{https://github.com/jonghyunharrylee/pyPCGA}, and the inversion takes approximately four to five minutes to converge with six to eight Gauss-Newton iterations. We train the W model on a single Quadro RTX 6000, and it takes around 1 hour for the model at the 50000 steps checkpoint. For the online phase or prediction, the W model takes about 0.002 seconds. Hence, the W model could provide a speed-up of 120000. We note that even though the W model must be trained, it could be used repeatedly to estimate the permeability field much faster than the pyPCGA approach.  \par

\begin{figure}[!ht]
   \centering
         \includegraphics[keepaspectratio, height=3.5cm]{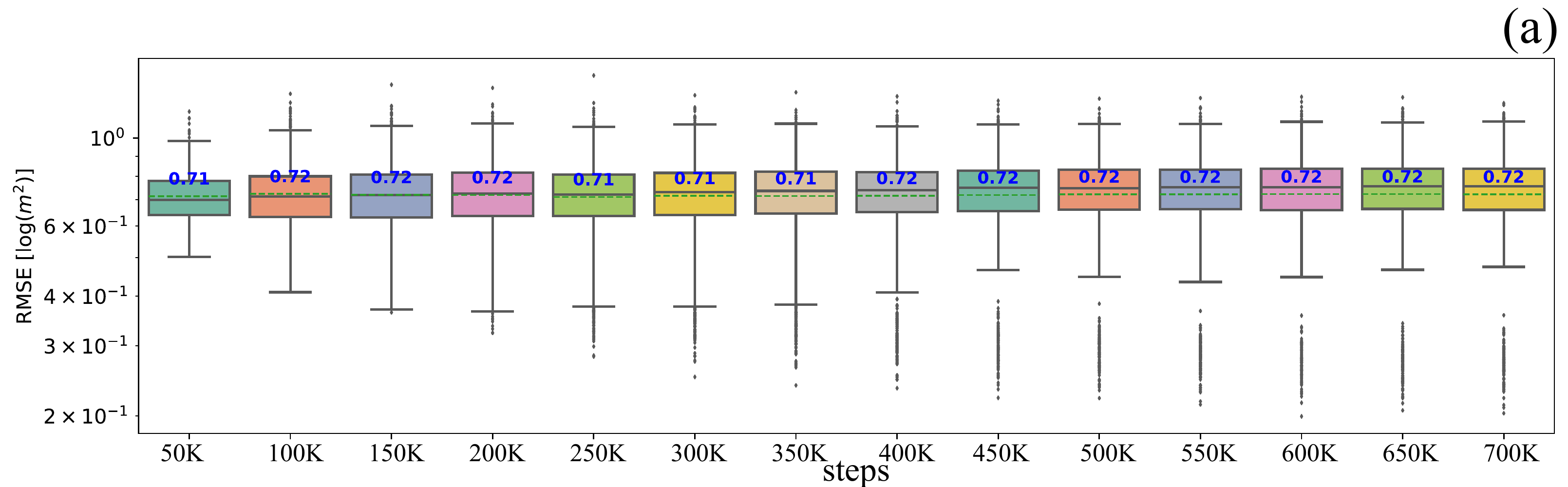}
         \includegraphics[keepaspectratio, height=3.5cm]{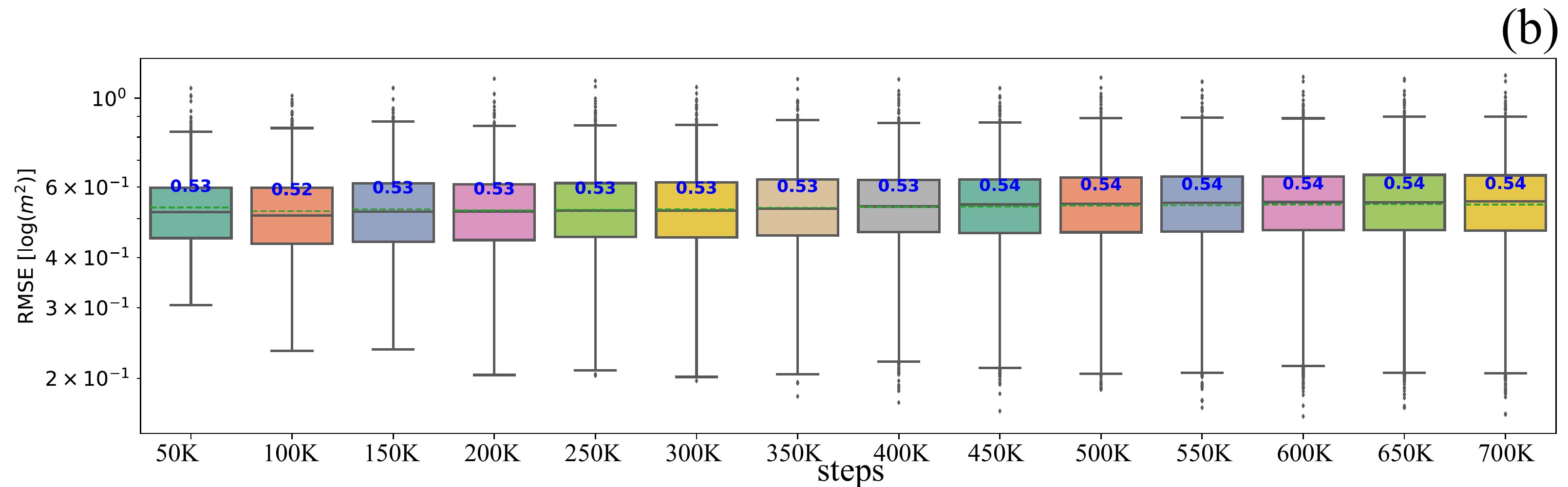}
   \caption{Root Mean Square Error (RMSE) of W model using 3\% of input fields (a) pressure as input and (b) both pressure and displacement as input. The number of training examples is 10000. Each step refers to each time we perform back-propagation including updating both generator and discriminator's parameters.}
   \label{fig:ex8_1_loss}
\end{figure}

We also illustrate that with more input data (pressure and displacement measurements as input), the W model could provide a better approximation, see Figure \ref{fig:ex8_1_loss}. The average RMSE values of the checkpoint at 50000 steps are 0.71 log(\si{m^2}) and 0.52 log(\si{m^2}) for the W model using only pressure as input and both pressure and displacement as input, respectively. The RMSE behavior clearly shows the overfitting behavior, as also observed in \ref{sec:inverse_random_vs_uniform} and \ref{sec:inverse_random_input_size}. This observation suggests that as the number of available data decreases, we should train the model using fewer steps. \par

\subsubsection{Training and testing data is corrupted by noise} \label{sec:8_2}

Next, we perform a study of the effect of added noise, which is created from the true data (${\bm{X}_h}$) as follows \cite{raissi2019physics,kadeethum2020pinn,Kadeethum2020ARMA}
\begin{equation}\label{eq:add_noise}
{\bm{X}_h}_{\text{noise}} = {\bm{X}_h} + \epsilon \operatorname{SD}\left({\bm{X}_h}\right) \mathcal{G}\left(0,1\right),
\end{equation}
\noindent
where ${\bm{X}_h}_{\text{noise}}$ is the input data with noise. The constant $\epsilon$, which is set to $0.05$, determines the noise level as it is multiplied with the standard deviation of the input field. $\mathcal{G}\left(0,1\right)$ is a random value which is sampled from the standard normal distribution with mean and standard deviation of zero and one, respectively. \par

\begin{figure}[!ht]
   \centering
         \includegraphics[keepaspectratio, height=3.5cm]{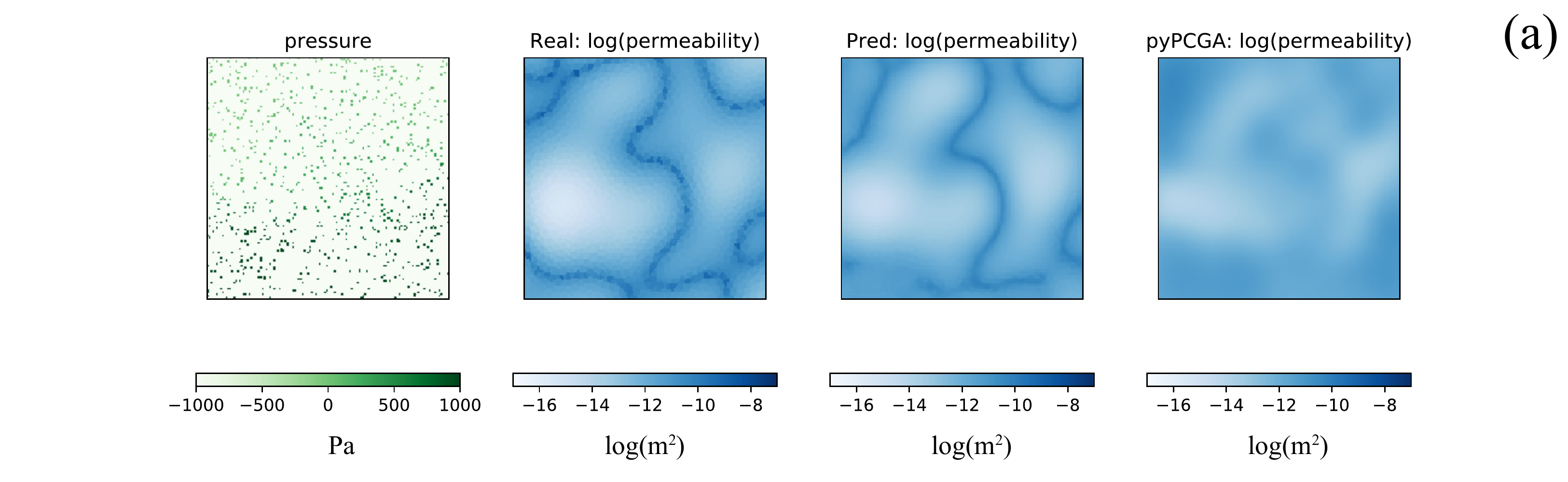}
        \includegraphics[keepaspectratio, height=3.5cm]{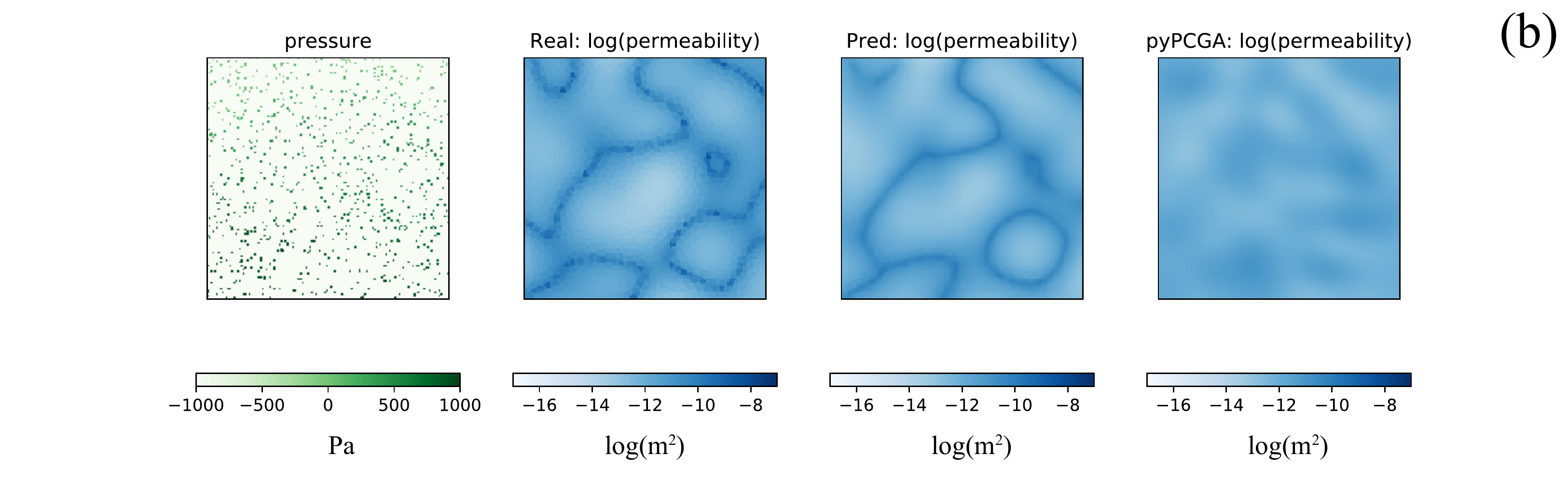}
        \includegraphics[keepaspectratio, height=3.5cm]{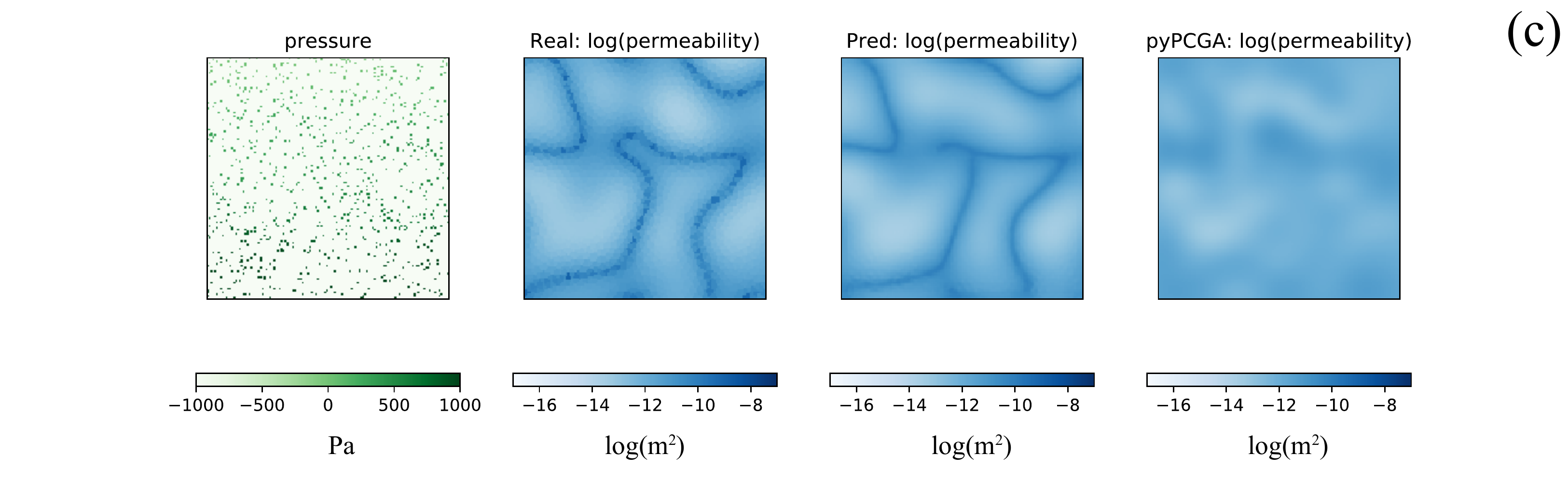}
   \caption{Three test cases' results (a), (b), and (c) with a noisy data of the W model and pyPCGA using 3\% of input fields. We use only pressure as its input, and for the W model, the number of training examples is 10000.}
   \label{fig:ex8_2_test}
\end{figure}

The comparison between the W model and PCGA with noisy data is illustrated in Figure \ref{fig:ex8_2_test}. Like the previous section, the W model could approximate the permeability field using only the pressure field as an input with RMSE from 0.39 to 0.63 log(\si{m^2}). The PCGA has approximately from 0.62 to 0.82 log(\si{m^2}) of RMSE value. These results illustrate that the W model is tolerant against noise in the input data. Again, during the online phase, the W model could estimate the permeability field was significantly more efficient with respect to computational time. We note that the variance of the measurement error is set to $25^2$ \si{Pa^2}.  \par

\begin{figure}[!ht]
   \centering
         \includegraphics[keepaspectratio, height=3.5cm]{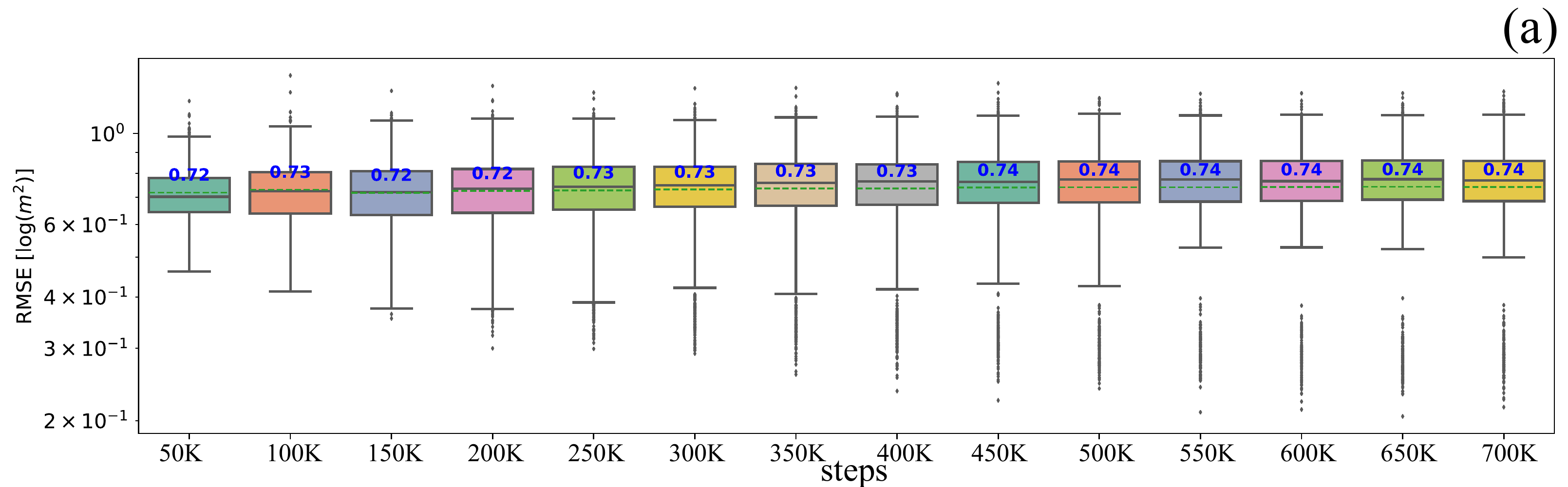}
         \includegraphics[keepaspectratio, height=3.5cm]{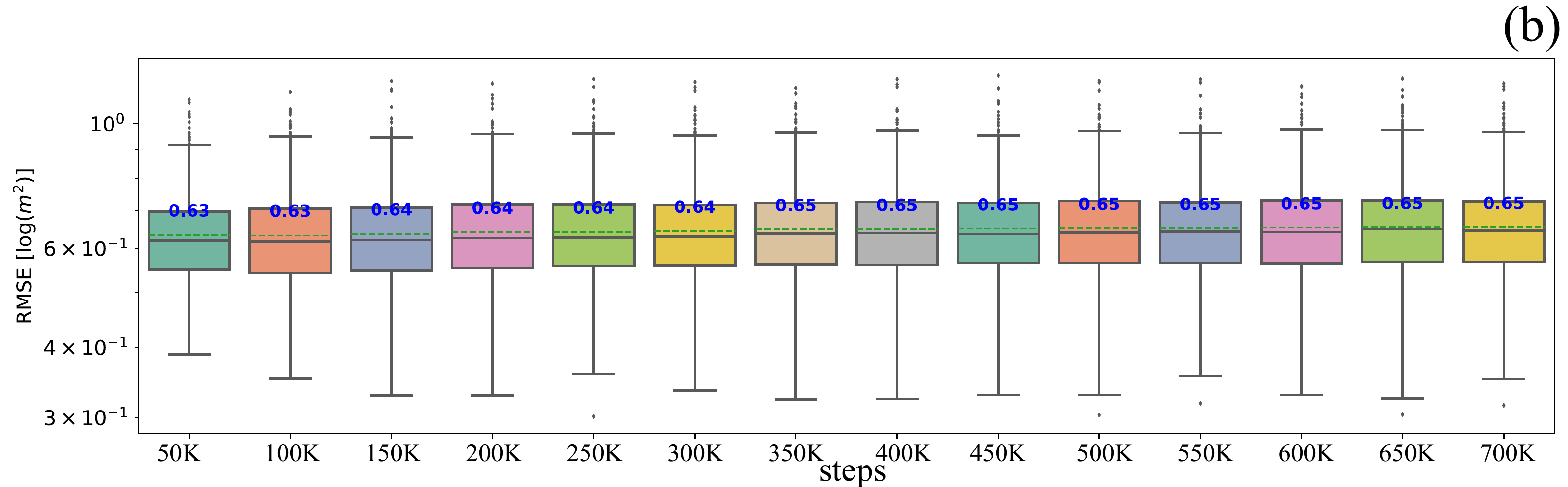}
   \caption{Root Mean Square Error (RMSE) of W model using 3\% of input fields (a) pressure as input and (b) pressure and displacement as input. The number of training examples is 10000. Each step refers to each time we perform back-propagation including updating both generator and discriminator's parameters.}
   \label{fig:ex8_2_loss}
\end{figure}

Figure \ref{fig:ex8_2_loss} presents the evolution of the RMSE behavior of the W model using (a) pressure as input and (b) both pressure and displacement as input. The average RMSE values of the checkpoint at 50000 steps are 0.72 log(\si{m^2}) and 0.63 log(\si{m^2}) for the W model using only pressure as input and pressure and displacement as input, respectively. While the RMSE is not much affected by the noise for the model with only pressure as input, the model with pressure and displacement as input suffers from the noise more severely (i.e., more input data, more noise effects). Similar to \ref{sec:inverse_random_vs_uniform} and \ref{sec:inverse_random_input_size}, and the previous section, the RMSE values illustrate the overfitting behavior, and it implies that as the number of available data decreases, we should train the model with an early stopping strategy. \par

\section{Discussion}

This manuscript presents a data-driven framework for solving a system of partial differential equations in a forward and inverse setting, focusing on coupled hydro-mechanical processes in heterogeneous porous media (\ref{sec:governing_equations}) where discontinuities commonly arise. Our framework is developed using cGAN image-to-image translation concept, which is composed of the U-Net generator and patch-based discriminator (\ref{sec:cgan}), as a means to learn the forward and inverse solution operator of a system of PDEs. Our approach uses the cGAN image-to-image translation concept to enable efficient parametrization of spatially heterogeneous input fields while mapping multiscale features to the output (see \ref{sec:no_unet} for more detail). The model has three variations: (1) base model with classic adversary loss (\ref{sec:cgan_base}), (2) SN model with spectral normalization (\ref{sec:cgan_sn}), and (3) W model with Wasserstein loss and gradient penalty (\ref{sec:cgan_w}). 


Based on \ref{sec:forward_gs}, \ref{sec:forward_bm}, and \ref{sec:forward_zh}, we have illustrated that the proposed framework can efficiently learn the forward solution operator and generate solutions outside of the training set, which have acceptable accuracy with respect to the full-order model (a Discontinuous Galerkin Finite Element framework in our case). It does so, even in exploring highly heterogeneous fields of material parameters, and more specifically, permeability fields in fluid-filled porous media. This observation suggests that our framework can be utilized to solve other linear and nonlinear PDEs, exploring the effect of heterogeneous material properties such as Young's modulus, Poisson's ratio, or porosity in the material and structural response. Although we have two primary variables in our HM problem, corresponding to the pressure and displacement fields, we only present the results of the pressure field for conciseness. The displacement field results also have the same range of relative RMSE. The RMSE values are two to three orders of magnitude less than the maximum value of the primary variables. Moreover, the ROM with W loss, referred to as the W model in this manuscript, showcases the best accuracy and the most stable training behavior. However, the W model is computationally more expensive to train than the base and SN models. The W model's training time for 700000 back-propagation steps is two hours greater than that of the base and SN models. \par

We then focus on the ability of the proposed framework to generalize in the \textbf{Results: Forward modeling} section. There, we demonstrate that the framework can approximate primary variables given highly heterogeneous permeability fields, which have significantly different characteristics and ranges. Again, we utilize three types of highly heterogeneous permeability fields: (1) a permeability field given as a Gaussian distribution, (2) a permeability field defined by a bimodal transformation, and (3) a permeability field from a Zinn \& Harvey transformation. The model with only 7500 training examples (2500 examples for each field) could provide the RMSE of two orders of magnitude less than the maximum value of the primary variable. This behavior shows that our model does not require extensive training sets, which is very beneficial in practice. The developed framework could give a speedup of approximately 2000 and still be able to capture most of the high-fidelity model results (finite element method). In \ref{sec:com_no}, we illustrate that the W model could handle a binary function as an input. Moreover, the W model has RMSE approximately four-order of magnitude less than the neural operator approach \cite{li2020neural}. This characteristic would be beneficial to various applications such as optimization, uncertainty quantification, and multiscale modeling. \par

\ref{sec:inverse_full} shows that the framework could also be used for inverse modeling (i.e., with given fields of primary variables, we inquire the model to approximate the material property distributions - the permeability field in this case). As we observe that the W model outperforms the base and SN models in terms of training stability and model's accuracy on the test set, we only use the W model in the inverse modeling part. The RMSE values are two orders of magnitude less than the maximum value of the permeability field. In addition, using both pressure and displacement fields as input, the model does not provide any significant incremental accuracy compared to the model with only pressure field as input since the pressure data contains information on the displacement field through the coupled HM PDEs. We note that this statement is valid only when the full data of the input field is provided. \par

In many realistic cases, however, the input data is sparse, especially in subsurface operations such as geothermal, groundwater, or hydrocarbon harvesting, since measurement points are strictly limited to the location of the wells or non-intrusive geophysics and InSAR data. Therefore, we use \ref{sec:inverse_random_vs_uniform} and \ref{sec:inverse_random_input_size} to demonstrate that our framework could also provide reasonable accuracy (i.e., still approximately two orders of magnitude less than the maximum value of the permeability field) in cases where available input data is incomplete (as low as six \% of the completed data). As expected, the more insufficient the data set is, the less accurate the model becomes. In contrast to \ref{sec:inverse_full}, where input data is complete, we could not observe a significant difference between using both pressure and displacement fields and using only pressure field as input; the W model with both pressure and displacement fields as input has higher accuracy than the W model with only a pressure field in cases where input data is incomplete. This observation reflects the fact that as the available input data is decreased (spatially), the additional information from the other fields, displacement field, in this case, could provide more information to the model and resulting in better accuracy. \par

We compare our framework in an inverse setting with Principal Component Geostatistical Approach (PCGA) method in \textbf{Results: Inverse modeling} section. We further reduce the available data from six \% to three \% of the completed data. Moreover, we test our model with cases where we have experimental measurements are corrupted by noise. Our results illustrate that the framework still could provide decent approximations of the permeability field as the RMSE values are two orders of magnitude less than the maximum value of the permeability field. Moreover, our model's performance is better than the PCGA approach since the estimated permeability fields of the PCGA approach are smoothed due to Gaussian prior assumption and lack of sensitivity in the pressure data alone. We also observe that the model with both pressure and displacement fields as input is more affected by noise in data than the model with pressure as an input alone. For a computational time comparison, our framework provides approximately a speed-up of 120000.  \par

\section{Problem description and methods} \label{sec:problem_statement}

\begin{figure}[!ht]
   \centering
         \includegraphics[keepaspectratio, height=7.5cm]{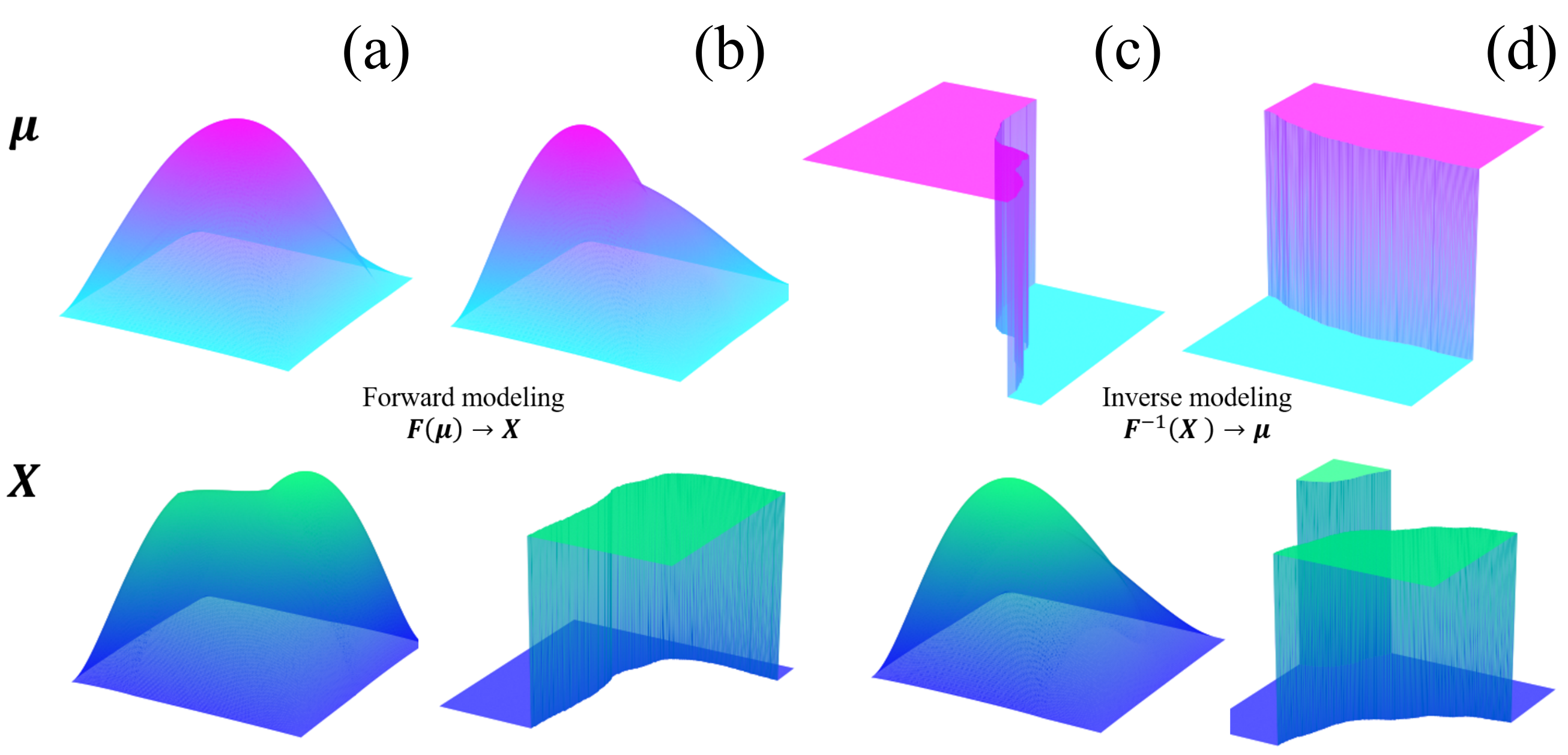}
\caption{Main goals of data-driven model reduction framework of PDEs: (a) $\bm{\mu}$ and $\bm{X}$ are continuous, (b) $\bm{\mu}$ is continuous, but $\bm{X}$ is discontinuous, (c) $\bm{\mu}$ is discontinuous, but $\bm{X}$ is continuous, and (d) $\bm{\mu}$ and $\bm{X}$ are discontinuous.}
\label{fig:idea_explain}
\end{figure}

Our main target is illustrated in Figure \ref{fig:idea_explain}. We consider a system of time-independent PDEs
\begin{equation} \label{eq:gen_pdes}
\begin{split}
\bm{F}\left( \bm{X}; \bm{\mu}\right) = \bm{0}  &\text { \: in \: } \Omega, \\
\bm{X} =\bm{f}_{D} &\text { \: on \: } \partial \Omega_{D},\\
- \nabla \bm{X} \cdot \mathbf{n}=\bm{f}_N &\text { \: on \:} \partial \Omega_{N}.
\end{split}
\end{equation}
\noindent
corresponding to quasi-static or steady-state problems, where $\Omega \subset \mathbb{R}^{n_d}$ (${n_d} \in \{1,2,3\}$) denotes the computational domain, and $\partial \Omega_{D}$ and $\partial \Omega_{N}$ denote the Dirichlet and Neumann boundaries respectively. $\bm{X}$ is a set of scalar ($\bm{X} \in  \mathbb{R}$) or tensor valued (e.g. $\bm{X} \in  \mathbb{R}^{n_d}\,\,\mathrm{or}\,\,\mathbb{R}^{n_d}\times\mathbb{R}^{n_d}$) generalized primary variables, and $\bm{\mu}$ are scalar ($\bm{\mu} \in  \mathbb{R}$) or tensor valued (e.g. $\bm{\mu} \in  \mathbb{R}^{n_d}\,\,\mathrm{or}\,\,\mathbb{R}^{n_d}\times\mathbb{R}^{n_d}$) generalized parameters for which a solution can be obtained in a finite-dimensional setting through a FOM. We are interested in developing a data-driven ROM to efficiently (1) approximate primary variables $\bm{X}$ for a solution of the system of PDEs given heterogeneous fields for the generalized parameters $\bm{\mu}$, and (2) estimate the generalized parameters $\bm{\mu}$ when given information from the solution of the primary variable field $\bm{X}$ that is full, partial or corrupted by noise. We note that $\bm{\mu}$ and $\bm{X}$ must be able to correspond to both continuous or discontinuous fields over the domain of interest (see Figure \ref{fig:idea_explain}) for both the forward and the inverse problem.   \par

\begin{figure}[!ht]
   \centering

         \includegraphics[keepaspectratio, height=10.0cm]{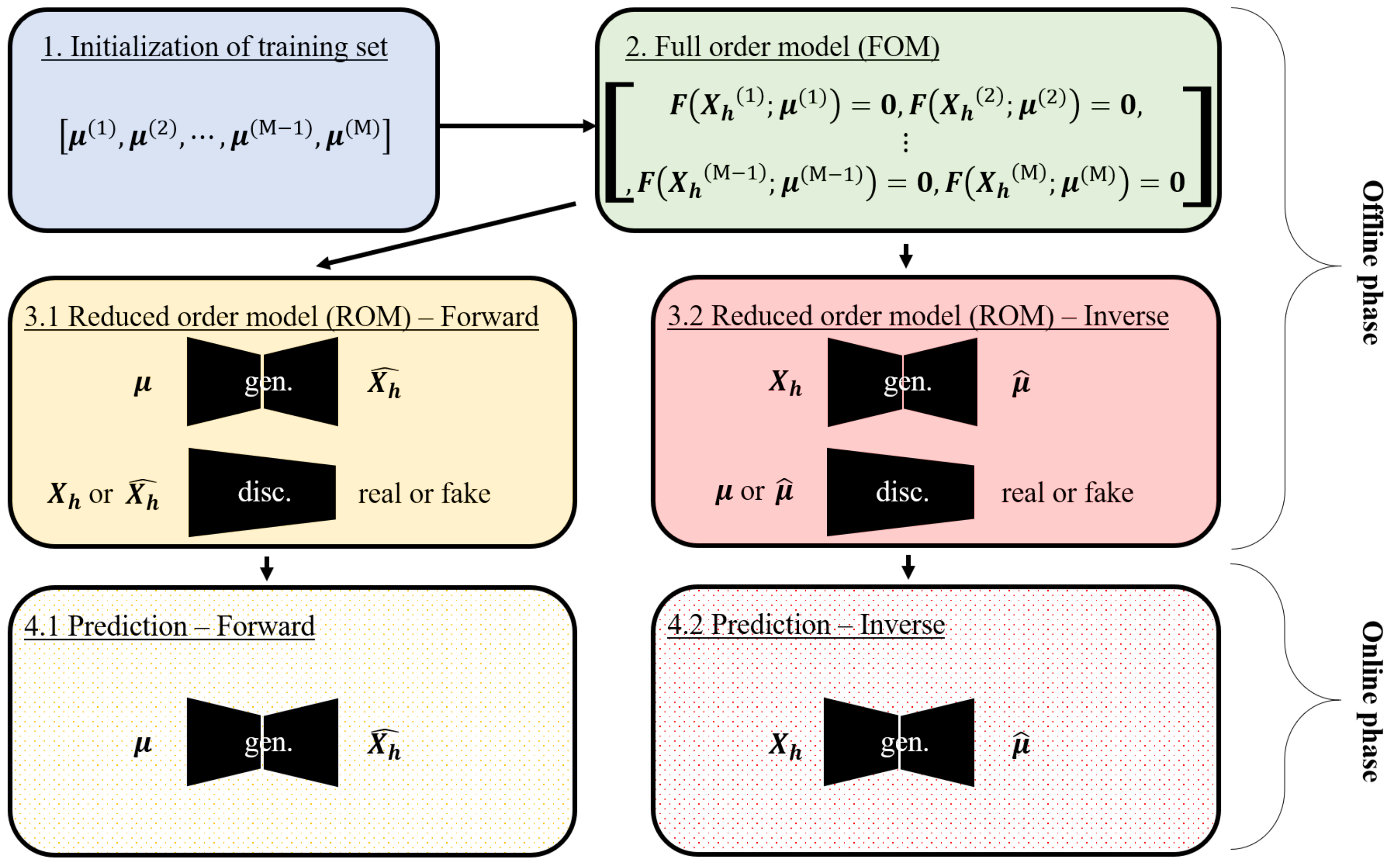}

   \caption{Summary of data-driven model reduction framework of PDEs. Gen. represents generator, and disc is discriminator. We note that $\bm{X}_h$ is an approximation of $\bm{X}$ obtaining from FOM, and $\widehat{\bm{X}_h}$ is an approximation of $\bm{X}_h$ obtaining from ROM in forward modeling setting. $\widehat{\bm{\mu}}$ is an estimation of $\bm{\mu}$ using ROM in inverse modeling setting.}
   \label{fig:pod_nn_explain}
   \end{figure}

A summary of the proposed framework built upon the offline-online paradigm is presented in Figure \ref{fig:pod_nn_explain}. The first step of the offline stage represents the initialization of a training set of parameters ($\bm{\mu}$) of cardinality $\mathrm{M}$ (colored in blue in Figure \ref{fig:pod_nn_explain}). These coefficients $\bm{\mu}$ could represent physical properties, geometric characteristics, or boundary conditions. However, in this work, we focus on using $\bm{\mu}$ to represent collections of spatially heterogeneous scalar coefficients, defining the physical characteristics of complex microstructures. \par

In the second step (colored in green in Figure \ref{fig:pod_nn_explain}), we query the FOM, which can provide a solution in a finite-dimensional setting, for each parameter $\bm{\mu}$ in the training set. The FOM is used to approximate primary variables $\bm{X}_h$, which could correspond to material displacement and fluid pressure fields given the field of parameters $\bm{\mu}$ as input. We note that $\bm{X}_h$ is a finite-dimensional approximation of $\bm{X}$ corresponding to the FOM results, which our model takes as the ground truth. Even though the proposed framework could be applied to a wide range of PDEs, here, we will focus on the steady-state solution of the linear poroelasticity equations coupling solid deformation and fluid diffusion in porous media with highly heterogeneous permeability. For the FOM, we use the Discontinuous Galerkin finite element solver, discussed in \ref{sec:governing_equations}, as required to handle the high degree of heterogeneity of the system. \par

In the third step, we build the ROM based on the cGAN image-to-image translation framework, which is discussed in detail in \ref{sec:cgan}. We compare the performance of three different variants of the proposed ROM (modifying the architecture and loss function) with respect to both accuracy and computational costs. These variants of the model are extensively discussed  in \ref{sec:cgan_base}, \ref{sec:cgan_sn}, and \ref{sec:cgan_w}. The training of the ROM must be adapted depending on whether it is to be used in forward or inverse modeling. For the forward setting (colored in yellow in Figure \ref{fig:pod_nn_explain}), the input to the generator is $\bm{\mu}$, and the output is $\widehat{\bm{X}_h}$. The input to the discriminator is $\bm{X}_h$ or $\widehat{\bm{X}_h}$, and the output quantifies the proximity of the generated data to the real data. For the inverse modeling (colored in red in Figure \ref{fig:pod_nn_explain}), the input to the generator is $\bm{X}_h$, and the output is $\widehat{\bm{\mu}}$. The input to the discriminator is $\bm{X}_h$ or $\widehat{\bm{\mu}}$, and the output again quantifies the proximity of the generated data to the real data.  \par


Finally, during the online phase for the forward modeling (colored in dotted yellow in Figure \ref{fig:pod_nn_explain}), we used the trained generator to predict $\widehat{\bm{X}_h}$ with given $\bm{\mu}$. For the inverse modeling (colored in dotted red in Figure \ref{fig:pod_nn_explain}), the trained generator is used to estimate $\widehat{\bm{\mu}}$ with given $\bm{X}_h$. In practice, the input field for the inverse model might be incomplete because the measurement points are limited and are located sparsely. Moreover, the input might also be contaminated with noise due to uncertainty in experimental measurement. Hence, the ROM in the inverse setting is tested with incomplete and noisy data. 

\section{Acknowledgments}

DO acknowledges support from Los Alamos National Laboratory's Laboratory Directed Research and Development Early Career Award (20200575ECR). HV is grateful to the funding support from the U.S. Department of Energy (DOE) Basic Energy Sciences (LANLE3W1).
NB acknowledges startup funds from Cornell University. YC acknowledges LDRD funds (21-FS-042) from Lawrence Livermore National Laboratory. Lawrence Livermore National Laboratory is operated by Lawrence Livermore National Security, LLC, for the U.S. Department of Energy, National Nuclear Security Administration under Contract DE-AC52-07NA27344 (LLNL-JRNL-823007).

\section{CRediT authorship contribution statement}

\textbf{T. Kadeethum}: Conceptualization, Formal analysis, Software, Validation, Writing - original draft, Writing - review \& editing. \textbf{D. O'Malley}: Conceptualization, Formal analysis, Supervision, Validation, Writing - review \& editing. \textbf{J.N. Fuhg}: Software, Validation, Writing - review \& editing. \textbf{Y. Choi}: Conceptualization, Formal analysis, Supervision, Validation, Writing - review \& editing. \textbf{J. Lee}: Software, Formal analysis, Supervision, Writing - review \& editing. \textbf{H.S. Viswanathan}: Conceptualization, Supervision, Writing - review \& editing. \textbf{N. Bouklas}: Conceptualization, Formal analysis, Funding acquisition, Supervision, Writing - review \& editing.

\section{Competing interests}
The authors declare no competing interests






\newpage

\beginsupplement
\import{}{supplement_arxiv.tex}

\bibliographystyle{unsrt}

\bibliography{references.bib}

\end{document}

%% file: layers/init.tex
\usetikzlibrary{quotes,arrows.meta}
\usetikzlibrary{positioning}

\def\edgecolor{rgb:blue,4;red,1;green,4;black,3}
\newcommand{\midarrow}{\tikz \draw[-Stealth,line width =0.8mm,draw=\edgecolor] (-0.3,0) -- ++(0.3,0);}

\usepackage{Ball}
\usepackage{Box}
\usepackage{RightBandedBox}

%% file: supplement_arxiv.tex
\section{Supplementary information}
\beginsupplementsub
\section{Coupled hydro-mechanical processes in porous media}\label{sec:governing_equations}

Even though the reduced order model presented in this manuscript (Section \ref{sec:cgan}) could be applied to any types of differential equations, we here focus on coupled hydro-mechanical (HM) processes in porous media, in which fluid flow and solid deformation tightly interact. These coupled multiphysics processes are involved in various problems ranging from groundwater and contaminant hydrology to biomedical engineering \cite{bisdom2016geometrically,juanes2016were,lee2016pressure,nick2013reactive,choo2018cracking,kadeethum2020enriched,yu2020poroelastic,kadeethum2019investigation,bouklas2015nonlinear,salimzadeh2019effect}. We follow the Biot's formulation for linear poroelasticity \cite{biot1941general,biot1957elastic}. Note that although linear poroelasticity theory may oversimplify deformations in soft porous materials such as soils \cite{Choo2016,Borja2016,Macminn2016,Zhao2020}, this description is reasonably good for stiff materials such as rocks, which are the focus of this work. The theory of linear poroelasticity theory describes the HM problem through two coupled governing equations, namely linear momentum and mass balance equations.


Let $\Omega \subset \mathbb{R}^d$ ($d \in \{1,2,3\}$) denote the physical domain and $\partial \Omega$ denote its boundary. The time domain is denoted by $\mathbb{T} = \left(0,\mathrm{T}\right]$ with $\mathrm{T}>0$. Primary variables used in this paper are $p : \Omega \times  \mathbb{T} \to \mathbb{R}$, which is a scalar-valued fluid pressure (\si{Pa}) and $\bm{u} : \Omega \times \mathbb{T} \to \mathbb{R}^d$, which is a vector-valued displacement (\si{m}). The infinitesimal strain tensor $\bm{\varepsilon}$ is defined as
\begin{equation}
\bm{\varepsilon}(\bm{u}) :=\frac{1}{2}\left(\nabla \bm{u}+(\nabla \bm{u})^{\intercal}\right),
\end{equation}

Further, $\bm{\sigma}$ is the total Cauchy stress tensor, which may be related to the effective Cauchy stress tensor $\bm{\sigma}^{\prime}$ and the pore pressure $p$ as
\begin{equation}
\bm{\sigma} (\bm{u},p) = \bm{\sigma}^{\prime}(\bm{u}) - \alpha p \mathbf{I}.
\end{equation}
Here, $\mathbf{I}$ is the second-order identity tensor, and
$\alpha$ is the Biot coefficient defined as \cite{jaeger2009fundamentals}:
\begin{equation} \label{eq:biot_coeff}
\alpha = 1-\frac{K}{K_{{s}}},
\end{equation}
\noindent
with {$K$} and {$K_s$} being the bulk moduli of the bulk porous material and the solid matrix, respectively.
According to linear elasticity, the effective stress tensor relates to the infinitesimal strain tensor, and therefore to the displacement, through the following constitutive relationship, which can be written as
\begin{equation}\label{eq:constitutive}
\bm{\sigma}^{\prime}(\bm{u}) =
\lambda_{l}  \tr(\bm{\varepsilon}(\bm{u})) \mathbf{I}
+ 2 \mu_{l} \bm{\varepsilon}{(\bm{u})}.
\end{equation} %
\noindent
where $\lambda_{l}$ and $\mu_{l}$ are the Lam\'e constants, which are related to the bulk modulus $K$ and the Poisson ratio $\nu$ of the porous solid as

\begin{equation}\label{eq:lambda_l}
\lambda_{l}=\frac{3 K \nu}{1+\nu}, \quad\text{ and }\quad \mu_{l}=\frac{3 K(1-2 \nu)}{2(1+\nu)}.
\end{equation}
\noindent

Under quasi-static conditions, the linear momentum balance equation can be written as
\begin{equation}
\nabla \cdot \bm{\sigma} (\bm{u},p) + \bm{f} = \bm{0},
\end{equation}
where $\bm{f}$ is the body force term defined as $\rho \phi \mathbf{g}+\rho_{s}(1-\phi) \mathbf{g}$, where $\rho$ is the fluid density, $\rho_s$ is the solid density, $\phi$ is the porosity, and $\mathbf{g}$ is the gravitational acceleration vector.
The gravitational force will be neglected in this study, but the body force term will be kept in the succeeding formulations for a more general case.

For this solid deformation problem, the domain boundary $\partial \Omega$ is assumed to be suitably decomposed into displacement and traction boundaries, $\partial \Omega_u$ and $\partial \Omega_{t}$, respectively.
Then the linear momentum balance equation is supplemented by the boundary and initial conditions as:
%
\begin{equation} \label{eq:linear_balance}
\begin{split}
\nabla \cdot \bm{\sigma}^{\prime}(\bm{u}) -\alpha \nabla \cdot \left(p \mathbf{I}\right)
+ \bm{f} = \bm{0}  &\text { \: in \: } \Omega \times \mathbb{T}, \\
\bm{u} =\bm{u}_{D} &\text { \: on \: } \partial \Omega_{u} \times \mathbb{T},\\
\bm{\sigma} {(\bm{u})} \cdot  \mathbf{n}=\bm{t_D} &\text { \: on \: } \partial \Omega_{t} \times \mathbb{T}, \\
\bm{u}=\bm{u}_{0}  &\text { \: in \: } \Omega \text { at } t = 0,
\end{split}
\end{equation}

\noindent
where $\bm{u}_D$ and ${\bm{t}_D}$ are prescribed displacement and traction values at the boundaries, respectively, and $\mathbf{n}$ is the unit normal vector to the boundary.

Next, the mass balance equation is given as \cite{coussy2004poromechanics,kim2011stability}:
\begin{equation} \label{eq:mass_balance_old}
\frac{1}{M} \dfrac{\partial p}{\partial t} +
\alpha  \frac{\partial {\varepsilon_{v}}}{\partial t}
+ \nabla \cdot  \bm{q} = g \text { in } \Omega \times \mathbb{T},
\end{equation}

\noindent
where
\begin{equation} \label{eq:1/M}
\frac{1}{M}  = \left(\phi c_{f}+\dfrac{\alpha-\phi}{K_{s}}\right)
\end{equation}
is the Biot modulus.
\noindent
Here, $c_f$ is the fluid compressibility, ${\varepsilon_{v}}$ := $\operatorname{tr}(\bm{\varepsilon}) = \nabla \cdot \bm{u}$ is the volumetric strain, and $g$ is a sink/source term. The superficial velocity vector $\bm{q}$ is given by Darcy's law as
\begin{equation} \label{eq:darcy}
\bm{q} =- \bm{\kappa}(\nabla p-\rho \mathbf{g}).
\end{equation}

\noindent
Here $\bm{\kappa}=\frac{\bm{k}}{\mu_f}$ is the porous media conductivity, $\mu_f$ is the fluid viscosity. Again, the gravitational force, $\rho \mathbf{g}$, will be neglected in this work, without loss of generality. In addition, $\bm{k}$ is the matrix permeability tensor defined as

\begin{equation} \label{eq:permeability_matrix}
\bm{k} :=
\begin{cases}
 \left[ \begin{array}{lll}{{k}_{xx}} & {{k}_{xy}} & {{k}_{xz}} \\ {{k}_{yx}} & {{k}_{yy}} & {{k}_{yz}} \\ {{k}_{zx}} & {{k}_{zy}} & {k}_{zz}\end{array}\right] & \text{if} \ d = 3, \ \\ \\
 \left[ \begin{array}{ll}{{k}_{xx}} & {{k}_{xy}}  \\ {{k}_{yx}} & {{k}_{yy}} \\ \end{array}\right]  & \text{if} \ d = 2, \ \\ \\
 \ {k}_{xx}  & \text{if} \ d = 1,
\end{cases}
\end{equation}

\noindent

To simplify our problem, we assume all off-diagonal terms of \eqref{eq:permeability_matrix} to be zero and all diagonal terms have similar value. For the fluid flow problem, the domain boundary $\partial \Omega$ is also suitably decomposed into the pressure and flux boundaries,  $\partial \Omega_p$ and $\partial \Omega_q$, respectively.
In what follows, we apply the fixed stress split scheme \cite{kim2011stability,mikelic2013convergence}, assuming $\left(\sigma_{v}-\sigma_{v, 0}\right)+\alpha \left(p-p_{0}\right)=K \varepsilon_{v}$.
Then we write the fluid flow problem with boundary and initial conditions as

\begin{equation} \label{eq:mass_balance}
\begin{split}
\left(\frac{1}{M}+\frac{\alpha^{2}}{K}\right) \frac{\partial p}{\partial t}+\frac{\alpha}{K} \frac{\partial \sigma_{v}}{\partial t}- \bm{\kappa} \nabla p = g  &\text { \: in \: } \Omega \times \mathbb{T}, \\
p=p_{D} &\text { \: on \: } \partial \Omega_{p} \times \mathbb{T}, \\
- \bm{\kappa} \nabla p \cdot \mathbf{n}=q_{D} &\text { \: on \:} \partial \Omega_{q} \times \mathbb{T}, \\
p=p_{0} &\text { \: in \: } \Omega \text { at } t = 0,
\end{split}
\end{equation}

\noindent
where $\sigma_{v}:=\frac{1}{3} \tr(\bm{\sigma})$ is the volumetric stress, and $p_D$ and $q_D$ are the given boundary pressure and flux, respectively.

\begin{figure}[!ht]
   \centering
    \includegraphics[width=6.0cm,keepaspectratio]{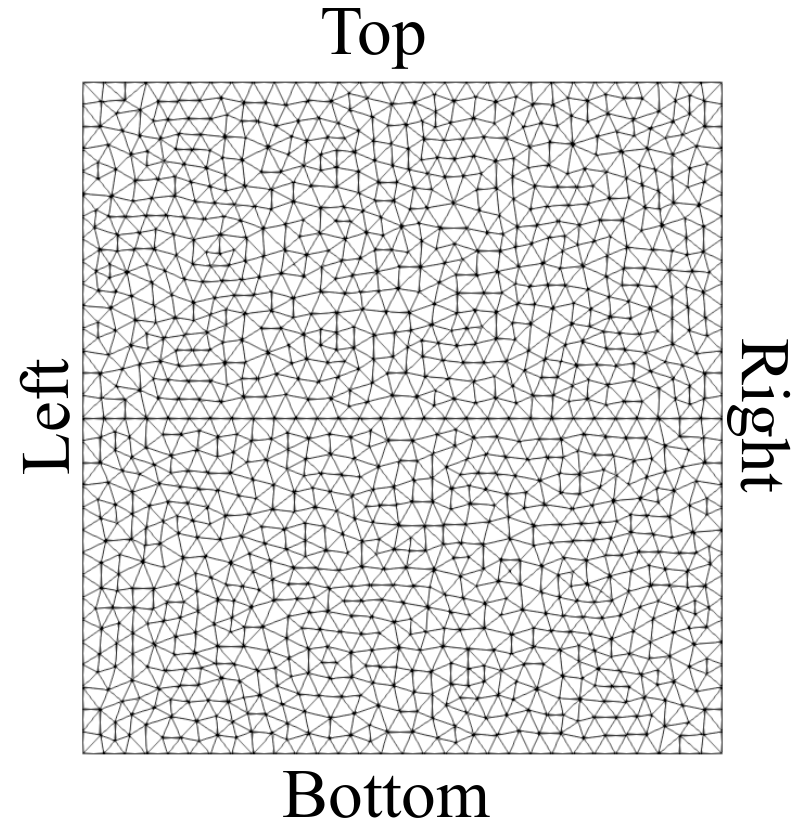} 
   \caption{Domain, its boundaries, and mesh used for all numerical examples.}
   \label{fig:mesh}
\end{figure}

We utilize the discontinuous Galerkin finite element model of linear poroelasticity developed in \cite{Kadeethum2019ARMA, kadeethum2020enriched, kadeethum2020finite} for this study. To simplify the process, we only investigate the pressure and displacement fields at the steady-state solutions and use these solutions to train our ROM model. The mesh and boundary conditions adapted from \cite{kadeethum2021non} are presented in Fig. \ref{fig:mesh} . We take $\Omega = \left(0, 1\right)^2$ corresponding to a square domain of $1\mathrm{m}^2$ area, and decompose its boundary $\partial \Omega$ with the following labels

\begin{equation}
\begin{aligned}
\mathrm{Left} & = \{0\} \times [0,1],  \\
\mathrm{Top} &= [0,1] \times \{1\},   \\
\mathrm{Right} &= \{1\} \times[0,1],   \\
\mathrm{Bottom} &= [0,1] \times \{0\}.
\end{aligned}
\end{equation}

\noindent
Throughout this study, the boundary conditions are described as follows

\begin{equation} \label{eq:ex1_bound_u}
\begin{split}
\bm{u}_{D} \cdot  \mathbf{n}= 0 \quad \si{m} &\text { \: on \: } \mathrm{Left} \times \mathbb{T},\\
\bm{t_D} =[0, -1]  \quad  \si{kPa} &\text { \: on \: } \mathrm{Top} \times \mathbb{T}, \\
\bm{u}_{D} \cdot  \mathbf{n}= 0 \quad  \si{m} &\text { \: on \: } \mathrm{Right} \times \mathbb{T},\\
\bm{u}_{D} \cdot  \mathbf{n}= 0 \quad \si{m} &\text { \: on \: } \mathrm{Bottom} \times \mathbb{T},
\end{split}
\end{equation}

\noindent
for \eqref{eq:linear_balance}, and

\begin{equation} \label{eq:ex1_bound_p}
\begin{split}
{q}_{D} = 0 \quad \si{m/s} &\text { \: on \: } \mathrm{Left} \times \mathbb{T},\\
{p_D} = 0  \quad  \si{Pa} &\text { \: on \: } \mathrm{Top} \times \mathbb{T}, \\
{q}_{D} = 0 \quad \si{m/s} &\text { \: on \: } \mathrm{Right} \times \mathbb{T},\\
{p_D} = 1000 \quad \si{Pa} &\text { \: on \: } \mathrm{Bottom} \times \mathbb{T},
\end{split}
\end{equation}

\noindent
for \eqref{eq:mass_balance}. The degrees of freedom associated with this mesh are $9722$ for the continuous approximation of the displacement field $\bm{u}$, and $7110$ for the discontinuous approximation of the pressure $p$.

The following set of input parameters are used throughout this study. We fix $\alpha \approx 1$, as the porous matrix is characterized by $K=1000$ $\mathrm{kPa}$ while the bulk solid is modeled by $K_{\mathrm{s}} \to \infty$ $\mathrm{kPa}$, $c_f = 1.0 \times 10^{-10}$ $\mathrm{Pa^{-1}}$, $\phi = 0.2$, fluid viscosity - $\mu_f=10^{-3}$ $\mathrm{Pa.s}$, and Poisson ratio $\nu = 0.25$. The permeability in x-direction (${k}_{xx}$) is uniquely populated for each simulation by \cite{muller2020}. Then the permeability field ($\bm{k}$) is set as

\begin{equation} \label{eq:permeability_used}
    \bm{k} := \left[ \begin{array}{ll}{{k}_{xx}} & 0.0  \\ 0.0 & {{k}_{xx}} \\ \end{array}\right]
\end{equation}

As porous media are generally heterogeneous, and the sharp contrast of phases of the porous microstructure leads to discontinuous material properties that can span several orders of magnitude, it is very challenging and time-consuming to capture multiphysics processes occurred in these media using finite volume or finite element methods \cite{matthai2009upscaling,kadeethum2020finite,baker2015practical,jia2017comprehensive,kadeethum2018investigation,bisdom2016geometrically,muljadi2016impact,nicolaides2015impact,KadNickLeeBallarin_2019_mixed}. Hence, we aim to reduce this problem's computational cost while maintaining the acceptable accuracy through a generative model presented in Section \ref{sec:cgan}.

\begin{remark}
 We want to emphasize that we use the discontinuous Galerkin finite element model of linear poroelasticity developed in \cite{Kadeethum2019ARMA, kadeethum2020enriched, kadeethum2020finite, kadeethum2021non} to generate training and test examples. Each FEM simulation uses the same set of input parameters except the permeability field ($\bm{k}$), which is uniquely generated for each simulation by \cite{muller2020}. Besides, to simplify the process, we only employ the steady-state solutions in this study. Time-dependent problems will be incorporated in our subsequent studies. 
\end{remark}

\beginsupplementsub
\section{Conditional generative adversarial network}\label{sec:cgan}

We propose a data-driven model order reduction for physics-based problems using conditional generative adversarial network (cGAN) \cite{shen2020interpreting,mirza2014conditional,chen2016infogan}. The framework is adapted from the idea of paired image-to-image translation developed in \cite{isola2017image,ma2017pose}. The word `paired' reflects the fact that there is a one-to-one mapping between input fields and output fields. In the past several years, this technique and its variations have been applied to many problems both supervised \cite{wang2018high,ledig2017photo,xu2018attngan} and unsupervised learning \cite{zhu2017unpaired,liu2017unsupervised,huang2018multimodal}. \par

In contrast to the classical cGAN in which noise and class vectors are used as an input to the generator \cite{mirza2014conditional}, the image-to-image translation framework employs a conditional matrix (or a conditional field) as its input \cite{isola2017image}. This conditional field could be edges, masks, points, or segmentation maps in typical image analysis studies. These fields then are mapped to the output, which is an image. \cite{ma2017pose, wang2018high,ledig2017photo,xu2018attngan,zhu2017unpaired,liu2017unsupervised,huang2018multimodal}.  This study, however, uses physics fields that could represent material properties, boundary and initial conditions, and measurable quantities such as pressure, temperature, or displacements as the generator's input and output fields. \par

\begin{figure}
    \centering
    \scalebox{0.5}{\begin{tikzpicture}
\tikzstyle{connection}=[ultra thick,every node/.style={sloped,allow upside down},draw=\edgecolor,opacity=0.7]
\tikzstyle{copyconnection}=[ultra thick,every node/.style={sloped,allow upside down},draw={rgb:blue,4;red,1;green,1;black,3},opacity=0.7]

\node[canvas is zy plane at x=0] (temp) at (-1,0,0) {\includegraphics[width=10cm,height=10cm]{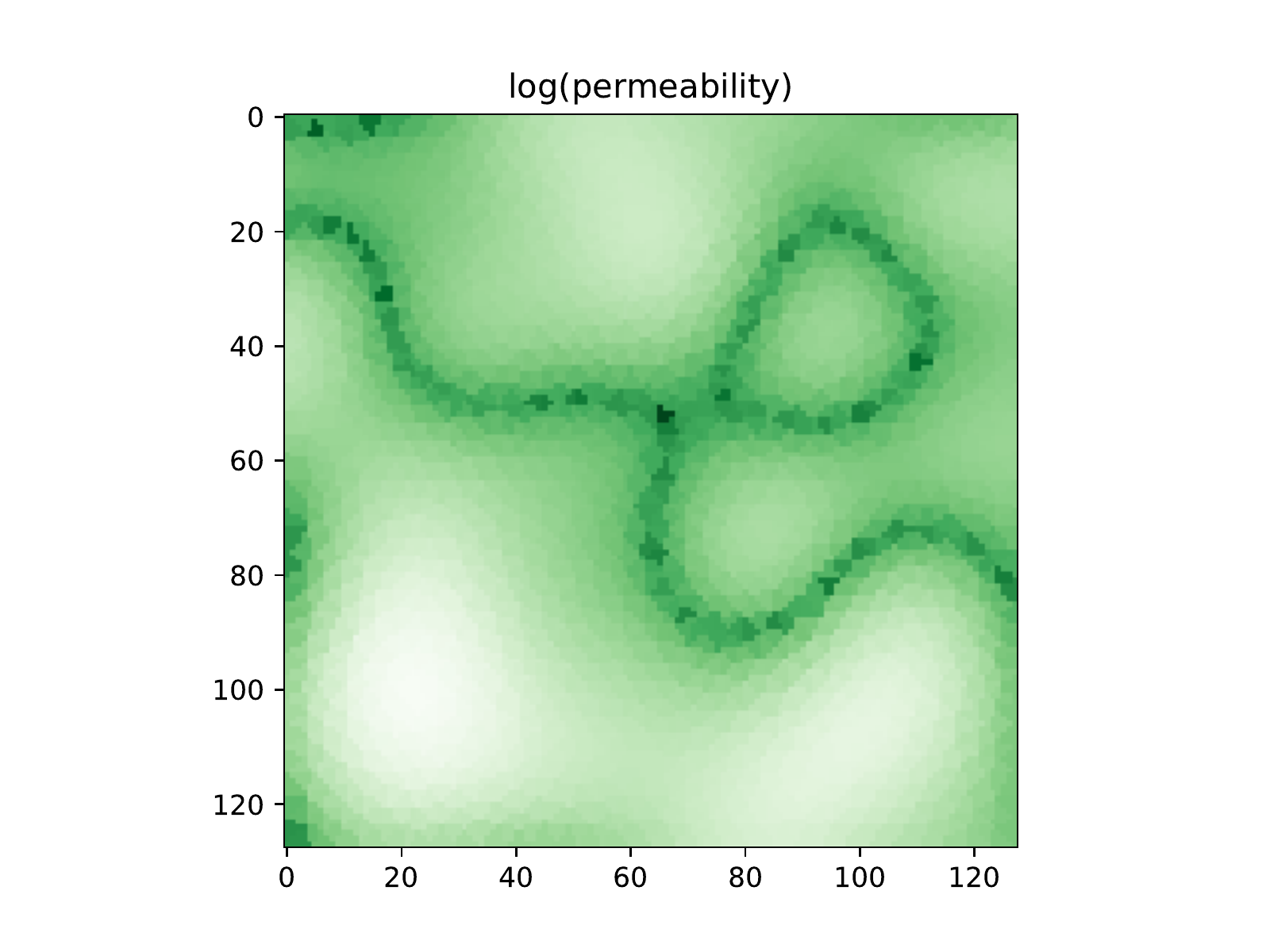}};

\pic[shift={ (0,0,0) }] at (0,0,0) 
    {RightBandedBox={
        name=ccr_b1,
        caption= ,
        fill=\ConvColor,
        bandfill=\ConvReluColor,
        height=40,
        width={ 2 , 2 },
        depth=40
        }
    };

\pic[shift={ (0,0,0) }] at (ccr_b1-east) 
    {Box={
        name=pool_b1,
        caption= ,
        fill=\PoolColor,
        opacity=0.5,
        height=32,
        width=1,
        depth=32
        }
    };

\pic[shift={ (1,0,0) }] at (pool_b1-east) 
    {RightBandedBox={
        name=ccr_b2,
        caption= ,
        fill=\ConvColor,
        bandfill=\ConvReluColor,
        height=32,
        width={ 3.5 , 3.5 },
        depth=32
        }
    };

\pic[shift={ (0,0,0) }] at (ccr_b2-east) 
    {Box={
        name=pool_b2,
        caption= ,
        fill=\PoolColor,
        opacity=0.5,
        height=24,
        width=1,
        depth=24
        }
    };

\draw [connection]  (pool_b1-east)    -- node {\midarrow} (ccr_b2-west);

\pic[shift={ (1,0,0) }] at (pool_b2-east) 
    {RightBandedBox={
        name=ccr_b3,
        caption= ,
        fill=\ConvColor,
        bandfill=\ConvReluColor,
        height=25,
        width={ 4.5 , 4.5 },
        depth=25
        }
    };

\pic[shift={ (0,0,0) }] at (ccr_b3-east) 
    {Box={
        name=pool_b3,
        caption= ,
        fill=\PoolColor,
        opacity=0.5,
        height=19,
        width=1,
        depth=19
        }
    };

\draw [connection]  (pool_b2-east)    -- node {\midarrow} (ccr_b3-west);




\pic[shift={ (2,0,0) }] at (pool_b3-east) 
    {RightBandedBox={
        name=ccr_b5,
        caption=Bottleneck,
        fill=\ConvColor,
        bandfill=\ConvReluColor,
        height=8,
        width={ 8 , 8 },
        depth=8
        }
    };

\draw [connection]  (pool_b3-east)    -- node {\midarrow} (ccr_b5-west);

\pic[shift={ (2.1,0,0) }] at (ccr_b5-east) 
    {Box={
        name=unpool_b7,
        caption= ,
        fill=\UnpoolColor,
        opacity=0.5,
        height=25,
        width=1,
        depth=25
        }
    };

\pic[shift={ (0,0,0) }] at (unpool_b7-east) 
    {RightBandedBox={
        name=ccr_res_b7,
        caption= ,
        fill={rgb:white,1;black,3},
        bandfill={rgb:white,1;black,2},
        opacity=0.5,
        height=25,
        width=4.5,
        depth=25
        }
    };

\pic[shift={(0,0,0)}] at (ccr_res_b7-east) 
    {Box={
        name=ccr_b7,
        caption= ,
        fill=\ConvColor,
        height=25,
        width=4.5,
        depth=25
        }
    };

\pic[shift={ (0,0,0) }] at (ccr_b7-east) 
    {RightBandedBox={
        name=ccr_res_c_b7,
        caption= ,
        fill={rgb:white,1;black,3},
        bandfill={rgb:white,1;black,2},
        opacity=0.5,
        height=25,
        width=4.5,
        depth=25
        }
    };

\pic[shift={(0,0,0)}] at (ccr_res_c_b7-east) 
    {Box={
        name=end_b7,
        caption= ,
        fill=\ConvColor,
        height=25,
        width=4.5,
        depth=25
        }
    };

\draw [connection]  (ccr_b5-east)    -- node {\midarrow} (unpool_b7-west);

\path (ccr_b3-southeast) -- (ccr_b3-northeast) coordinate[pos=1.25] (ccr_b3-top) ;
\path (ccr_res_b7-south)  -- (ccr_res_b7-north)  coordinate[pos=1.25] (ccr_res_b7-top) ;
\draw [copyconnection]  (ccr_b3-northeast)  
-- node {\copymidarrow}(ccr_b3-top)
-- node {\copymidarrow}(ccr_res_b7-top)
-- node {\copymidarrow} (ccr_res_b7-north);

\pic[shift={ (2.1,0,0) }] at (end_b7-east) 
    {Box={
        name=unpool_b8,
        caption= ,
        fill=\UnpoolColor,
        opacity=0.5,
        height=32,
        width=1,
        depth=32
        }
    };

\pic[shift={ (0,0,0) }] at (unpool_b8-east) 
    {RightBandedBox={
        name=ccr_res_b8,
        caption= ,
        fill={rgb:white,1;black,3},
        bandfill={rgb:white,1;black,2},
        opacity=0.5,
        height=32,
        width=3.5,
        depth=32
        }
    };

\pic[shift={(0,0,0)}] at (ccr_res_b8-east) 
    {Box={
        name=ccr_b8,
        caption= ,
        fill=\ConvColor,
        height=32,
        width=3.5,
        depth=32
        }
    };

\pic[shift={ (0,0,0) }] at (ccr_b8-east) 
    {RightBandedBox={
        name=ccr_res_c_b8,
        caption= ,
        fill={rgb:white,1;black,3},
        bandfill={rgb:white,1;black,2},
        opacity=0.5,
        height=32,
        width=3.5,
        depth=32
        }
    };

\pic[shift={(0,0,0)}] at (ccr_res_c_b8-east) 
    {Box={
        name=end_b8,
        caption= ,
        fill=\ConvColor,
        height=32,
        width=3.5,
        depth=32
        }
    };

\draw [connection]  (end_b7-east)    -- node {\midarrow} (unpool_b8-west);

\path (ccr_b2-southeast) -- (ccr_b2-northeast) coordinate[pos=1.25] (ccr_b2-top) ;
\path (ccr_res_b8-south)  -- (ccr_res_b8-north)  coordinate[pos=1.25] (ccr_res_b8-top) ;
\draw [copyconnection]  (ccr_b2-northeast)  
-- node {\copymidarrow}(ccr_b2-top)
-- node {\copymidarrow}(ccr_res_b8-top)
-- node {\copymidarrow} (ccr_res_b8-north);

\pic[shift={ (2.1,0,0) }] at (end_b8-east) 
    {Box={
        name=unpool_b9,
        caption= ,
        fill=\UnpoolColor,
        opacity=0.5,
        height=40,
        width=1,
        depth=40
        }
    };

\pic[shift={ (0,0,0) }] at (unpool_b9-east) 
    {RightBandedBox={
        name=ccr_res_b9,
        caption= ,
        fill={rgb:white,1;black,3},
        bandfill={rgb:white,1;black,2},
        opacity=0.5,
        height=40,
        width=2.5,
        depth=40
        }
    };

\pic[shift={(0,0,0)}] at (ccr_res_b9-east) 
    {Box={
        name=ccr_b9,
        caption= ,
        fill=\ConvColor,
        height=40,
        width=2.5,
        depth=40
        }
    };

\pic[shift={ (0,0,0) }] at (ccr_b9-east) 
    {RightBandedBox={
        name=ccr_res_c_b9,
        caption= ,
        fill={rgb:white,1;black,3},
        bandfill={rgb:white,1;black,2},
        opacity=0.5,
        height=40,
        width=2.5,
        depth=40
        }
    };

\pic[shift={(0,0,0)}] at (ccr_res_c_b9-east) 
    {Box={
        name=end_b9,
        caption= ,
        fill=\ConvColor,
        height=40,
        width=2.5,
        depth=40
        }
    };

\draw [connection]  (end_b8-east)    -- node {\midarrow} (unpool_b9-west);

\path (ccr_b1-southeast) -- (ccr_b1-northeast) coordinate[pos=1.25] (ccr_b1-top) ;
\path (ccr_res_b9-south)  -- (ccr_res_b9-north)  coordinate[pos=1.25] (ccr_res_b9-top) ;
\draw [copyconnection]  (ccr_b1-northeast)  
-- node {\copymidarrow}(ccr_b1-top)
-- node {\copymidarrow}(ccr_res_b9-top)
-- node {\copymidarrow} (ccr_res_b9-north);

\node[canvas is zy plane at x=0] (temp) at (28,0,0) {\includegraphics[width=10cm,height=10cm]{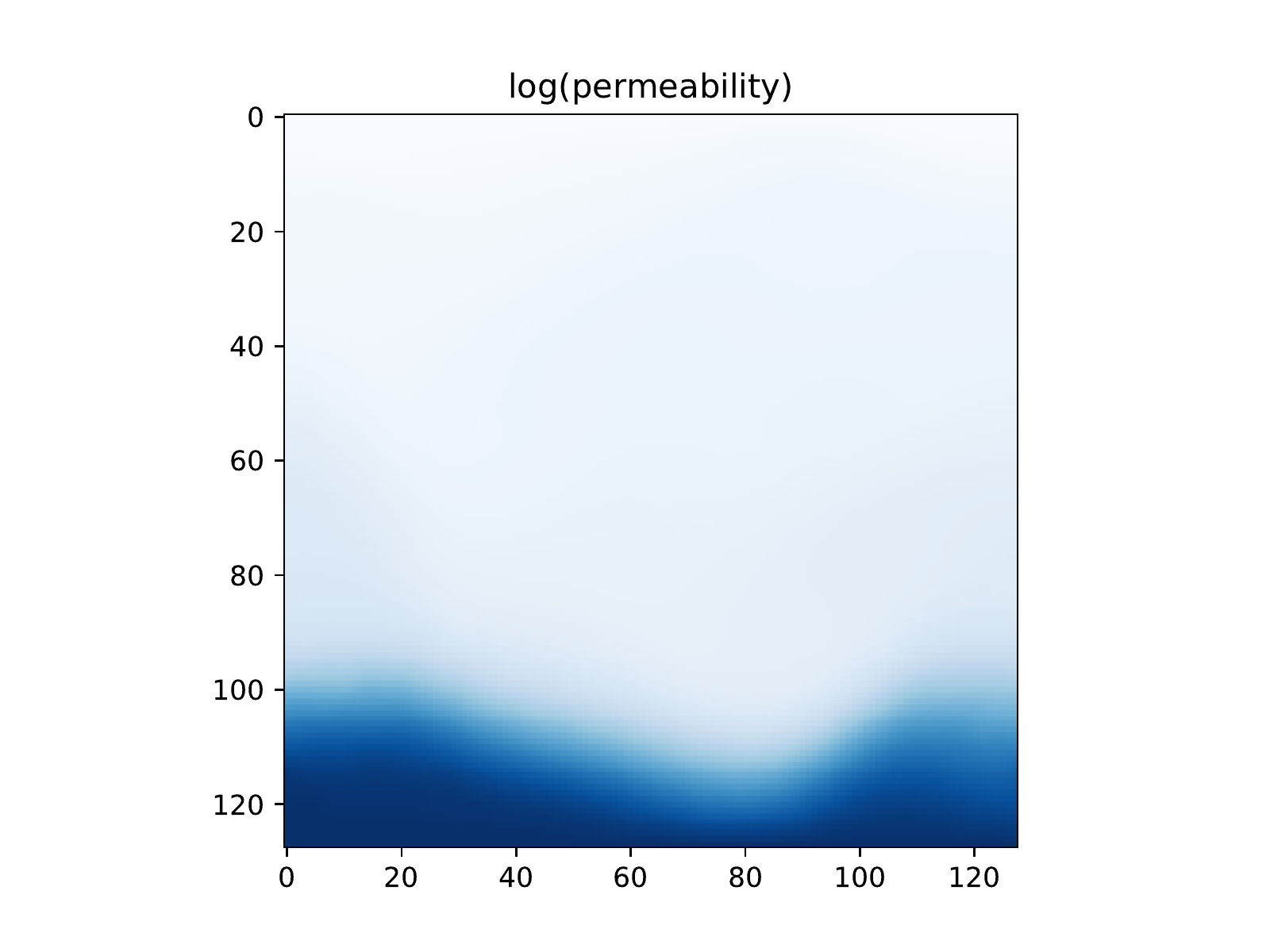}};



\end{tikzpicture}}
    \caption{Generator: U-net visualized using PlotNeuralNet: \url{https://github.com/HarisIqbal88/PlotNeuralNet}}
    \label{fig:unet}
\end{figure}

The generator used in this framework is illustrated in Fig. \ref{fig:unet}, and it resembled the well-established architecture of U-net, which typically is used for image segmentation \cite{ronneberger2015u}. The first component of the generator is a contracting block that performs two convolutions followed by a max pool operation (see Fig. \ref{fig:unet} - light and dark orange blocks). The second component is an expanding block (see Fig. \ref{fig:unet} - blue and pale yellow) in which it performs an upsampling, a convolution, a concatenation of its two inputs. The generator used in this study consists of six contracting and six expanding blocks. \par 

We provide a detailed architecture of the generator used in this study in Table \ref{tab:unet}. Note that each contracting block uses LeakyReLU with a negative slope of 0.2 as its activation function, each expanding block uses ReLU as its activation function, the $1^{\mathrm{st}}$ convolutional layer is used to map input channel ($\mathbb{C}_{\mathrm{in}}$) to hidden layer size ($\mathbb{H}$), and it does not subject to any activation function, and the $2^{\mathrm{nd}}$ convolutional layer is used to map hidden layer size ($\mathbb{H}$) to output channel ($\mathbb{C}_{\mathrm{out}}$), and it subjects to the Sigmoid activation function. For the generator, $\mathbb{C}_{\mathrm{in}}$ $=$ $\mathbb{C}_{\mathrm{out}}$ $=$ $\mathbb{C}$, and $\mathbb{H} = 32$. The domain size is governed by $\mathbb{DOF_X}$ $\times$ $\mathbb{DOF_Y}$, which is 128 $\times$ 128 throughout this manuscript. We note that $\mathbb{B}$ is batch size. \par 

\begin{table}[!ht]
\centering
\caption{Generator: U-net's detail used in this study (input and output sizes are represented by {[}$\mathbb{B}$, $\mathbb{C}$, $\mathbb{DOF_X}$, $\mathbb{DOF_Y}${]}. We use hidden layers $\mathbb{H} = 32$)}
\begin{tabular}{|l|c|c|c|c|}
\hline
Block                  & \multicolumn{1}{l|}{Input size} & \multicolumn{1}{l|}{Output size} & \multicolumn{1}{l|}{Batch normalization} & \multicolumn{1}{l|}{Dropout} \\ \hline
$1^{\mathrm{st}}$ convolutional layer & {[}$\mathbb{B}$, $\mathbb{C}$, 128, 128{]}            & {[}$\mathbb{B}$, 32, 128, 128{]}            &                                          &                              \\ \hline
$1^{\mathrm{st}}$ contracting block   & {[}$\mathbb{B}$, 32, 128, 128{]}           & {[}$\mathbb{B}$, 64, 64, 64{]}              & \checkmark                                        & \checkmark                            \\ \hline
$2^{\mathrm{nd}}$ contracting block   & {[}$\mathbb{B}$, 64, 64, 64{]}             & {[}$\mathbb{B}$, 128, 32, 32{]}             & \checkmark                                        & \checkmark                            \\ \hline
$3^{\mathrm{rd}}$ contracting block   & {[}$\mathbb{B}$, 128, 32, 32{]}            & {[}$\mathbb{B}$, 256, 16, 16{]}             & \checkmark                                        & \checkmark                            \\ \hline
$4^{\mathrm{th}}$ contracting block   & {[}$\mathbb{B}$, 256, 16, 16{]}            & {[}$\mathbb{B}$, 512, 8, 8{]}               & \checkmark                                        &                              \\ \hline
$5^{\mathrm{th}}$ contracting block   & {[}$\mathbb{B}$, 512, 8, 8{]}              & {[}$\mathbb{B}$, 1024, 4, 4{]}              & \checkmark                                        &                              \\ \hline
$6^{\mathrm{th}}$ contracting block   & {[}$\mathbb{B}$, 1024, 4, 4{]}             & {[}$\mathbb{B}$, 2048, 2, 2{]}              & \checkmark                                        &                              \\ \hline
$1^{\mathrm{st}}$ expanding block     & {[}$\mathbb{B}$, 2048, 2, 2{]}         & {[}$\mathbb{B}$, 1024, 4, 4{]}              & \checkmark                                        &                              \\ \hline
$2^{\mathrm{nd}}$ expanding block    & {[}$\mathbb{B}$, 1024, 4, 4{]}         & {[}$\mathbb{B}$, 512, 8, 8{]}               & \checkmark                                        &                              \\ \hline
$3^{\mathrm{rd}}$ expanding block     & {[}$\mathbb{B}$, 512, 8, 8{]}          & {[}$\mathbb{B}$, 256, 16, 16{]}             & \checkmark                                        &                              \\ \hline
$4^{\mathrm{th}}$ expanding block     & {[}$\mathbb{B}$, 256, 16, 16{]}      & {[}$\mathbb{B}$, 128, 32, 32{]}             & \checkmark                                        &                              \\ \hline
$5^{\mathrm{th}}$ expanding block     & {[}$\mathbb{B}$, 128, 32, 32{]}      & {[}$\mathbb{B}$, 64, 64, 64{]}              & \checkmark                                        &                              \\ \hline
$6^{\mathrm{th}}$ expanding block     & {[}$\mathbb{B}$, 64, 64, 64{]}       & {[}$\mathbb{B}$, 32, 128, 128{]}            & \checkmark                                        &                              \\ \hline
$2^{\mathrm{nd}}$ convolutional layer & {[}$\mathbb{B}$, 32, 128, 128{]}           & {[}$\mathbb{B}$, $\mathbb{C}$, 128, 128{]}             &                                          &                              \\ \hline
\end{tabular}
\label{tab:unet}
\end{table}

Both input or conditional ($\bm{I}$) and output or real ($\bm{O}$) fields of the generator are normalized to be in a range of $[0, 1]$ as follows:

\begin{equation} \label{eq:norm_input_output}
{I}_{i}=\frac{{I}_{i}-\min (\bm{I})}{\max (\bm{I})-\min (\bm{I})} \quad \text{and} \quad {O}_{i}=\frac{{O}_{i}-\min (\bm{O})}{\max (\bm{O})-\min (\bm{O})}.
\end{equation}

\noindent
where $i$ represents each member's index in the sets of $\bm{I}$ and $\bm{O}$. We also note that $\bm{I}$ and $\bm{O}$ have $\mathrm{M}$ members (as well as the approximated output or fake fields ($\widehat{\bm{O}}$)). As the last layer of the generator $2^{\mathrm{nd}}$ convolutional layer (see Table \ref{tab:unet}), is subjected to the Sigmoid activation function, the $\widehat{\bm{O}}$ is always in a range of 0 to 1.  \par

\begin{figure}
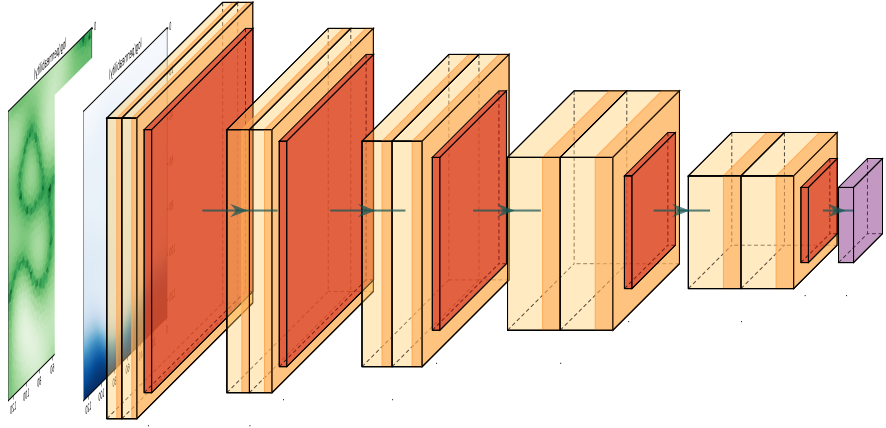

    \centering
     \scalebox{0.5}{\begin{tikzpicture}
\tikzstyle{connection}=[ultra thick,every node/.style={sloped,allow upside down},draw=\edgecolor,opacity=0.7]
\node[canvas is zy plane at x=0] (temp) at (-3,0,0) {\includegraphics[width=10cm,height=10cm]{pictures/perm_13_418.pdf}};
\node[canvas is zy plane at x=0] (temp) at (-1,0,0) {\includegraphics[width=10cm,height=10cm]{pictures/pressure_13_418.pdf}};
\pic[shift={(0,0,0)}] at (0,0,0) {RightBandedBox={name=cr1%
        ,fill=\ConvColor,bandfill=\ConvReluColor,%
        caption = ,height=40,width={2,2},depth=40}};

\pic[shift={(0,0,0)}] at (cr1-east) {Box={name=p1,%
        caption = ,fill=\PoolColor,opacity=0.5,height=35,width=1,depth=35}};
\pic[shift={(2,0,0)}] at (p1-east) {RightBandedBox={name=cr2,%
        caption = ,fill=\ConvColor,bandfill=\ConvReluColor,%
        height=35,width={3,3},depth=35}};
\pic[shift={(0,0,0)}] at (cr2-east) {Box={name=p2,%
        caption =,fill=\PoolColor,opacity=0.5,height=30,width=1,depth=30}};
\pic[shift={(2,0,0)}] at (p2-east) {RightBandedBox={name=cr3,%
        caption = ,fill=\ConvColor,bandfill=\ConvReluColor,%
        height=30,width={4,4},depth=30}};
\pic[shift={(0,0,0)}] at (cr3-east) {Box={name=p3,%
        caption = ,fill=\PoolColor,opacity=0.5,height=23,width=1,depth=23}};
\pic[shift={(1.8,0,0)}] at (p3-east) {RightBandedBox={name=cr4,%
        caption = ,fill=\ConvColor,bandfill=\ConvReluColor,%
        height=23,width={7,7},depth=23}};
\pic[shift={(0,0,0)}] at (cr4-east) {Box={name=p4,%
        caption =,fill=\PoolColor,opacity=0.5,height=15,width=1,depth=15}};
\pic[shift={(1.5,0,0)}] at (p4-east) {RightBandedBox={name=cr5,%
        caption = ,fill=\ConvColor,bandfill=\ConvReluColor,%
        height=15,width={7,7},depth=15}};
\pic[shift={(0,0,0)}] at (cr5-east) {Box={name=p5,%
        caption = ,fill=\PoolColor,opacity=0.5,height=10,width=1,depth=10}};

\pic[shift={(1,0,0)}] at (cr5-east) {Box={name=c8,%
    caption = ,fill=\SoftmaxColor,%
        height=10,width=2,depth=10}};
\draw [connection]  (p1-east)    -- node {\midarrow} (cr2-west);
\draw [connection]  (p2-east)    -- node {\midarrow} (cr3-west);
\draw [connection]  (p3-east)    -- node {\midarrow} (cr4-west);
\draw [connection]  (p4-east)    -- node {\midarrow} (cr5-west);
\draw [connection]  (p5-east)    -- node {\midarrow} (c8-west);
\end{tikzpicture}}
    \caption{Discriminator: PatchGAN visualized using PlotNeuralNet: \url{https://github.com/HarisIqbal88/PlotNeuralNet}}
    \label{fig:disc}
\end{figure}

In the classical cGAN, input to the discriminator is a concatenation of conditional vector (e.g., [0, 1, ..., 8, 9] for MNIST dataset) and output field produced from the generator or a real field (training/testing set) \cite{mirza2014conditional}. On the contrary, the image-to-image translation framework uses a combination of conditional and output fields produced from the generator or a real field (training/testing set) as its input \cite{isola2017image}. The schematic of the patch discriminator is shown in Fig. \ref{fig:disc}. The discriminator utilizes the contracting block (see Fig. \ref{fig:disc} - light and dark orange blocks), which is also used in the generator. The purple block is used to illustrate that our discriminator output is not a single value, but a matrix \cite{demir2018patch}.  \par

The detail of patch discriminator architecture is shown in Table \ref{tab:disc}. Each contracting block (see Fig. \ref{fig:disc} - light and dark orange blocks) uses LeakyReLU with a negative slope of 0.2 as its activation function, the $1^{\mathrm{st}}$ convolutional layer is used to map $\mathbb{C}$ $+$ conditional field to $\mathbb{H} = 8$, and it does not subject to any activation function, and and the $2^{\mathrm{nd}}$ convolutional layer is used to map $\mathbb{H} = 8$ to $\mathbb{C}$. Dissimilar to the generator where the input size is equal to output size $\mathbb{DOF_X}$ $\times$ $\mathbb{DOF_Y}$ $=$ 128 $\times$ 128, the discriminator input size is $\mathbb{DOF_X}$ $\times$ $\mathbb{DOF_Y}$ $=$ 128 $\times$ 128, while the output size is a patch matrix of size $\mathbb{PATCH_X}$ $\times$ $\mathbb{PATCH_Y}$ $=$ 8 $\times$ 8. We note that we have tried different generator and discriminator architectures such as fully connected layers, deep convolutional networks, or U-Net without residual blocks. We select these generator and discriminator architectures, Figs. \ref{fig:unet} and \ref{fig:disc} because they provide the best performance based on our trial-and-error process. \par

\begin{table}[!ht]
\centering
\caption{Discriminator: Patch discriminator's detail used in this study (input size is represented by {[}$\mathbb{B}$, $\mathbb{C}$, $\mathbb{DOF_X}$, $\mathbb{DOF_Y}${]}, and output size is represented by {[}$\mathbb{B}$, $\mathbb{C}$, $\mathbb{PATCH_X}$, $\mathbb{PATCH_Y}${]}. We use hidden layers $\mathbb{H} = 8$). Note that the spectral normalization is applied only for the SN model (see Section \ref{sec:cgan_sn}).}
\begin{tabular}{|l|c|c|c|c|c|}
\hline
Block                  & \multicolumn{1}{l|}{Input size} & \multicolumn{1}{l|}{Output size} & \multicolumn{1}{l|}{Batch normalization} & \multicolumn{1}{l|}{Dropout} & \multicolumn{1}{l|}{Spectral normalization} \\ \hline
$1^{\mathrm{st}}$ convolutional layer & {[}$\mathbb{B}$, $\mathbb{C}$+1, 128, 128{]}            & {[}$\mathbb{B}$, 8, 128, 128{]}             &                                          &          & \checkmark                     \\ \hline
$1^{\mathrm{st}}$ contracting block   & {[}$\mathbb{B}$, 8, 128, 128{]}            & {[}$\mathbb{B}$, 16, 64, 64{]}              &                                          &    & \checkmark                            \\ \hline
$2^{\mathrm{nd}}$ contracting block   & {[}$\mathbb{B}$, 16, 64, 64{]}             & {[}$\mathbb{B}$, 32, 32, 32{]}              & \checkmark                                        &       & \checkmark                         \\ \hline
$3^{\mathrm{rd}}$ contracting block   & {[}$\mathbb{B}$, 32, 32, 32{]}             & {[}$\mathbb{B}$, 64, 16, 16{]}              & \checkmark                                        &            & \checkmark                   \\ \hline
$4^{\mathrm{th}}$ contracting block   & {[}$\mathbb{B}$, 64, 16, 16{]}             & {[}$\mathbb{B}$, 128, 8, 8{]}               & \checkmark                                         &                & \checkmark               \\ \hline
$2^{\mathrm{nd}}$ convolutional layer & {[}$\mathbb{B}$, 128, 8, 8{]}              & {[}$\mathbb{B}$, $\mathbb{C}$, 8, 8{]}                 &                     &                     & \multicolumn{1}{c|}{ \checkmark }        \\ \hline
\end{tabular}
\label{tab:disc}
\end{table}

We note that the input of the discriminator is combined fields of input ($\bm{I}$) and output ($\bm{O}$) fields of the generator; hence, they are normalized by \eqref{eq:norm_input_output} and range from 0 to 1. In this study, we compare three different models' performance in both accuracy and computational costs. The detail of each model is provided in the following sections. \par

\subsection{Base model}\label{sec:cgan_base}

As previously discussed, the cGAN model is generally composed of one generator and one discriminator (see Figs. \ref{fig:unet} and \ref{fig:disc}). The goal of the generator is to generate an approximated output field ($\widehat{\bm{O}}$) that looks realistic and similar to a real output field ($\bm{O}$) with a given conditional field ($\bm{I}$). The discriminator, on the other hand, tries to differentiate between $\widehat{\bm{O}}$ and $\bm{O}$. Consequently, to train the framework (both generator and discriminator), we equivalently try to solve the following constraint


\begin{equation} \label{eq:total_min_max}
\min _{G} \max _{D} \left[ \ell_{a} +\lambda_{{r}}\ell_r \right].
\end{equation}

\noindent
Here, $\ell_{a}$ is an adversarial loss defined as \cite{goodfellow2014generative,liu2016coupled}

\begin{equation} \label{eq:adver_loss}
{\ell_a}= \frac{1}{\mathbb{B}}\sum_{i = 1}^{\mathbb{B}} \left( {O_{i}} \cdot \log \widehat{O_{i}}+\left(1-{O_{i}}\right) \cdot \log \left(1-\widehat{O_{i}}\right) \right),
\end{equation}

\noindent
where, again, $\mathbb{B}$ is batch size and $\mathbb{B} \leq \mathrm{M}$, which is the total number of training examples. The $\lambda_r$ is a  penalty constant selected by users, and we set $\lambda_r=500$ throughout this paper. The $\ell_r$ is read

\begin{equation} \label{eq:l1_loss}
\ell_r= \frac{1}{\mathbb{B}} \sum_{i = 1}^{\mathbb{B}}  \left|\widehat{O_{i}}-O_{i}\right|.
\end{equation}

\noindent
We choose L1 norm (i.e., \eqref{eq:l1_loss}) over L2 norm because it has shown to improve GAN performance \cite{pandey2018adversarial}. We would like to emphasize that according to \eqref{eq:total_min_max} and \eqref{eq:adver_loss}, the discriminator's goal is to maximize $\frac{1}{\mathbb{B}}\sum_{i = 1}^{\mathbb{B}} \left( {O_{i}} \cdot \log \widehat{O_{i}}+\left(1-{O_{i}}\right) \cdot \log \left(1-\widehat{O_{i}}\right) \right)$, which equivalently means that it tries to differentiate between $\widehat{\bm{O}}$ and $\bm{O}$. The generator of the base model, on the other hand, is not affected by ${O_{i}} \cdot \log \widehat{O_{i}}$ term in \eqref{eq:adver_loss}. Hence, the goal of the generator is to minimize $\frac{1}{\mathbb{B}}\sum_{i = 1}^{\mathbb{B}} \left(1-{O_{i}}\right) \cdot \log \left(1-\widehat{O_{i}}\right)$ and $\frac{1}{\mathbb{B}} \sum_{i = 1}^{\mathbb{B}}  \left|\widehat{O_{i}}-O_{i}\right|$. By satisfying these two constraints , the generator attempts to produce realistic $\widehat{\bm{O}}$.

We use the adaptive moment estimation (ADAM) algorithm \cite{kingma2014adam} to train the framework (both generator and discriminator). The learning rate ($\eta$) is calculated as \cite{loshchilov2016sgdr}

\begin{equation}
\eta_{c}=\eta_{\min }+\frac{1}{2}\left(\eta_{\max }-\eta_{\min }\right)\left(1+\cos \left(\frac{\mathrm{step_c}}{\mathrm{step_f}} \pi\right)\right)
\end{equation}

\noindent
where $\eta_{c}$ is a learning rate at step $\mathrm{step_c}$, $\eta_{\min }$ is the minimum learning rate, which is set as $1 \times 10^{-16}$, $\eta_{\max }$ is the maximum or initial learning rate, which is selected as $1 \times 10^{-4}$, $\mathrm{step_c}$ is the current step, and $\mathrm{step_f}$ is the final step. We note that each step refers to each time we perform back-propagation, including updating both generator and discriminator's parameters. In certain circumstances, one could do multiple rounds of back-propagating on the generator while updating the discriminator only once and vice versa to achieve more stable training. However, in this study, we update the generator and discriminator's parameters one time per step.    \par

\subsection{Spectral normalization generative adversarial networks (SN model)}\label{sec:cgan_sn}

Studies show that employing the loss terms and architecture in the base model in Section \ref{sec:cgan_base} may result in unstable training (e.g., mode collapse or vanishing gradient problem), which could lead to the early ceasing of the GAN learning process \cite{arjovsky2017wasserstein,gulrajani2017improved,miyato2018spectral}. Therefore, we adapt the concept of spectral normalization proposed in \cite{miyato2018spectral}. This spectral normalization enforces the constraint in the discriminator that the weights matrix exhibits Lipschitz continuity (Euclidean norm of discriminator's gradient is at most one), which could provide a certain level of functional smoothness in all directions during the training process. The spectral normalization is applied to weight matrices ($\mathbf{W}$) resulting in (see \cite{miyato2018spectral,NEURIPS2019_9015})

\begin{equation}
\mathbf{W}_{\text{SN}}=\frac{\mathbf{W}}{\text{SN}(\mathbf{W})}.
\end{equation}




We apply spectral normalization only to the discriminator as shown in Table \ref{tab:disc}. The loss terms for both generator and discriminator are similar to the base model (see \eqref{eq:total_min_max}). 

\subsection{Wasserstein generative adversarial networks (W model)}\label{sec:cgan_w}

To prevent the GAN model from mode collapse or vanishing gradient problem, we could also use the Wasserstein loss or W loss \cite{arjovsky2017wasserstein,gulrajani2017improved}. The W loss employs the Earth mover's distance, enforcing the distribution of the generator's output to be similar to that of the actual distribution \cite{frogner2015learning,arjovsky2017wasserstein}. Dissimilar to the previous models, the W model uses the following constraint  \par

\begin{equation} \label{eq:total_min_max_add_wloss}
\min _{G} \max _{D} \left[ \ell_{a} +\lambda_{{r}}\ell_r  +\lambda_{{p}}\wp_p \right].
\end{equation}

\noindent
where $\ell_{a}$ in this model is the Earth mover's distance defined as

\begin{equation}
\ell_{a} = \mathbb{E} \left(D(\bm{I} + \bm{O})\right)-\mathbb{E}\left(D\left(\bm{I} + \widehat{\bm{O}} \right)\right).
\end{equation}

\noindent
Here, $D(\bm{I} + \bm{O})$ is an output matrix from the patch discriminator (see Fig. \ref{fig:disc}) using a real output ($\bm{O}$); $D\left(\bm{I} + \widehat{\bm{O}} \right)$, on the other hand, is an output matrix from the patch discriminator using an approximated output $\widehat{\bm{O}}$ produced by the generator ($G(\bm{I}) = \widehat{\bm{O}}$); $\mathbb{E}(\cdot)$ denotes an expectation; $\lambda_{{p}}$ denotes a gradient penalty constant that we set to 10 throughout this study; $\wp_p$ gradient penalty regularization; $\wp_p$ is used to enforce Lipschitz continuity of weight matrices ($\mathbf{W}$), i.e., Euclidean norm of discriminator’s gradient is at most one. \par

We note that there are two common ways of ensuring Lipschitz continuity of $\mathbf{W}$; weight clipping and gradient penalty \cite{arjovsky2017wasserstein,gulrajani2017improved}. With weight clipping, $\mathbf{W}$ of the discriminator is forced to take values between a fixed interval, which means that weights over that interval, either too high or too low, will be set to the maximum or the minimum amount allowed. One of the disadvantages is forcing the discriminator's $\mathbf{W}$ to a limited range of values could limit the discriminator's ability to learn and ultimately resulting in an early termination of the learning process \cite{gulrajani2017improved}. \par

The gradient penalty, another method, is a much softer way to enforce the discriminator to be Lipschitz continuous. The gradient penalty adds a regularization term ($\lambda_{{p}}\wp_p$) to the loss function \eqref{eq:total_min_max_add_wloss}. This term penalizes the discriminator when its gradient norm is higher than one by

\begin{equation}
   \wp_p = \mathbb{E}\left(\|\nabla D(\bm{I} + \Bar{\bm{O}})\|_{2}-1\right)^{2}
\end{equation}

\noindent
where $\Bar{\bm{O}}$ is a mixing between $\widehat{\bm{O}}$ and $\bm{O}$, which is defined by

\begin{equation}
   \Bar{\bm{O}} = \bm{\epsilon} \bm{O} + (1 -  \bm{\epsilon})\widehat{\bm{O}}.
\end{equation}

\noindent
We randomly select $\epsilon_i \in \bm{\epsilon}$ for each $\Bar{{O}}_i$ from a uniform distribution on the interval of $[0, 1)$.

\begin{remark}
 We note that in some literature, the `discriminator' of Wasserstein generative adversarial networks or W model in this paper is referred to as `critic' since its output is not bounded by 0 or 1 \cite{gulrajani2017improved,arjovsky2017wasserstein}. Throughout this paper, however, we refer the `critic' to `discriminator' for the sake of simplicity.
\end{remark}

\begin{remark}
Finite element results obtained from \cite{Kadeethum2019ARMA, kadeethum2020enriched, kadeethum2020finite, kadeethum2021non} are based on unstructured grids. Hence, we pre-processing our data by interpolating the finite element results as well as permeability field populated by \cite{muller2020} to structured grids using cubic spline interpolation and use this processed data in our ROM framework.
\end{remark}

\beginsupplementsub
\section{Supplementary proper orthogonal decomposition (POD) analysis}\label{sec:sup_pod}

\begin{figure}[!ht]
   \centering
         \includegraphics[keepaspectratio, height=5.5cm]{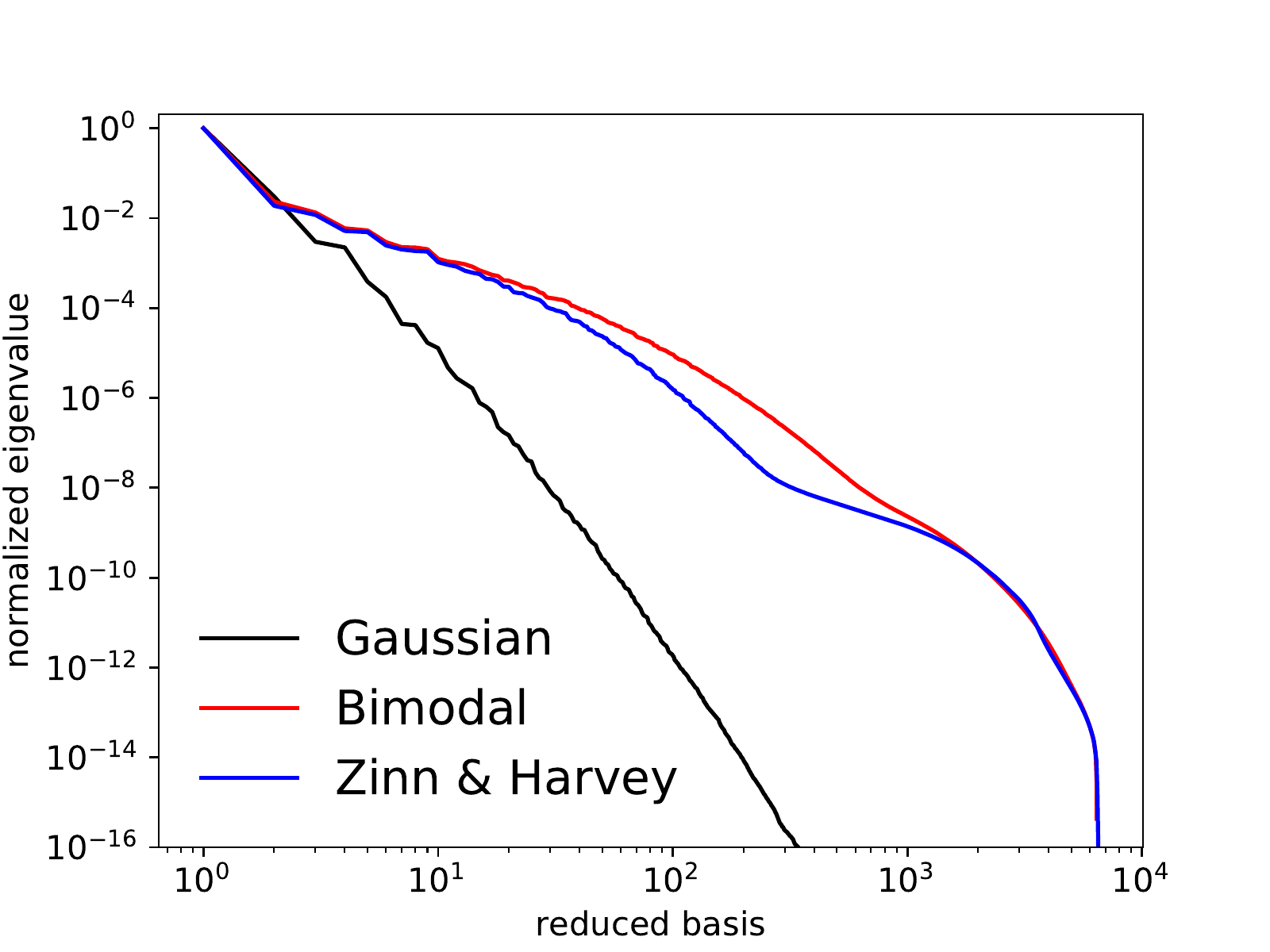}
   \caption{Normalized eigenvalue as a function of basis for fluid pressure field ($p_h$).}
   \label{fig:rb}
\end{figure}

We perform proper orthogonal decomposition (POD) on the pressure field $p_h$ obtained from FOM discussed in \ref{sec:governing_equations} using RBniCS project \cite{ballarin2015rbnics}. The decay of eigenvalues results of $p_h$ using (1) permeability field as Gaussian distribution \ref{sec:forward_gs}, (2) permeability field as bimodal transformation \ref{sec:forward_bm}, and (3) permeability field as Zinn \& Harvey transformation \ref{sec:forward_zh} are presented in Figure \ref{fig:rb}. From this figure, we could observe that the decay is very slow, which means POD, low-dimensional linear subspace, poorly approximates the $p_h$ field. We note that the total available data is 10000. The result is based on a "global" POD approach and using “local” POD method \cite{choi2020gradient} may improve a performance of POD.

\beginsupplementsub

\section{Supplementary numerical examples}\label{sec:sup_results}

\subsection{Supplementary forward modeling}\label{sec:sup_forward}

We utilize pressure and displacement fields at the steady-state solutions of discontinuous Galerkin finite element model of linear poroelasticity developed in \cite{Kadeethum2019ARMA, kadeethum2020enriched, kadeethum2020finite, kadeethum2021non, kadeethum2021locally} to train our ROM model. Throughout this section, we fix our framework parameters as follows; the model architecture is shown in Figs.\ref{fig:unet} and \ref{fig:disc} and discussed in Section \ref{sec:cgan}. The input and output fields have a resolution of $128 \times 128$. The input has one channel, but the output could have one, two, or three channels (i.e., pressure, the displacement in the x-direction, or the displacement in the y-direction). Note that we present here, the forward modeling part, only the results of pressure field ${p}_{h}$, for the sake of compactness. A very similar result is held for the displacement field ([$u_{x_h}$, $u_{y_h}$] $\in$  $\bm{u}_{h}$) as well. The batch size is fixed at four as it has been shown that GAN or cGAN model generally requires a small batch size to improve model accuracy \cite{karras2017progressive}. We use the adaptive moment estimation (ADAM) algorithm as an optimizer \cite{kingma2014adam}. Its learning rate is 0.0001, and $\beta$ is (0.5, 0.999). The L1 penalty coefficient ($\lambda_{{r}}$ in \eqref{eq:total_min_max} or \eqref{eq:total_min_max_add_wloss}) is 500. For the W model, we use gradient penalty coefficient ($\lambda_{{p}}$ in \eqref{eq:total_min_max_add_wloss}) of 10 as recommended by \cite{gulrajani2017improved}. \par

\subsubsection{Example 1: Permeability field as Gaussian distribution}\label{sec:forward_gs}

\begin{figure}[!ht]
   \centering
        \includegraphics[keepaspectratio, height=5.5cm]{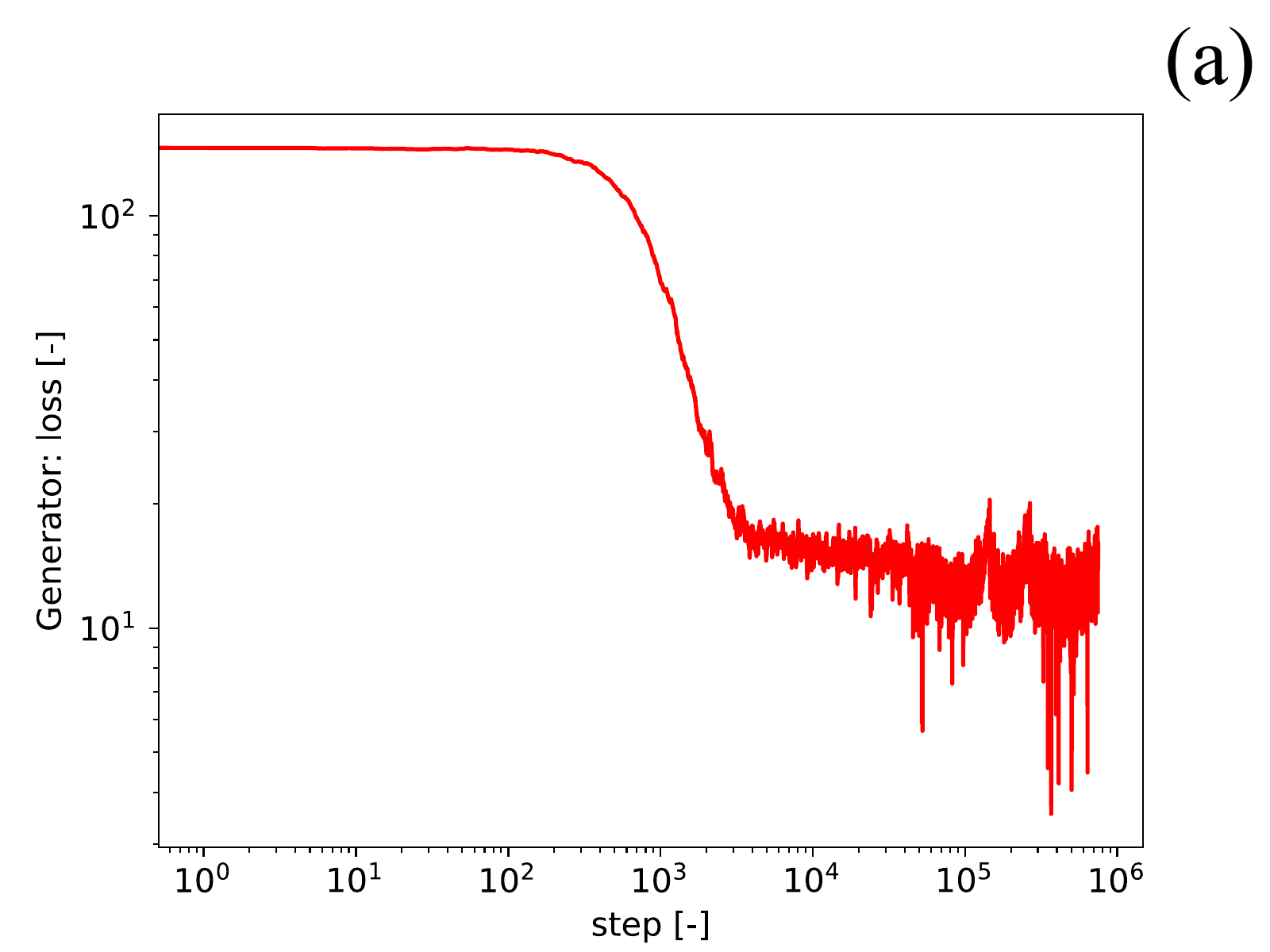}
         \includegraphics[keepaspectratio, height=5.5cm]{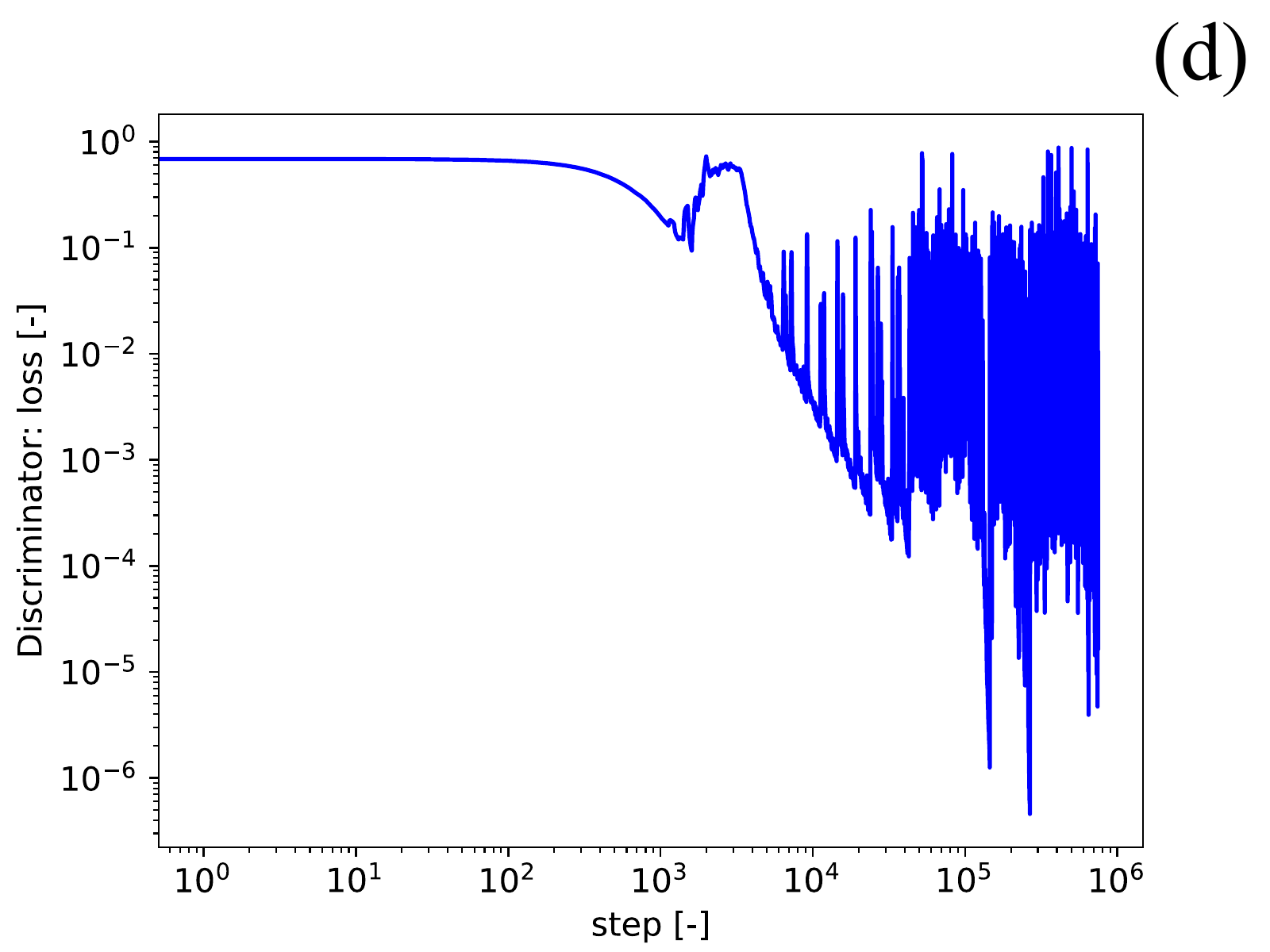}
         \includegraphics[keepaspectratio, height=5.5cm]{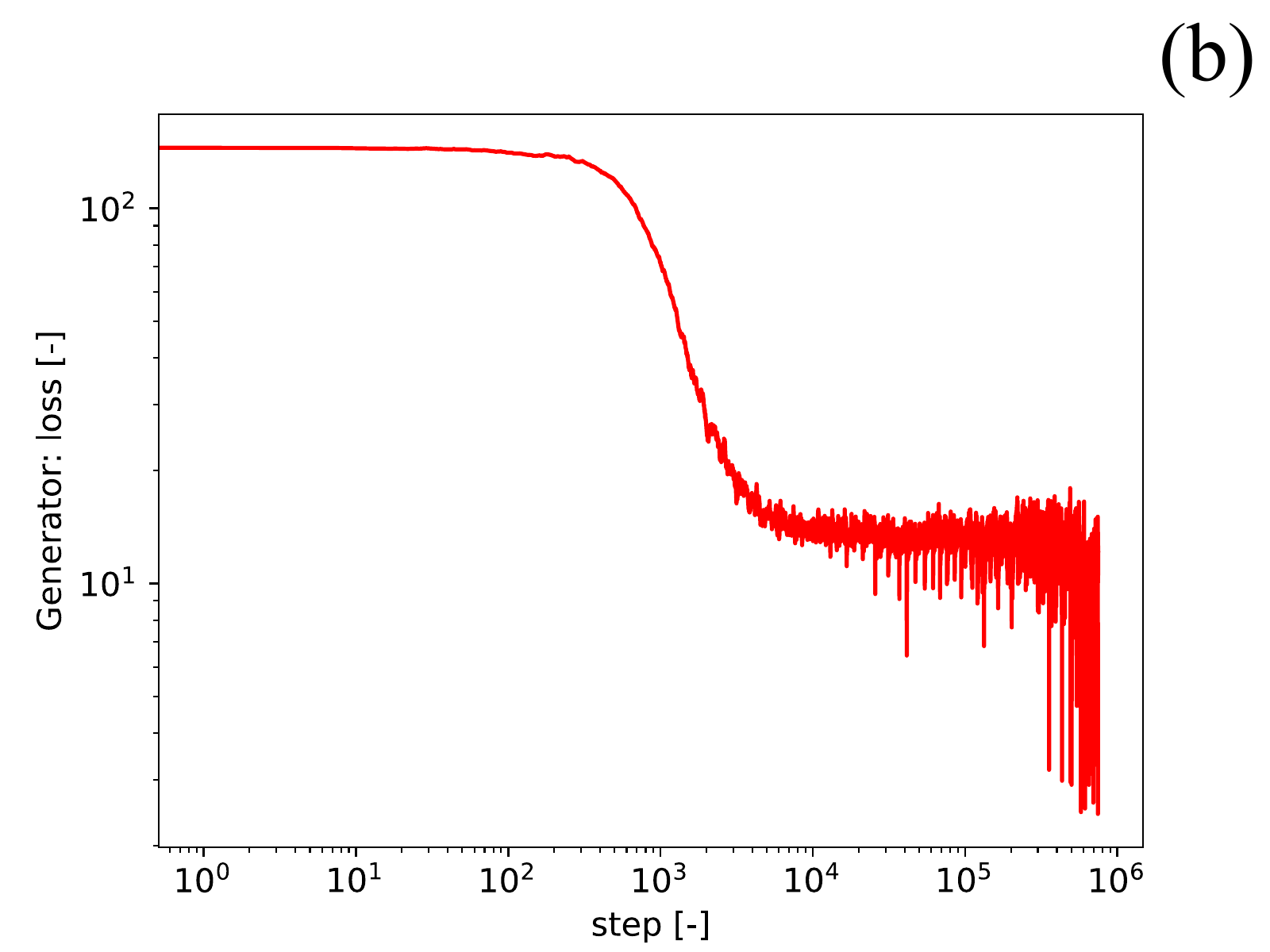}
         \includegraphics[keepaspectratio, height=5.5cm]{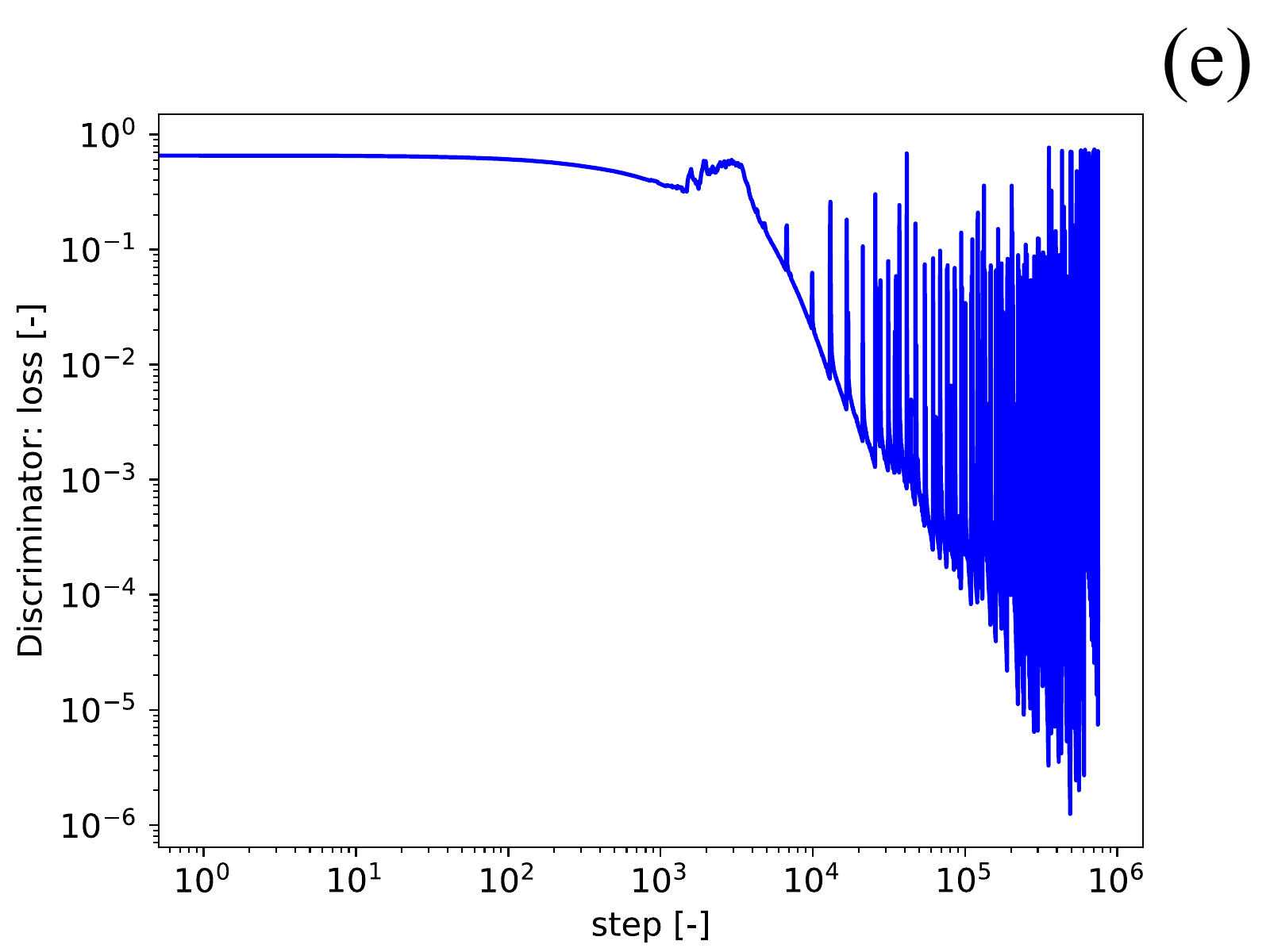}
         \includegraphics[keepaspectratio, height=5.5cm]{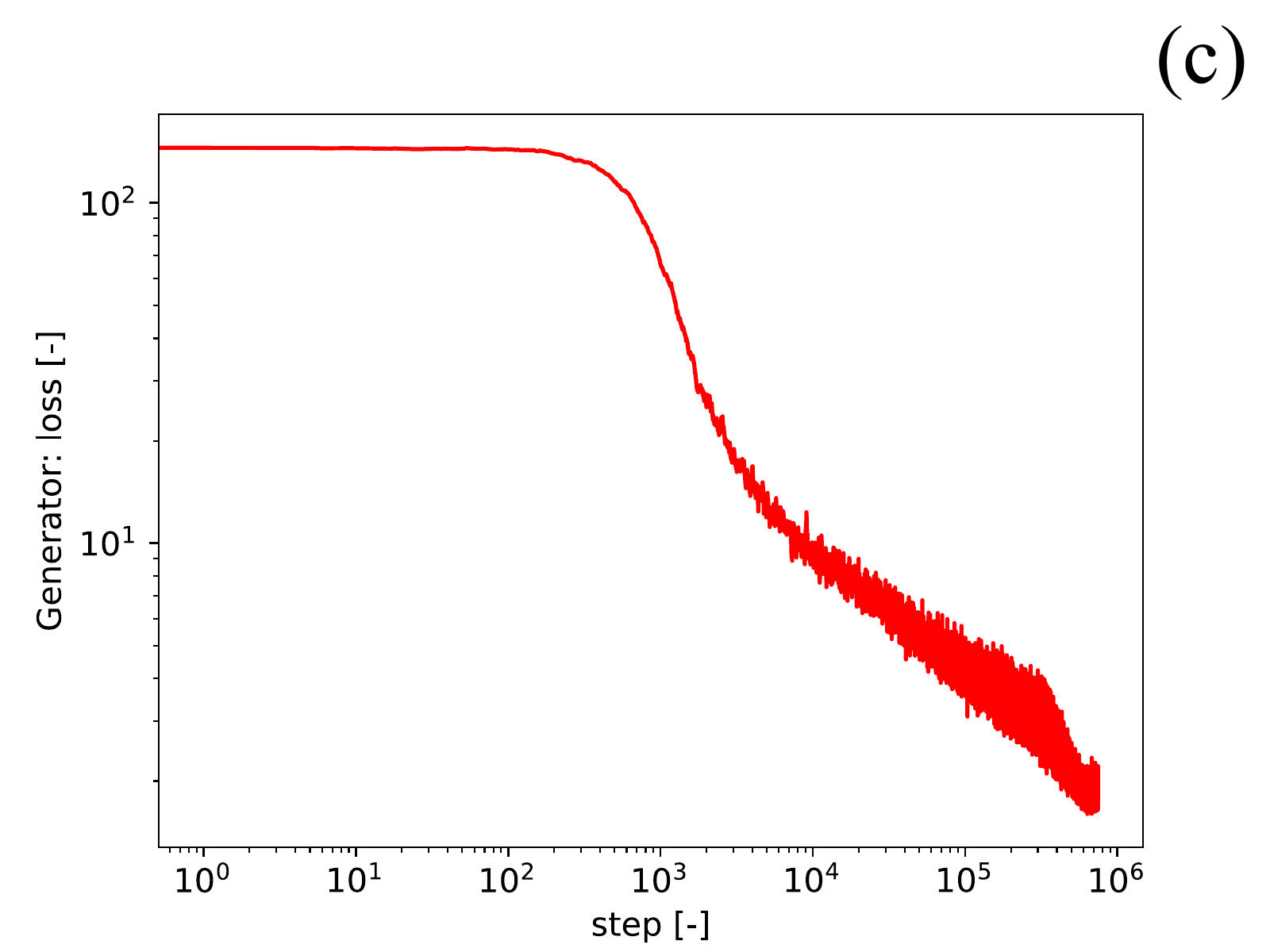}
         \includegraphics[keepaspectratio, height=5.5cm]{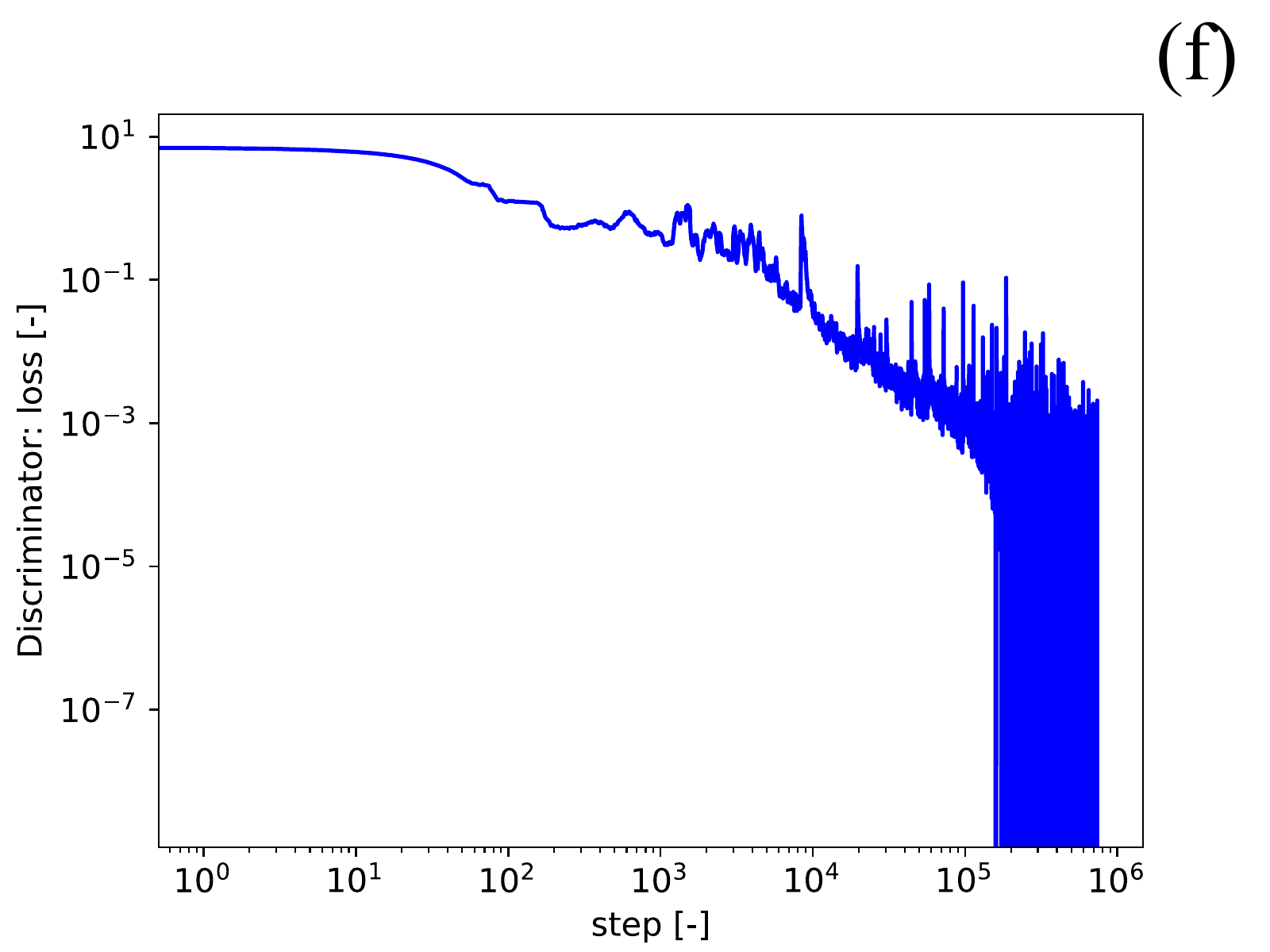}
   \caption{Example 1: 100$^{th}$ moving average of Generator loss of (a) base model, (b) SN model, and (c) W model and Discriminator loss of (d) base model, (e) SN model, and (f) W model   }
   \label{fig:ex1_gen_disc_loss}
\end{figure}

This section explores the ROM model's behavior using permeability fields populated using an unconditional Gaussian distribution \cite{jensen2000statistics,muller2020}. The mean of log-permeability (base 10) field is -12 \si{log(m^2)}, the variance is 1 \si{log(m^4)}, and the covariance kernel is Gaussian (i.e., Squared exponential) with the length-scale is 1.0 \si{m}. For the sake of brevity, all 'log' in this manuscript refers to 'log10.' We create 11000 pairs of permeability and primary variable fields. We use 10000 pairs to train the ROM model and test the model with 1000 cases. To reiterate, we have three types of model (1) base model, (2) SN model, and (3) W model as discussed in Section \ref{sec:cgan}. The training losses of both generator and discriminator are presented in Fig. \ref{fig:ex1_gen_disc_loss}.  \par


In Fig. \ref{fig:ex1_gen_disc_loss}, each step refers to each time we perform back-propagation including updating both generator and discriminator's parameters. Comparing Figs. \ref{fig:ex1_gen_disc_loss}a-c, we could observe that the generator loss of the W model behaves more stable than those of the base and SN models. Besides, there is no significant difference between the base and SN models. We also observe that the W model has better stability in training than the base and SN models for the discriminator losses. Its loss value reaches zero slower than those of the base and SN models (i.e., the learning process ceases later \cite{gulrajani2017improved,arjovsky2017wasserstein}) \par

\begin{figure}[!ht]
   \centering

         \includegraphics[keepaspectratio, height=3.5cm]{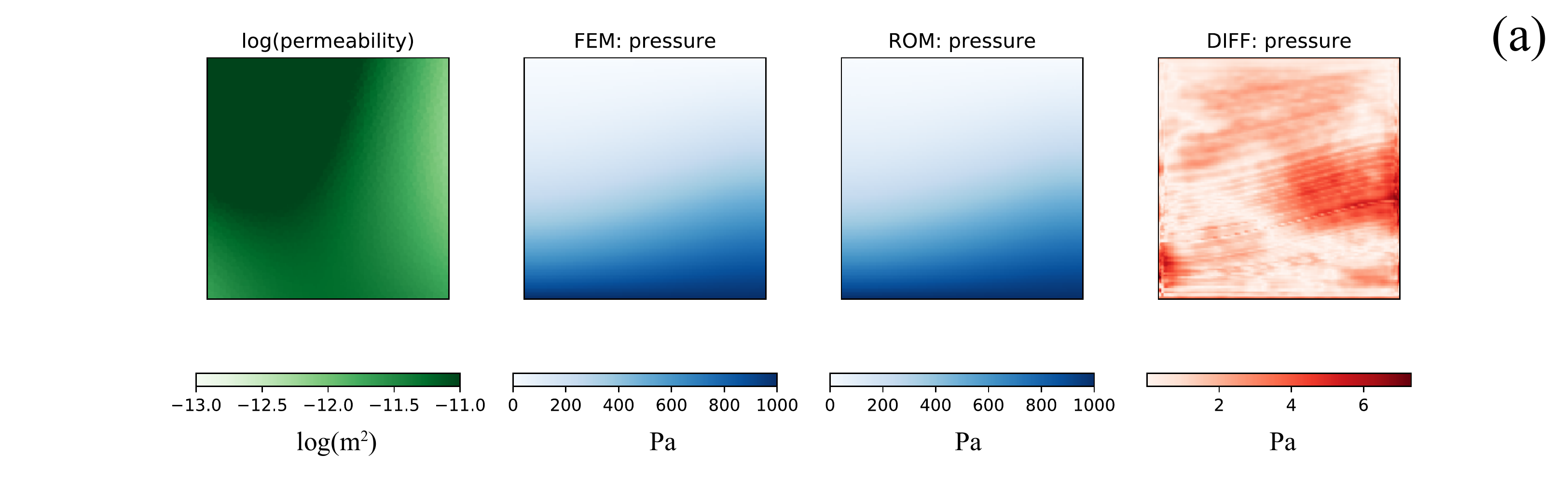}
         \includegraphics[keepaspectratio, height=3.5cm]{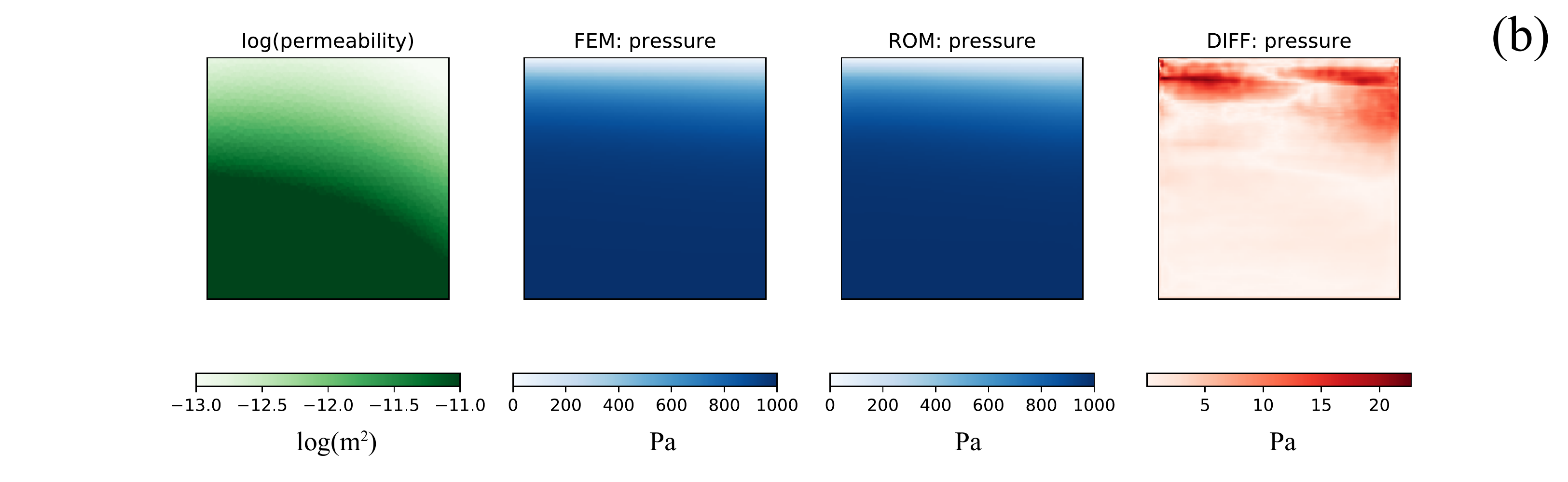}
         \includegraphics[keepaspectratio, height=3.5cm]{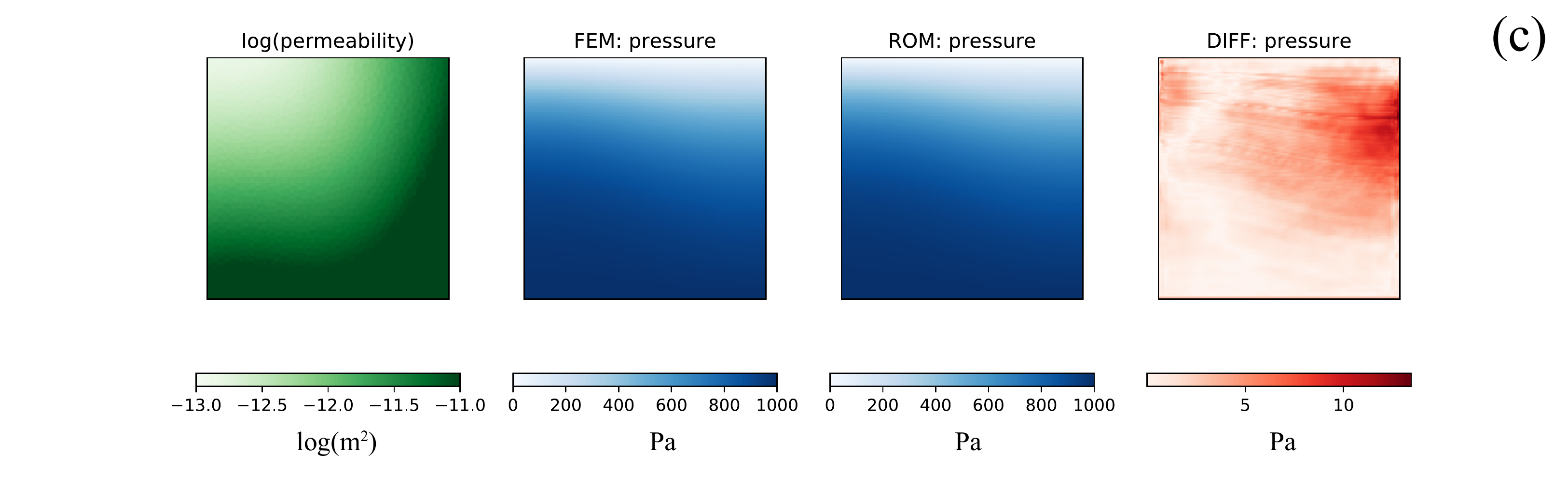}
   \caption{Example 1: test cases' results of the W model. Note that we train our three models (base, SN, and W models) using 10000 training examples and test them using 1000 test examples. These three cases shown here are randomly picked from 1000 test examples.}
   \label{fig:ex1_pic}

\end{figure}

We then test our model with 1000 testing examples. Three of those cases are presented in Fig. \ref{fig:ex1_pic}. From this figure, we observe that our models could provide a decent approximation of the pressure field. The range of pressure values is from 0.0 to 1000.0 Pa; the DIFF values \eqref{eq:diff} are approximately in the range of 0.0 to 10.0 Pa, which means the errors are approximately 0.1\%. Again, for the sake of compactness, we only show here the results of the pressure field, and the results of the displacement field are very similar.

\begin{figure}[!ht]
   \centering

         \includegraphics[keepaspectratio, height=3.5cm]{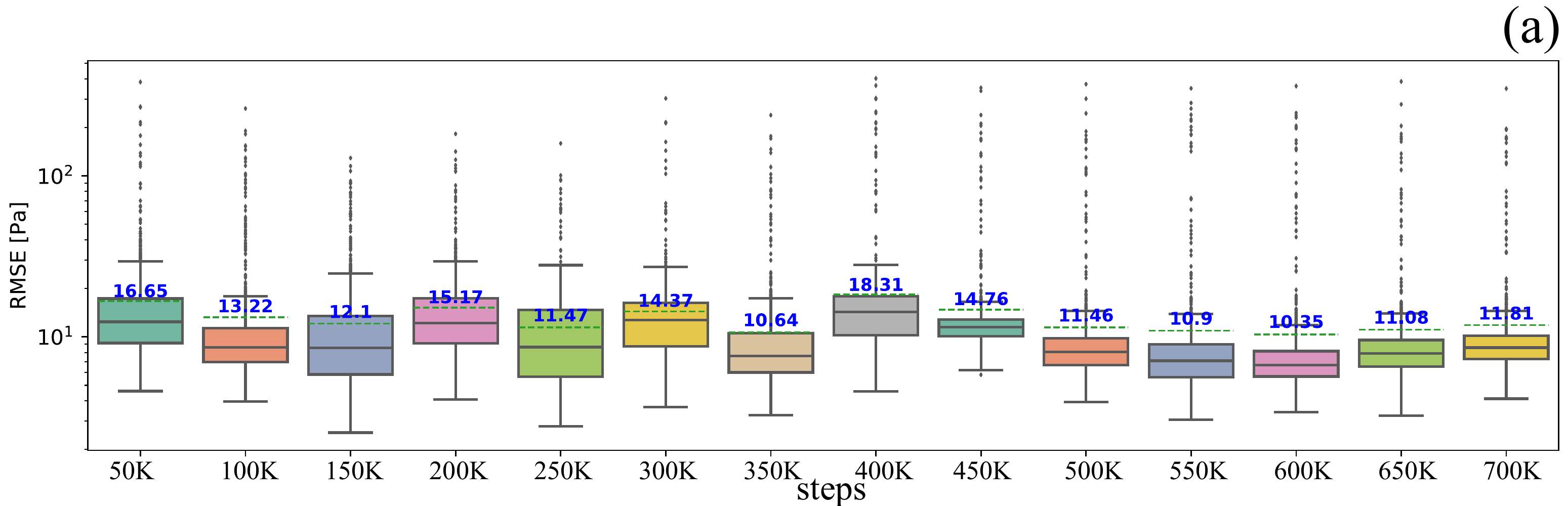}
         \includegraphics[keepaspectratio, height=3.5cm]{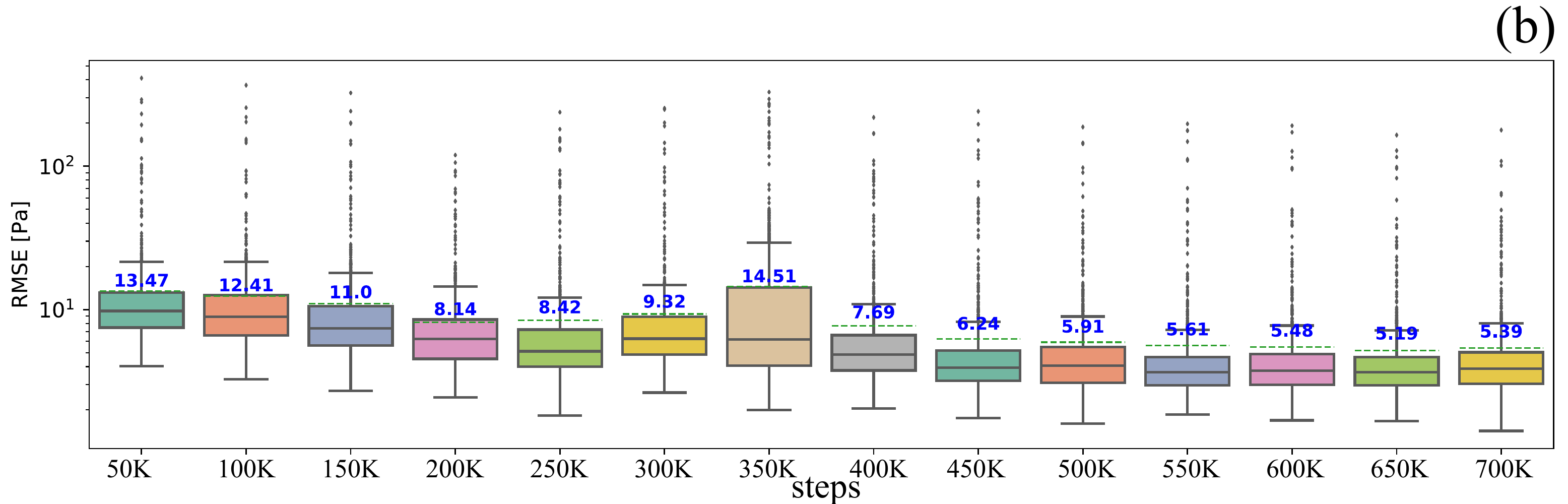}
         \includegraphics[keepaspectratio, height=3.5cm]{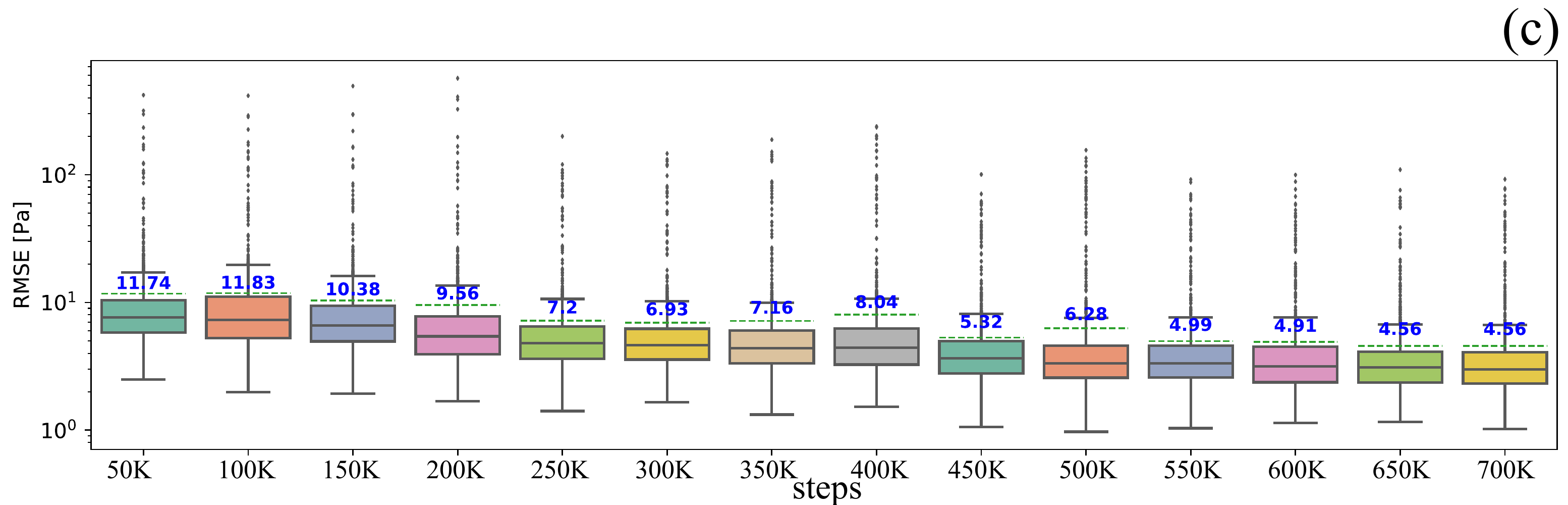}
   \caption{Example 1: Root Mean Square Error (RMSE) of (a) base model, (b) SN model, and (c) W model. Each step refers to each time we perform back-propagation including updating both generator and discriminator's parameters.}
   \label{fig:ex1_test}

\end{figure}

The RMSE of the base, SN, and W models are shown in Fig. \ref{fig:ex1_test}. We note that these RMSE values are calculated from 1000 test cases (i.e., $N=1000$). From Fig. \ref{fig:ex1_test}, one can observe that the RMSE values of the W model are the most stable, the ones from the SN model are a little bit less stable, while the results of the base model are not stable. \par

The best of an average value of RMSE is 10.35 \si{Pa}, which is about 1\% of the maximum pressure. The maximum and minimum of RMSE values are approximately 30\% and 0.4\% of the maximum pressure, respectively. For the SN model, the best RMSE has an average of 5.19 \si{Pa} or 0.5\% of the maximum pressure. The RMSE maximum and minimum of this model are about 20\% and 0.3\% of the maximum pressure, respectively. The W model's best RMSE has an average of 4.36 \si{Pa}, which is about 0.4\% of the maximum pressure. The maximum and minimum RMSE values of the W model are approximately 15\% and 0.1\% of the maximum pressure, respectively.

\subsubsection{Example 2: Bimodal permeability fields}\label{sec:forward_bm}

\begin{figure}[!ht]
   \centering
        \includegraphics[keepaspectratio, height=5.5cm]{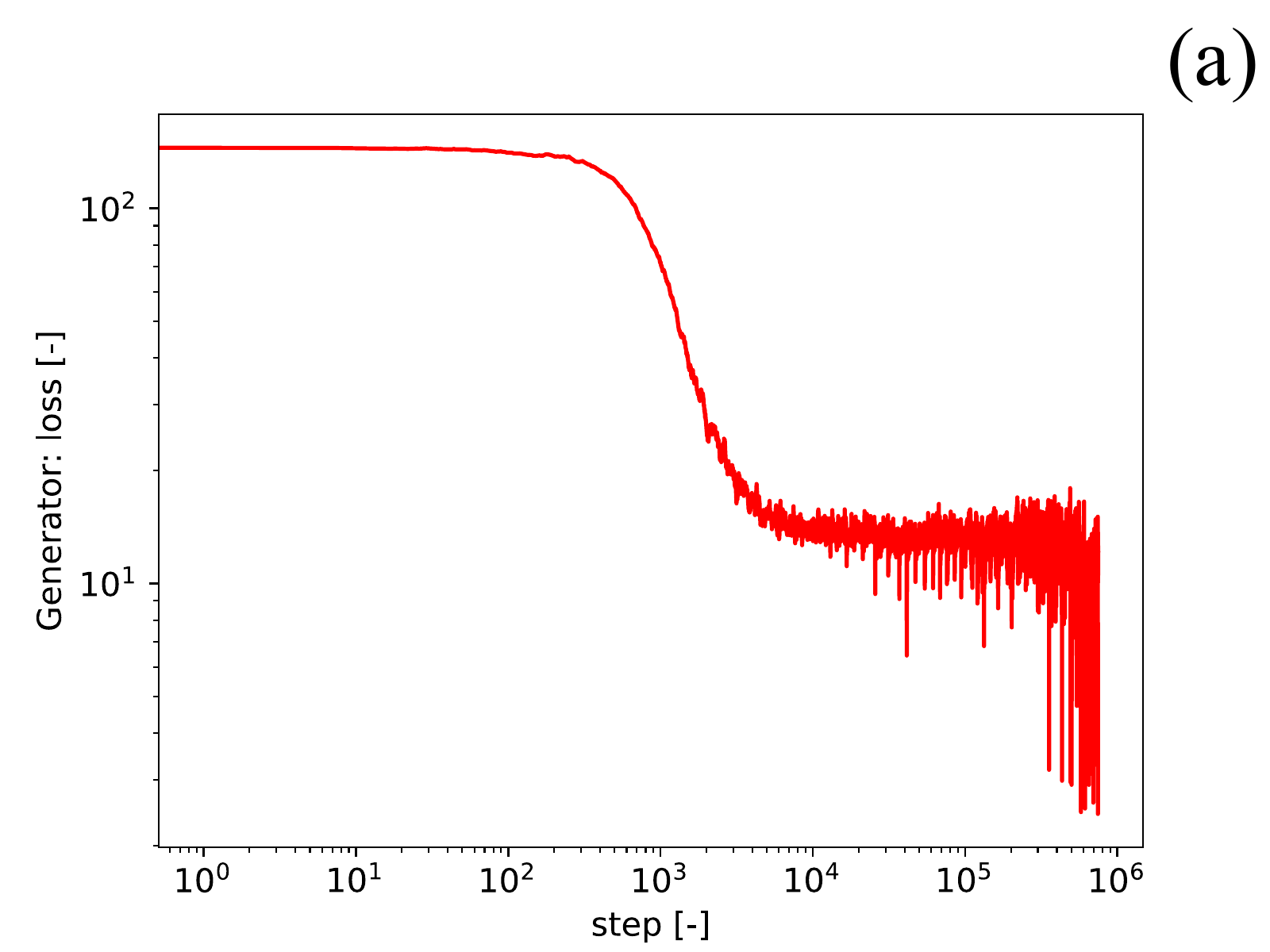}
         \includegraphics[keepaspectratio, height=5.5cm]{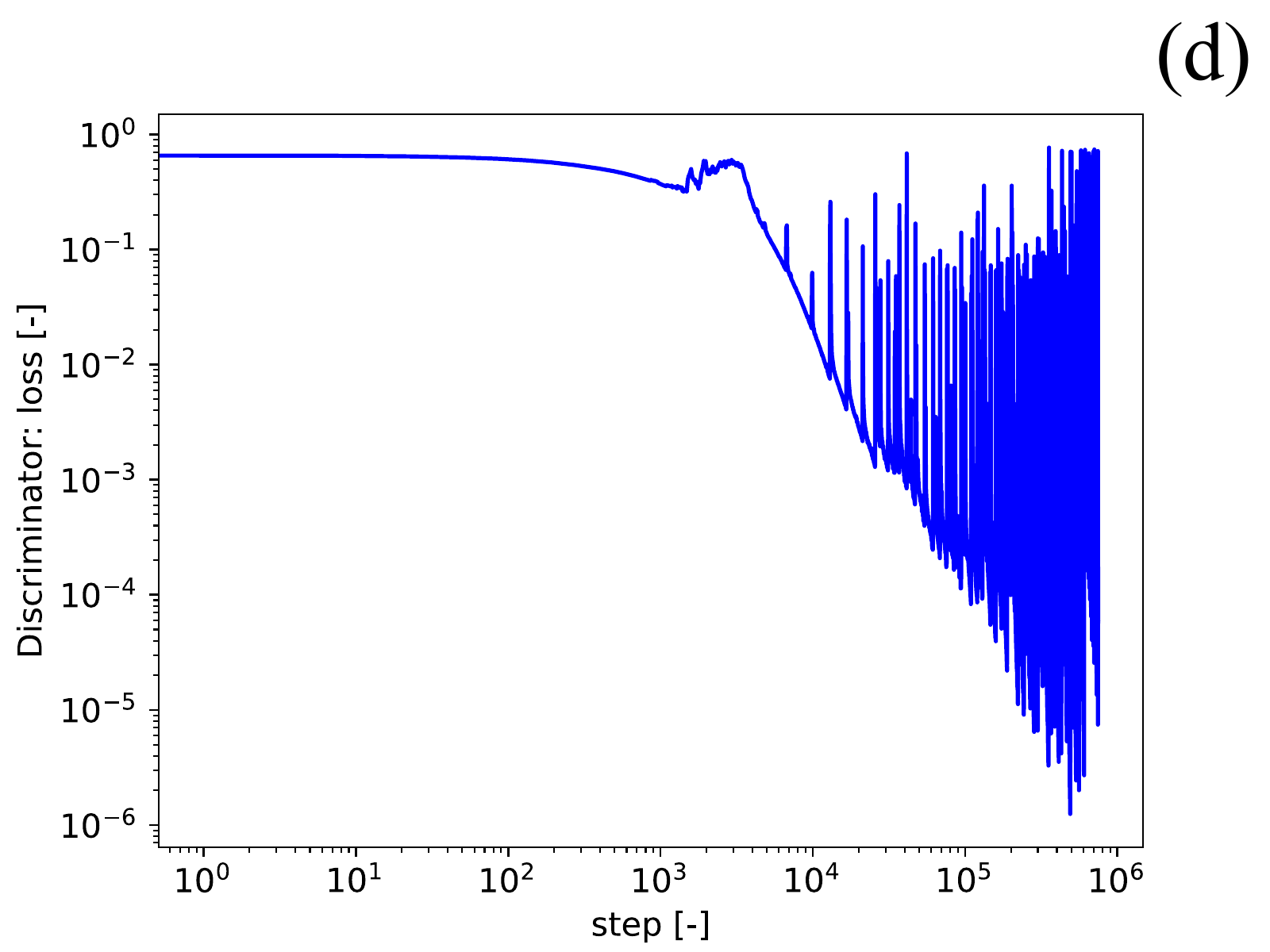}
         \includegraphics[keepaspectratio, height=5.5cm]{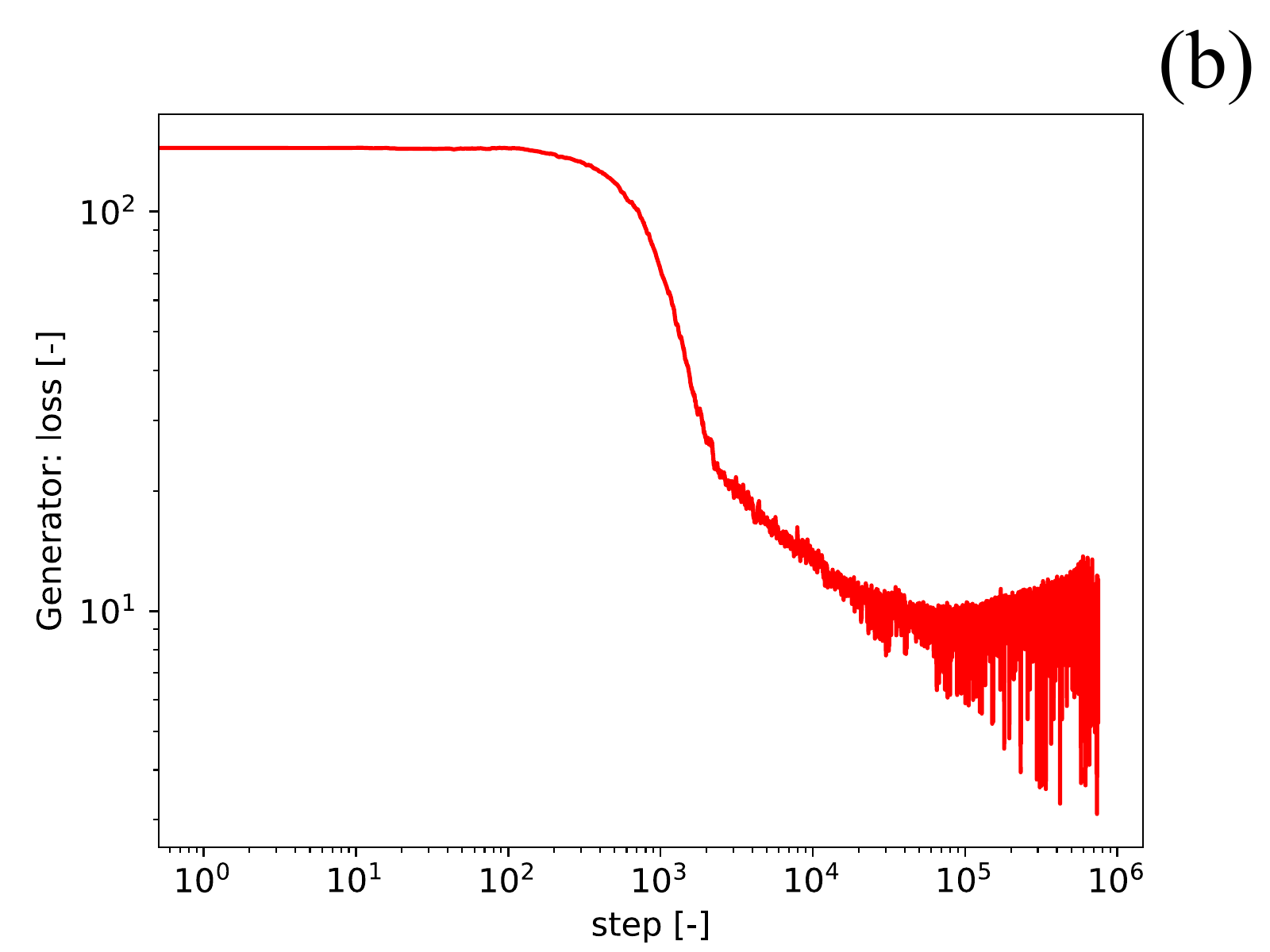}
         \includegraphics[keepaspectratio, height=5.5cm]{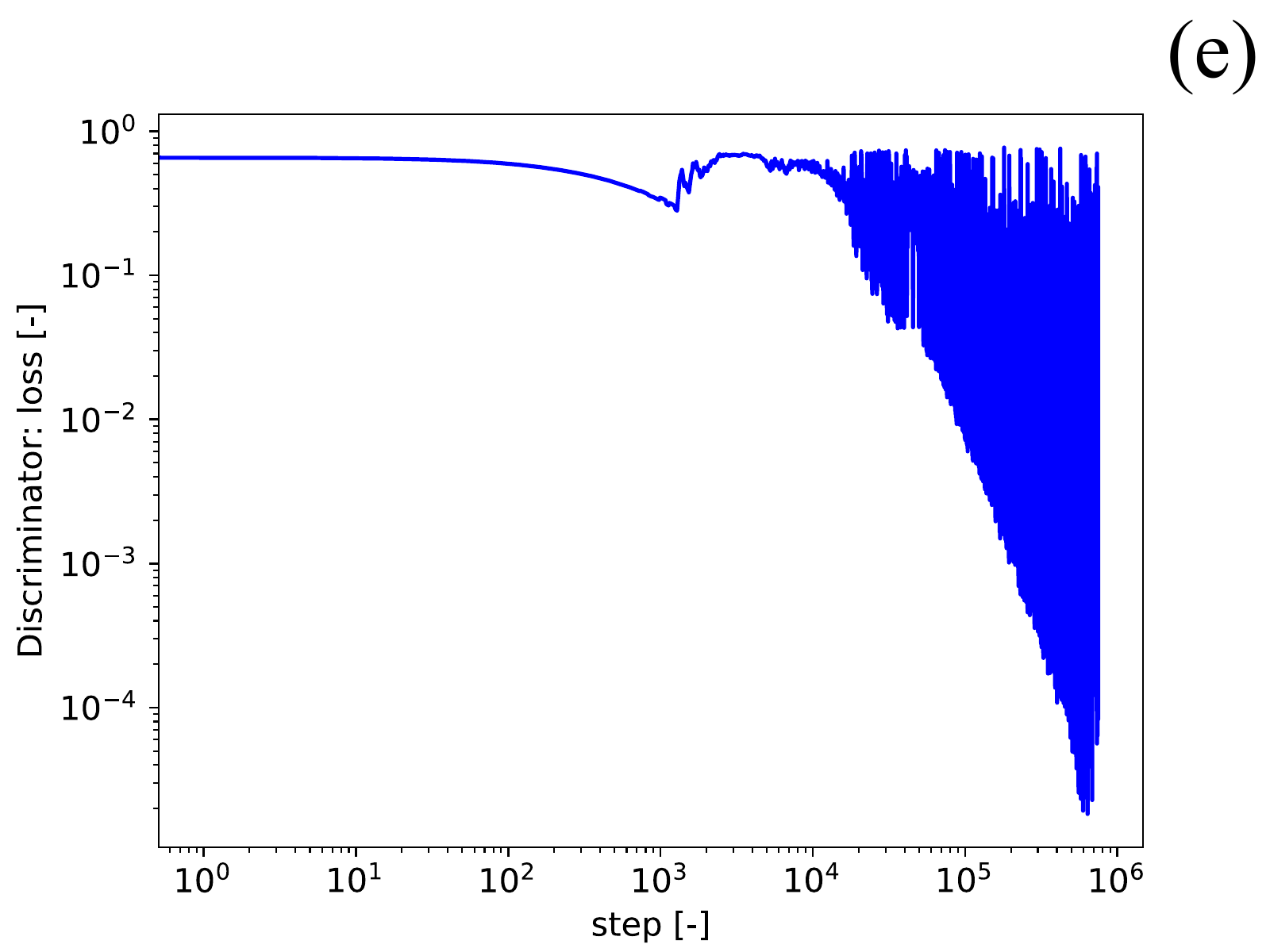}
         \includegraphics[keepaspectratio, height=5.5cm]{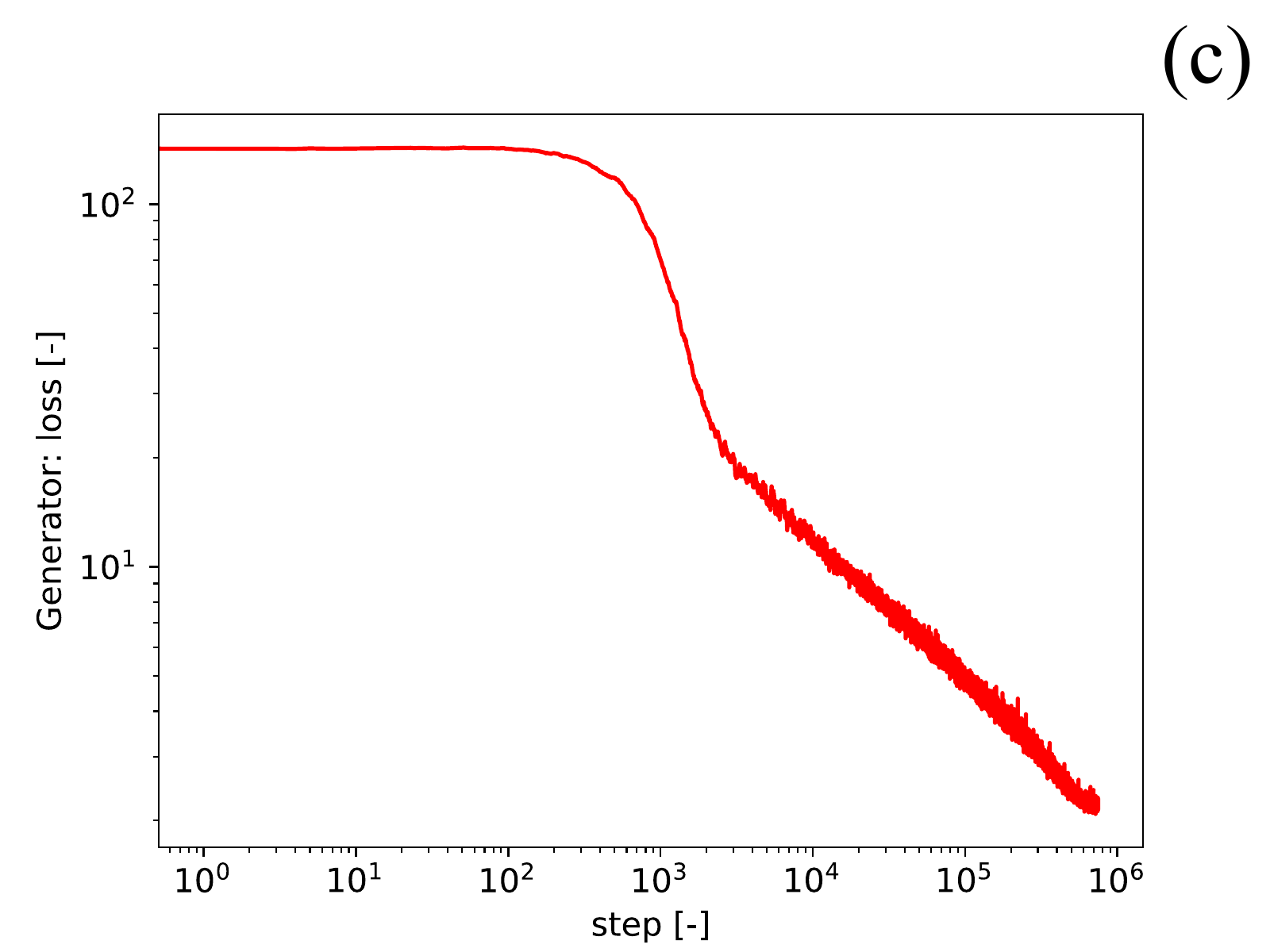}
         \includegraphics[keepaspectratio, height=5.5cm]{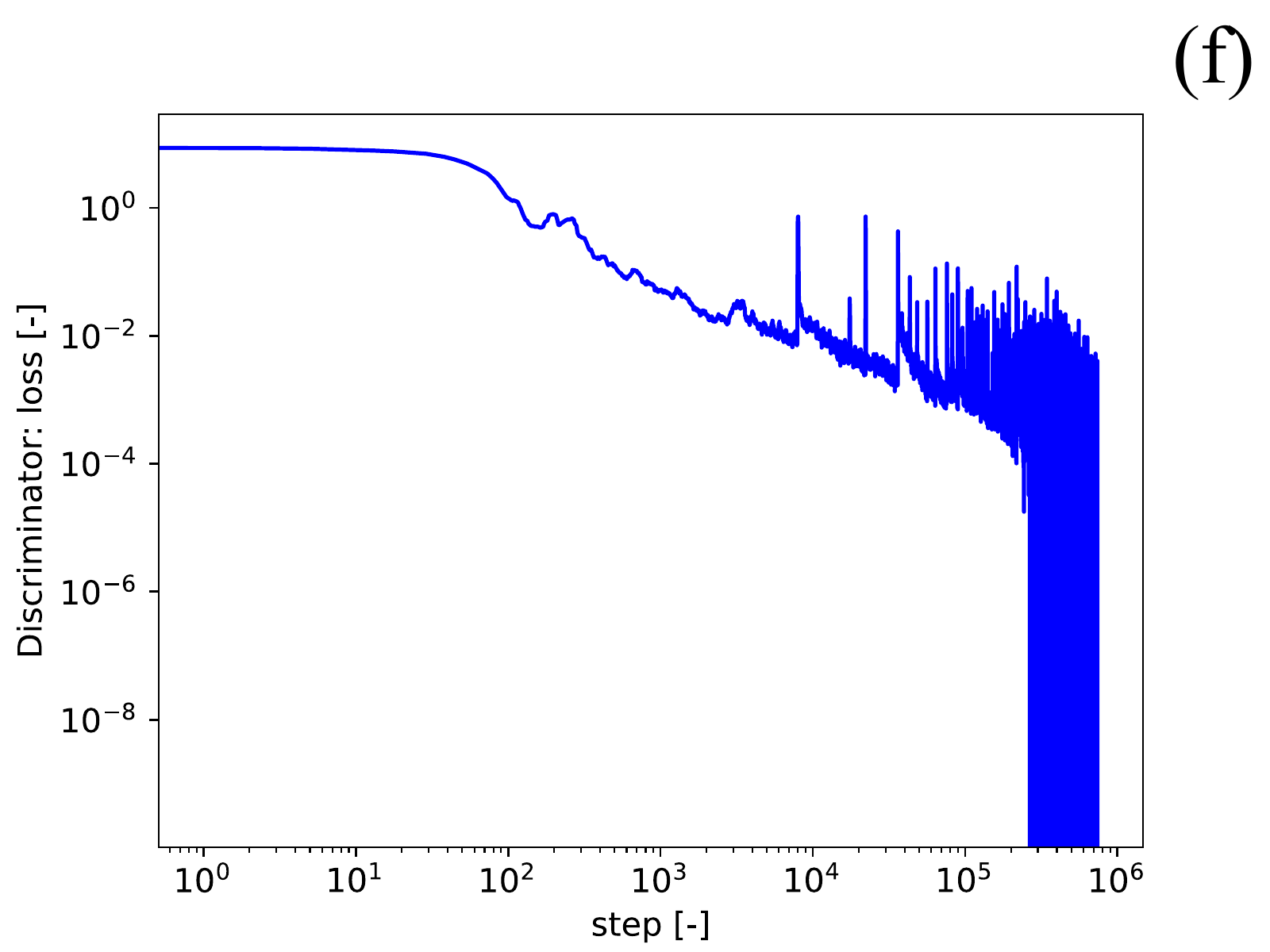}
   \caption{Example 2: 100$^{th}$ moving average of Generator loss of (a) base model, (b) SN model, and (c) W model and Discriminator loss of (d) base model, (e) SN model, and (f) W model   }
   \label{fig:ex2_gen_disc_loss}
\end{figure}

We then progress to a more challenging case of permeability fields that have a bimodal distribution. Again, we first populate 11000 realizations of a permeability field using a multivariate Gaussian distribution with a mean of log10 of permeability field is -12 \si{log(m^2)}, the variance is 1 \si{log(m^4)}, and the length-scale is 1.0 \si{m}. Subsequently, we transform the permeability fields using bimodal transformation \cite{muller2020}. Similar to the previous example, we use 10000 and 1000 examples for training and testing, respectively. The generator and discriminator losses' behaviors are illustrated in Fig. \ref{fig:ex2_gen_disc_loss}.  \par

These behaviors of the generator and discriminator losses are slightly different from ones in Fig. \ref{fig:ex1_gen_disc_loss}. From Figs. \ref{fig:ex2_gen_disc_loss}a-c, we observe that the W model is the most stable one, and in this case, the SN model's generator loss is more stable than the one of the base model. From Figs. \ref{fig:ex2_gen_disc_loss}d-f, the W model's discriminator loss is decreased steadily, while there are no significant differences between the SN model and base model's discriminator losses. Again, the W model's discriminator loss approaches zero much later than those of the base and SN models, which means the learning process continues for a longer time \cite{gulrajani2017improved,arjovsky2017wasserstein}.  \par

\begin{figure}[!ht]
   \centering

         \includegraphics[keepaspectratio, height=3.5cm]{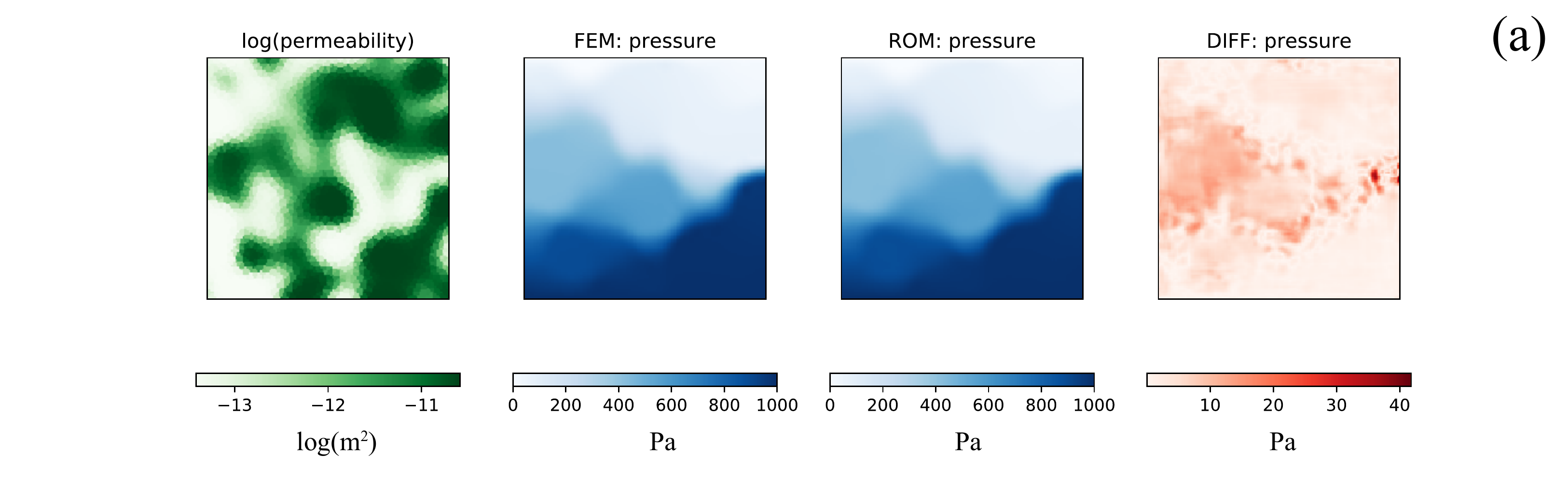}
         \includegraphics[keepaspectratio, height=3.5cm]{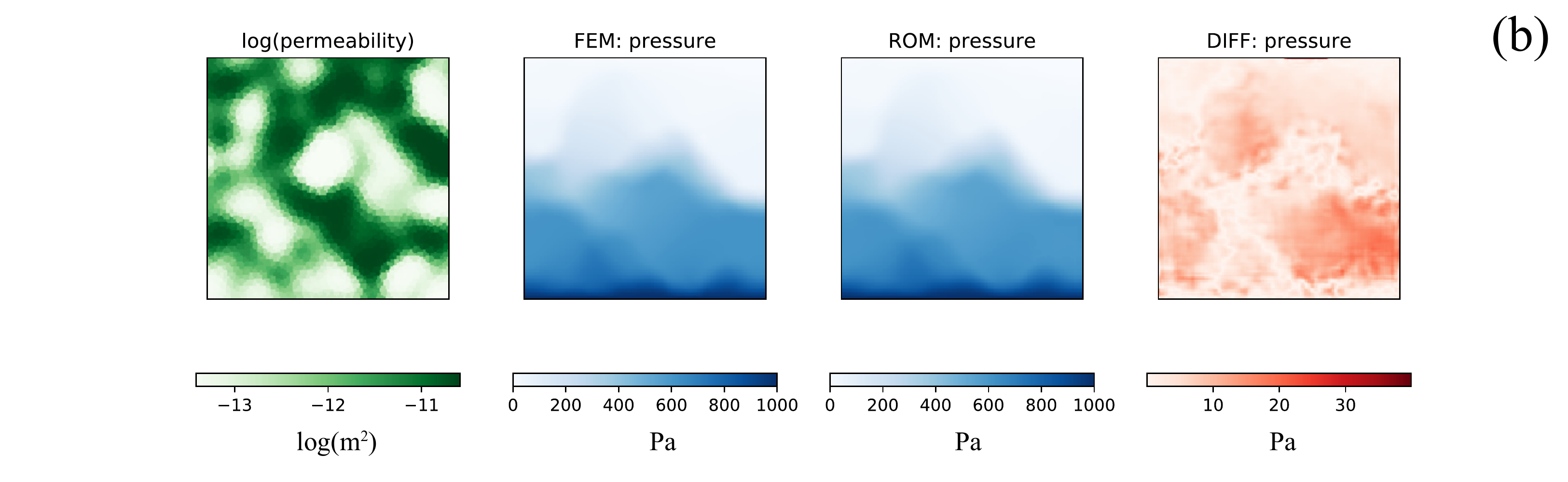}
         \includegraphics[keepaspectratio, height=3.5cm]{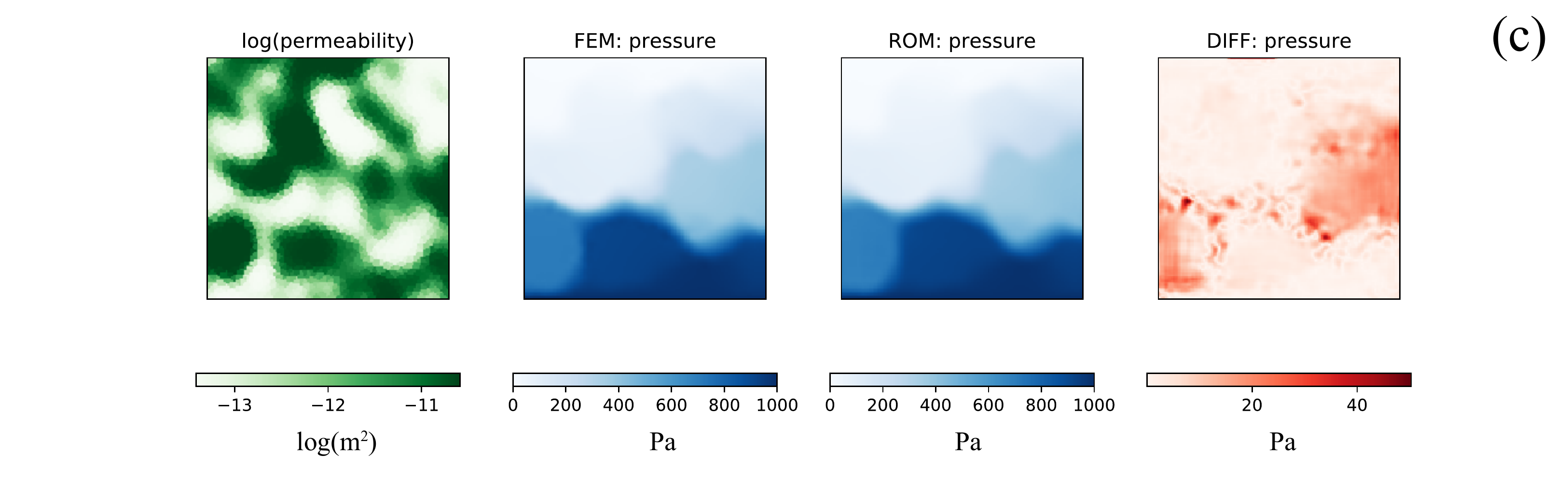}
   \caption{Example 2: test cases' results of the W model. Note that we train our three models (base, SN, and W models) using 10000 training examples and test them using 1000 test examples. These three cases shown here are randomly picked from 1000 test examples.}
   \label{fig:ex2_pic}

\end{figure}

We present three of our test cases in Fig. \ref{fig:ex2_pic}. Comparing between Figs. \ref{fig:ex1_pic} and \ref{fig:ex2_pic}, we observe that the results of the bimodal transformation example have higher errors (DIFF) as we expect. These higher errors stem from the fact that the permeability fields, in this case, contain sharper contrasts, which are harder to capture even in cases where finite element or finite volume methods are used \cite{nick2011hybrid, salinas2018discontinuous, kadeethum2020enriched}. \par

\begin{figure}[!ht]
   \centering

         \includegraphics[keepaspectratio, height=3.5cm]{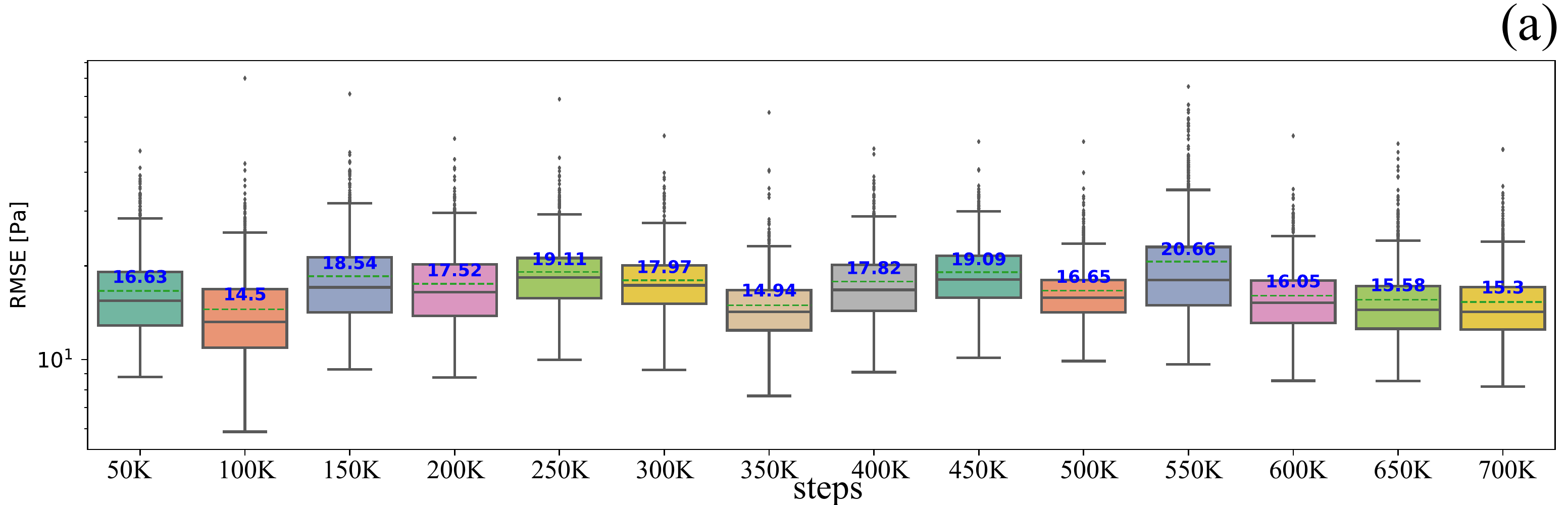}
         \includegraphics[keepaspectratio, height=3.5cm]{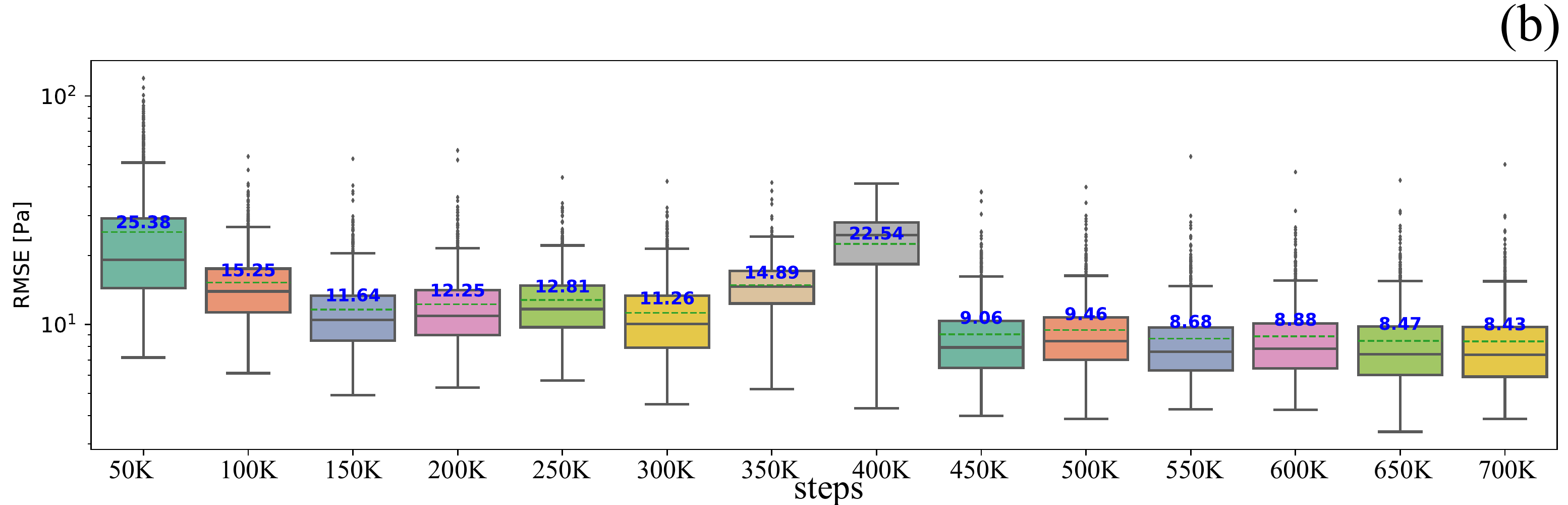}
         \includegraphics[keepaspectratio, height=3.5cm]{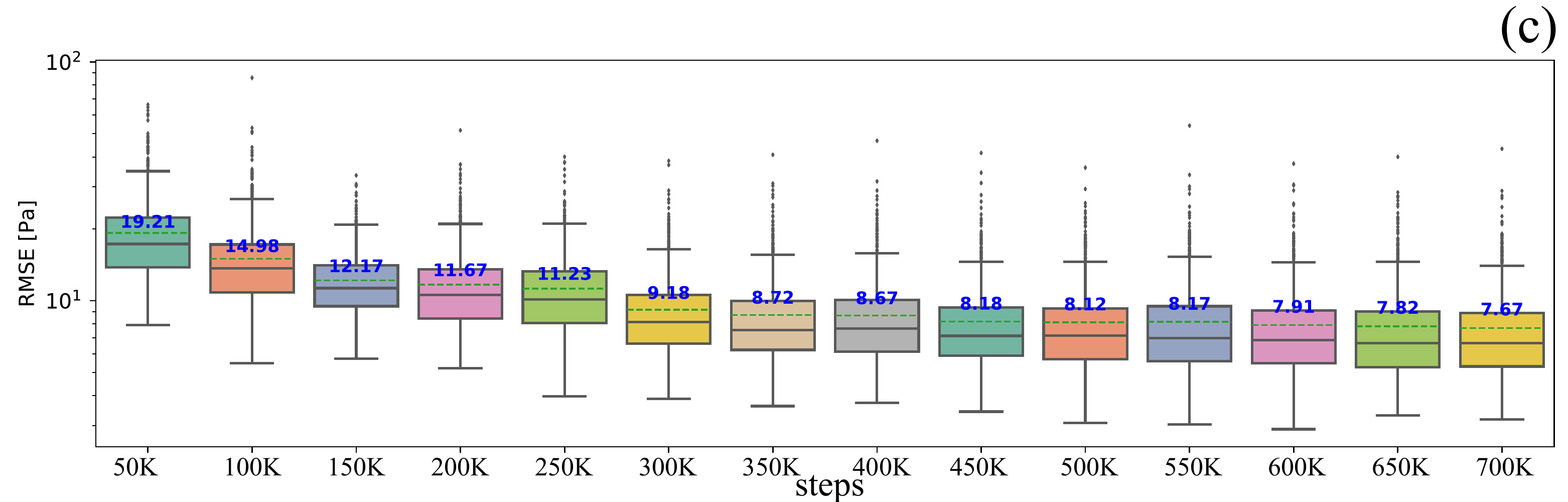}
   \caption{Example 2: Root Mean Square Error (RMSE) of (a) base model, (b) SN model, and (c) W model. Each step refers to each time we perform back-propagation including updating both generator and discriminator's parameters.}
   \label{fig:ex2_test}

\end{figure}

The RMSE results of the three models, base model, SN model, and W model, at different steps are presented in Fig. \ref{fig:ex2_test}. Similar to the previous example (Example 1), the W model provides the best accuracy and the most stable results (i.e., the RMSE of the test cases decreases as the training progresses). Furthermore, the SN model yields better RMSE results than those of the base model. As expected, the RMSE results of the permeability field as bimodal transformation example are generally higher than those of the permeability field as Gaussian distribution example.  \par 

From Fig. \ref{fig:ex2_test}, the base model has 0.06\%, 1.45\%, and 8.00\% minimum, average, and maximum RMSE values, respectively. The SN model has 0.04\%, 0.84\%, and 6.00\% minimum, average, and maximum RMSE values. The W model, which has the best performance, has 0.35\%, 0.77\%, and 6.00\% minimum, average, and maximum RMSE values, respectively.  \par


\subsubsection{Example 3: Permeability field as Zinn \& Harvey transformation}\label{sec:forward_zh}

\begin{figure}[!ht]
   \centering
        \includegraphics[keepaspectratio, height=5.5cm]{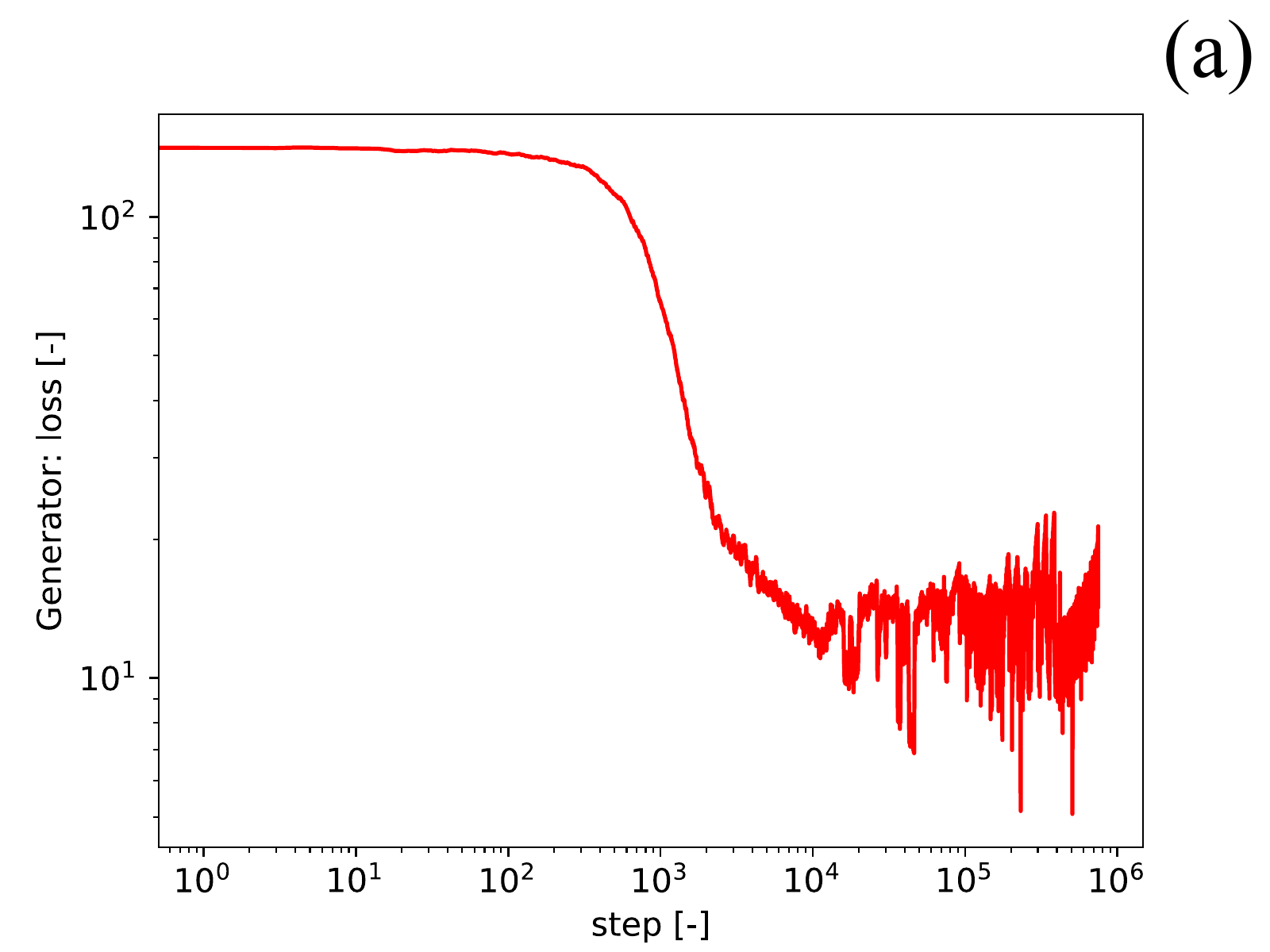}
         \includegraphics[keepaspectratio, height=5.5cm]{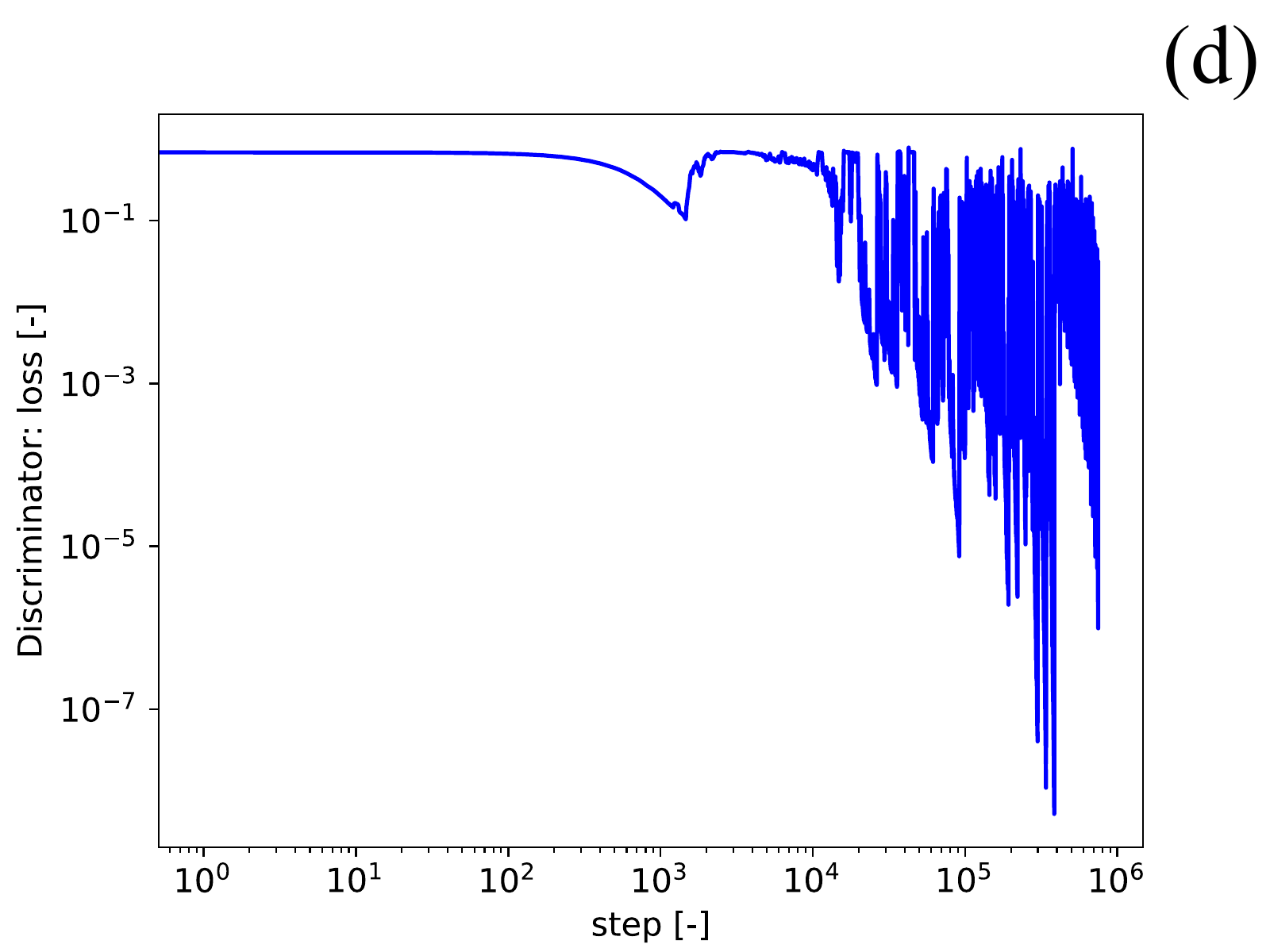}
         \includegraphics[keepaspectratio, height=5.5cm]{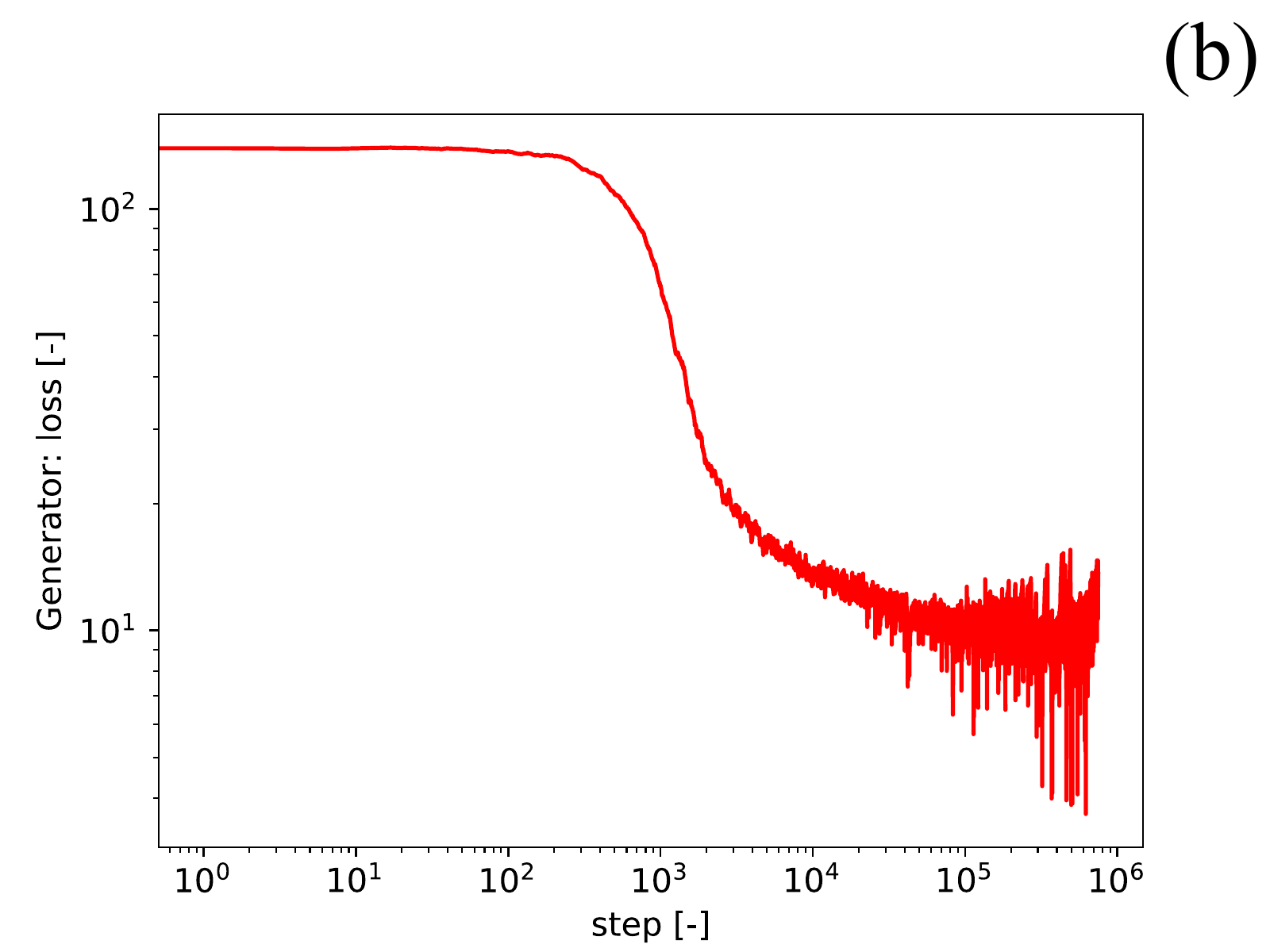}
         \includegraphics[keepaspectratio, height=5.5cm]{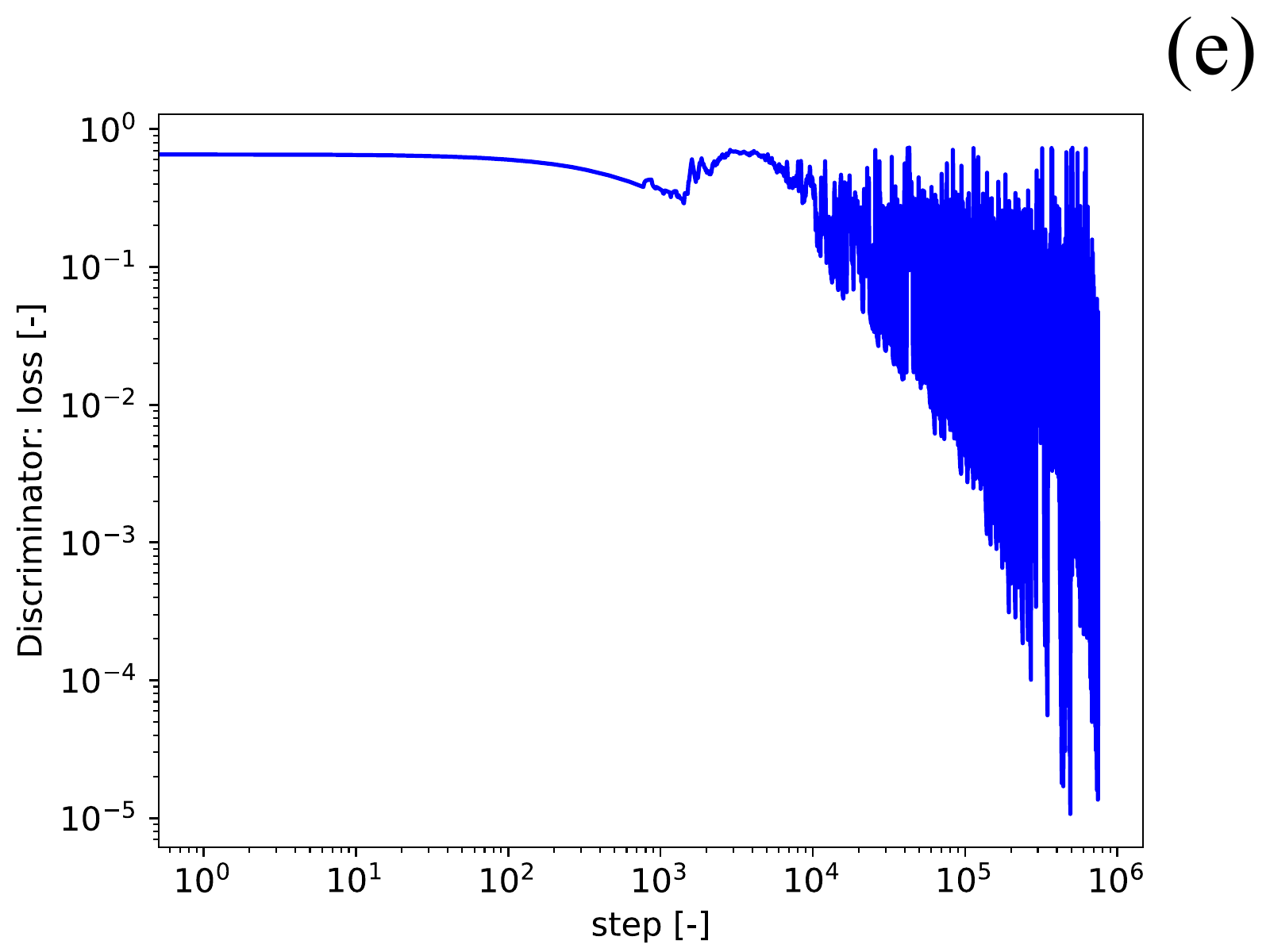}
         \includegraphics[keepaspectratio, height=5.5cm]{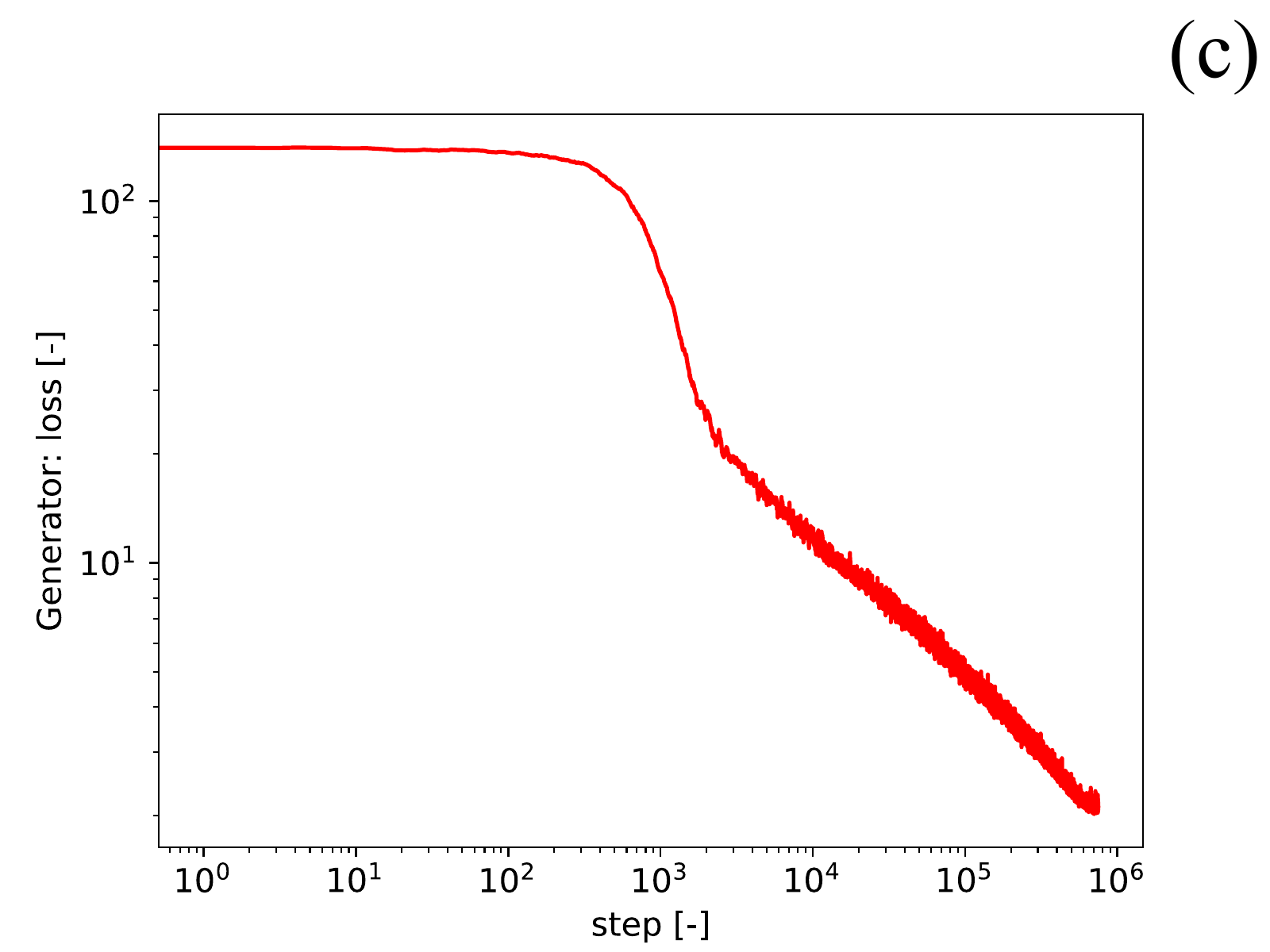}
         \includegraphics[keepaspectratio, height=5.5cm]{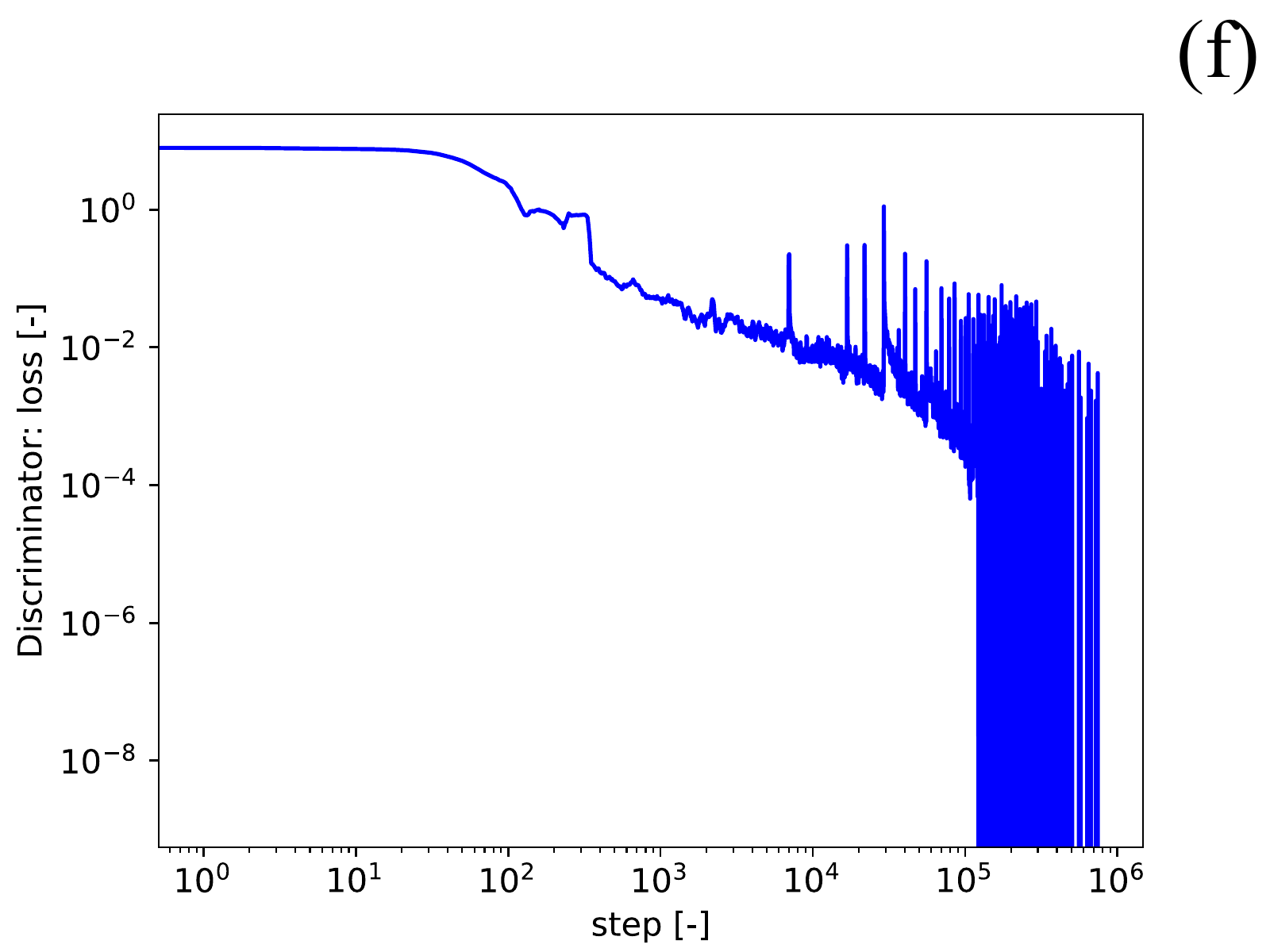}
   \caption{Example 3: 100$^{th}$ moving average of Generator loss of (a) base model, (b) SN model, and (c) W model and Discriminator loss of (d) base model, (e) SN model, and (f) W model}
   \label{fig:ex3_gen_disc_loss}
\end{figure}

For the third example, we move on to cases where there are high permeability contrasts and channels. Similar to two previous examples (Examples 1 and 2), we first populate 11000 realizations of a permeability field using an unconditional Gaussian distribution with a mean of log10 of permeability field is -12 \si{log(m^2)}, the variance is 1 \si{log(m^4)}, and the length-scale is 1.0 \si{m}. Then, we transform the generated permeability fields using Zinn \& Harvey transformation \cite{zinn2003good,muller2020} that promotes the connectivity in the high values of the original Gaussian field. To elaborate, this transformation transforms a Gaussian random field realization to a random field with enhanced connectivity (high or low values of the original realization are connected) without changing the first and second moments (mean and covariance of the random variable). Again, we populate 11000 permeability fields and use 10000 and 1000 of those fields for training and testing, respectively. The generator and discriminator losses' behaviors are illustrated in Fig. \ref{fig:ex3_gen_disc_loss}.  \par


Similar to two previous examples (Sections \ref{sec:forward_gs} and \ref{sec:forward_bm}), the W model has the most stable training behavior, see Figs \ref{fig:ex3_gen_disc_loss}a-c. The SN model exhibits a better training behavior than the base model (i.e., its generator loss shows a longer decreasing trend.). For the discriminator's loss (see Figs \ref{fig:ex3_gen_disc_loss}d-f), the W model illustrates the longest training process similar to previous examples. \par

\begin{figure}[!ht]
   \centering

         \includegraphics[keepaspectratio, height=3.5cm]{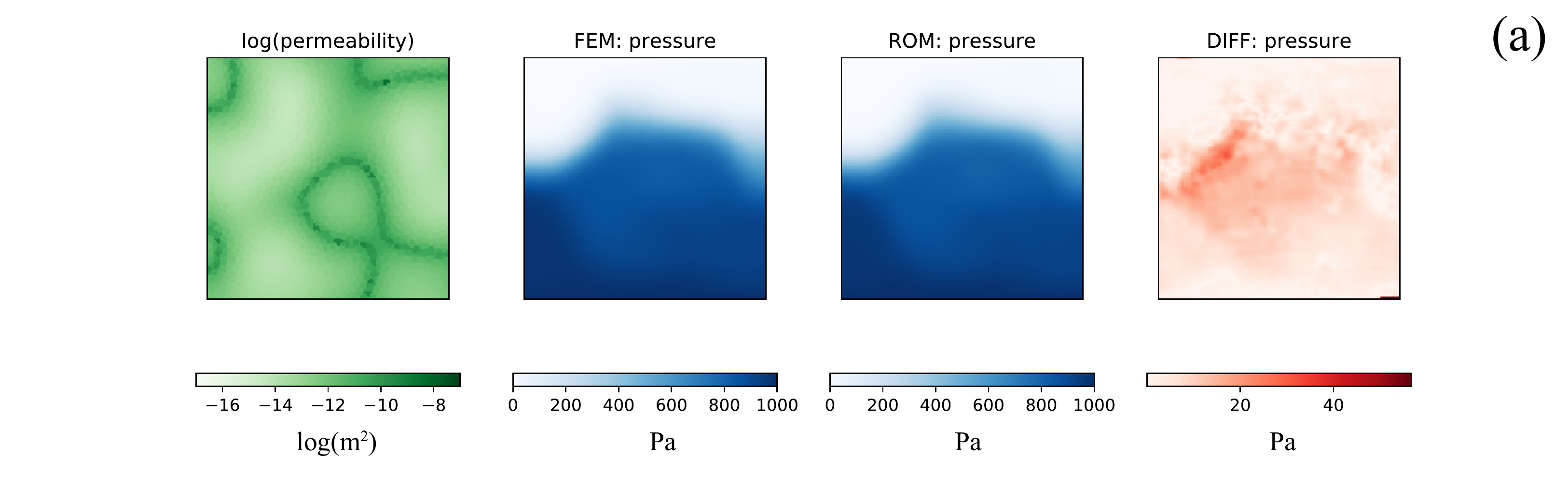}
         \includegraphics[keepaspectratio, height=3.5cm]{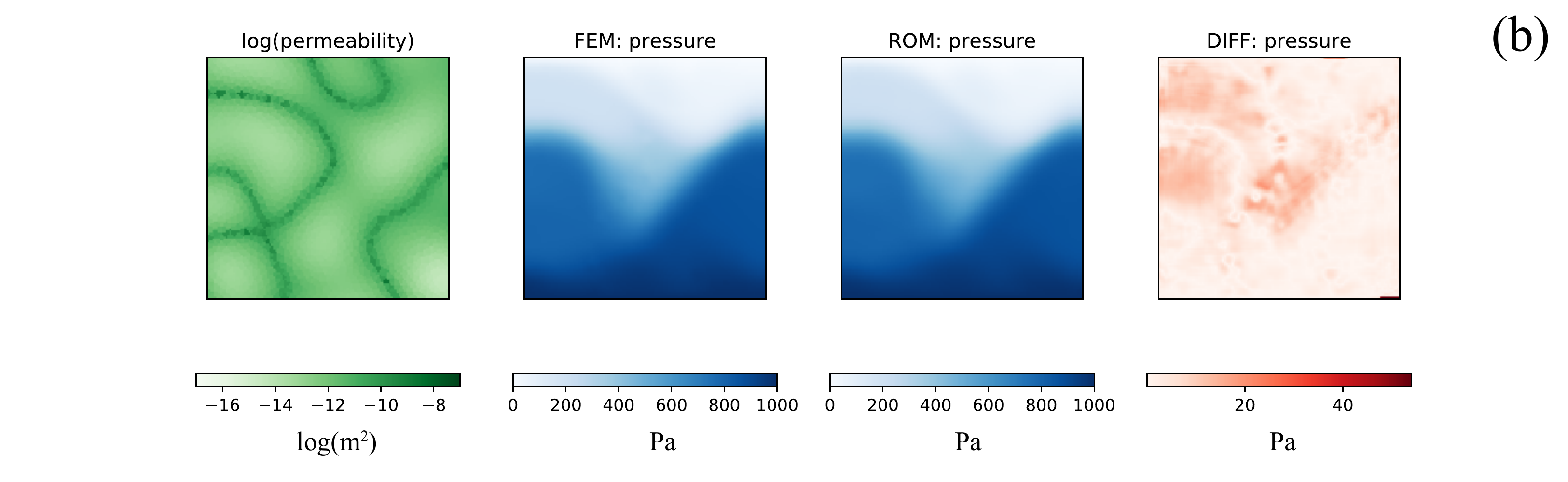}
         \includegraphics[keepaspectratio, height=3.5cm]{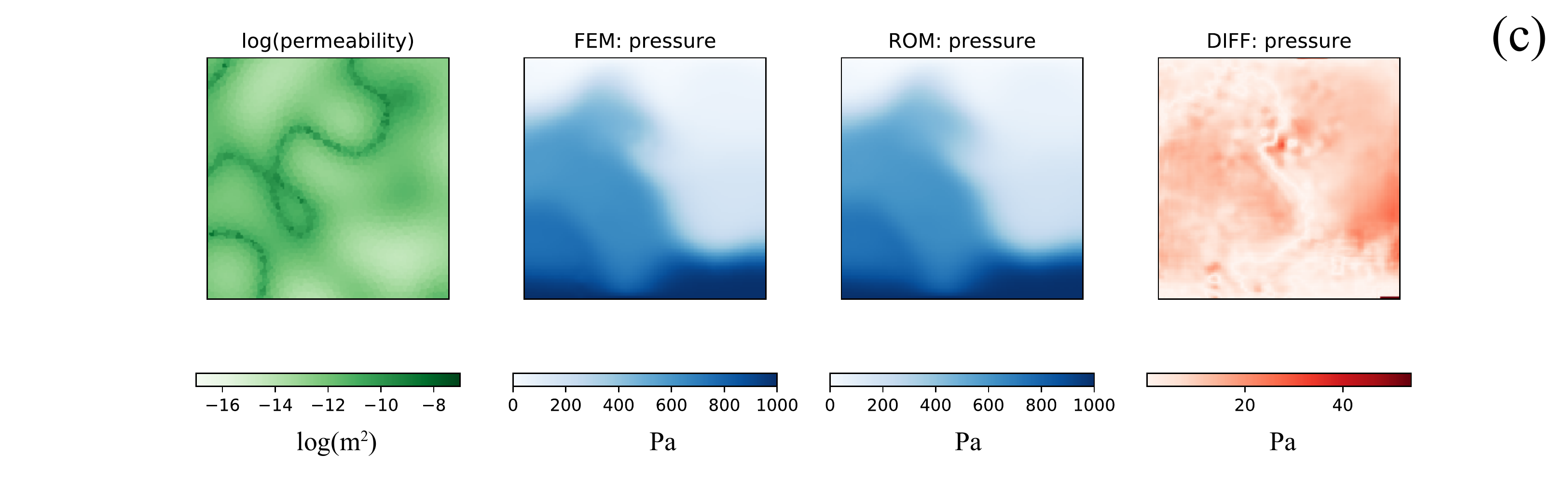}
   \caption{Example 3: test cases' results of the W model. Note that we train our three models (base, SN, and W models) using 10000 training examples and test them using 1000 test examples. These three cases shown here are randomly picked from 1000 test examples.}
   \label{fig:ex3_pic}

\end{figure}

We present three examples of the test cases in Fig. \ref{fig:ex3_pic}. We observe that the ROM model could provide a decent approximation of the FEM result. As one would expect, this example's error (DIFF) is higher than the two previous examples. However, the maximum DIFF value is still two-order of magnitude less than the maximum pressure value.  \par

\begin{figure}[!ht]
   \centering
         \includegraphics[keepaspectratio, height=3.5cm]{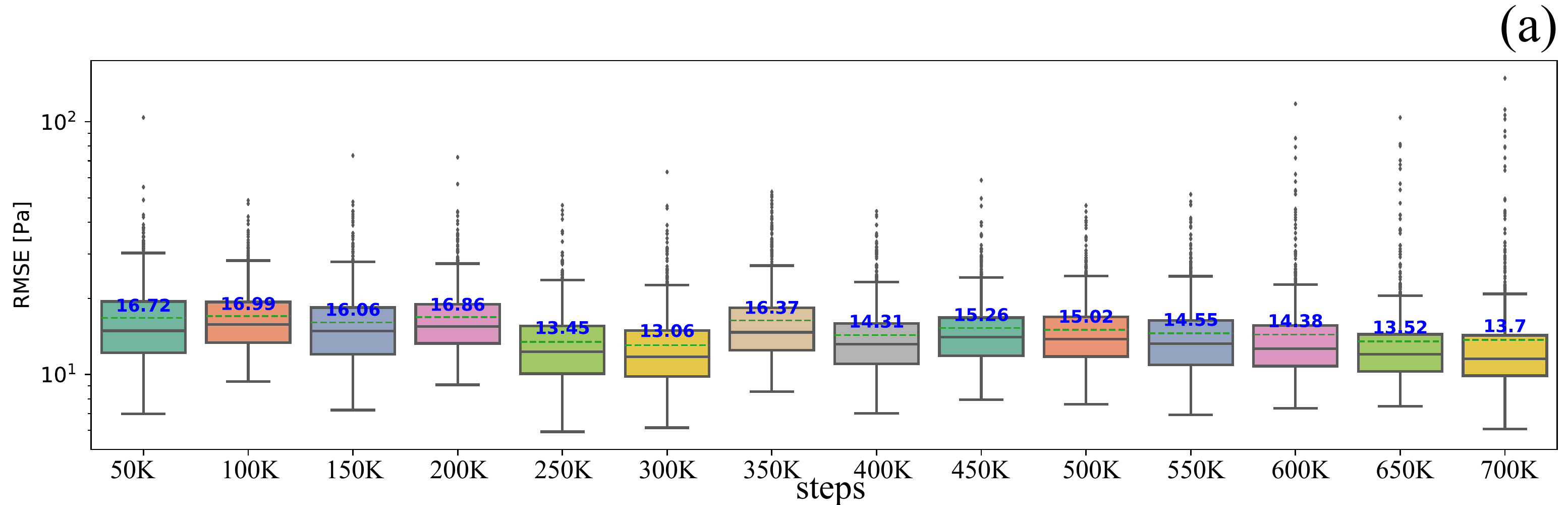}
         \includegraphics[keepaspectratio, height=3.5cm]{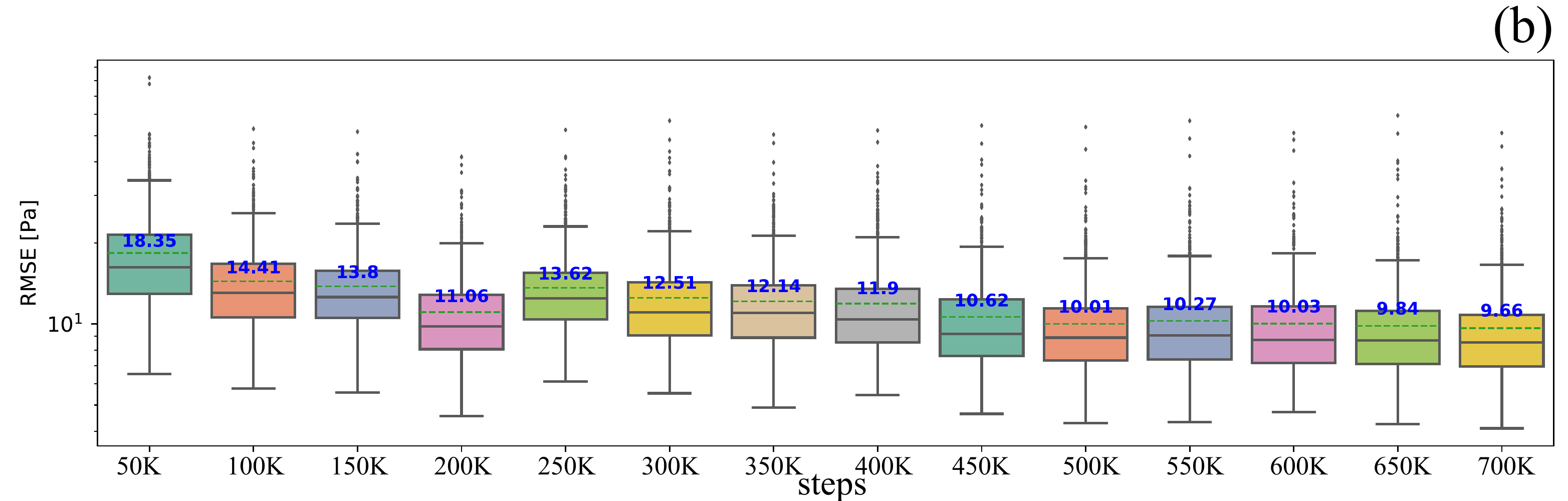}
         \includegraphics[keepaspectratio, height=3.5cm]{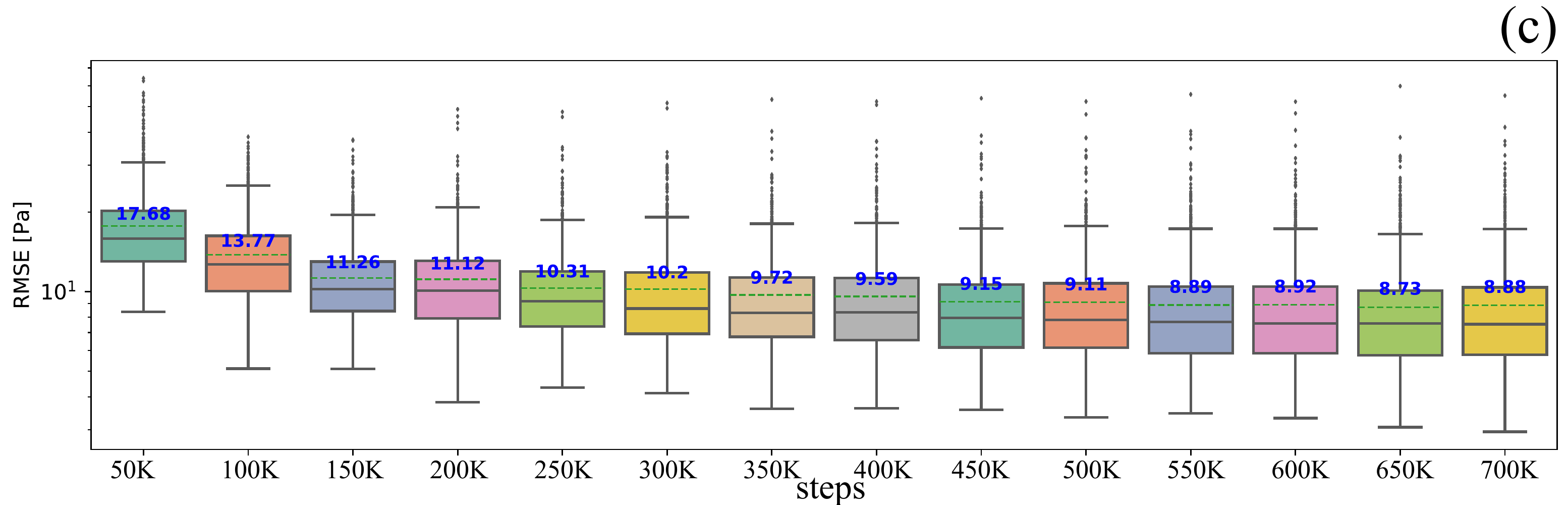}
   \caption{Example 3: Root Mean Square Error (RMSE) of (a) base model, (b) SN model, and (c) W model. Each step refers to each time we perform back-propagation including updating both generator and discriminator's parameters.}
   \label{fig:ex3_test}

\end{figure}

The RMSE results of this example are presented in Fig. \ref{fig:ex3_test}. Among three examples (Examples 1, 2, and 3), this example generally has the highest RMSE values for all three models (base, SN, and W models). Similar to two previous examples, the W model exhibits the most stable behavior (i.e., the RMSE average decreases as the training progress as well as the RMSE average is the lowest among the three models). The SN model typically has a better performance than the base model (see Fig. \ref{fig:ex3_test}a-b).   \par 

\subsection{Supplementary inverse modeling}\label{sec:sup_inverse}

As presented in the forward modeling sections, the W model exhibits the most stable training behavior as well as the testing accuracy; hence, throughout this section, we focus only on the results of the W model. Moreover, we use only Zinn \& Harvey transformation permeability fields. \par

\subsubsection{Example 4: input and output fields have the same resolution}\label{sec:inverse_full}

\begin{figure}[!ht]
   \centering
        \includegraphics[keepaspectratio, height=5.5cm]{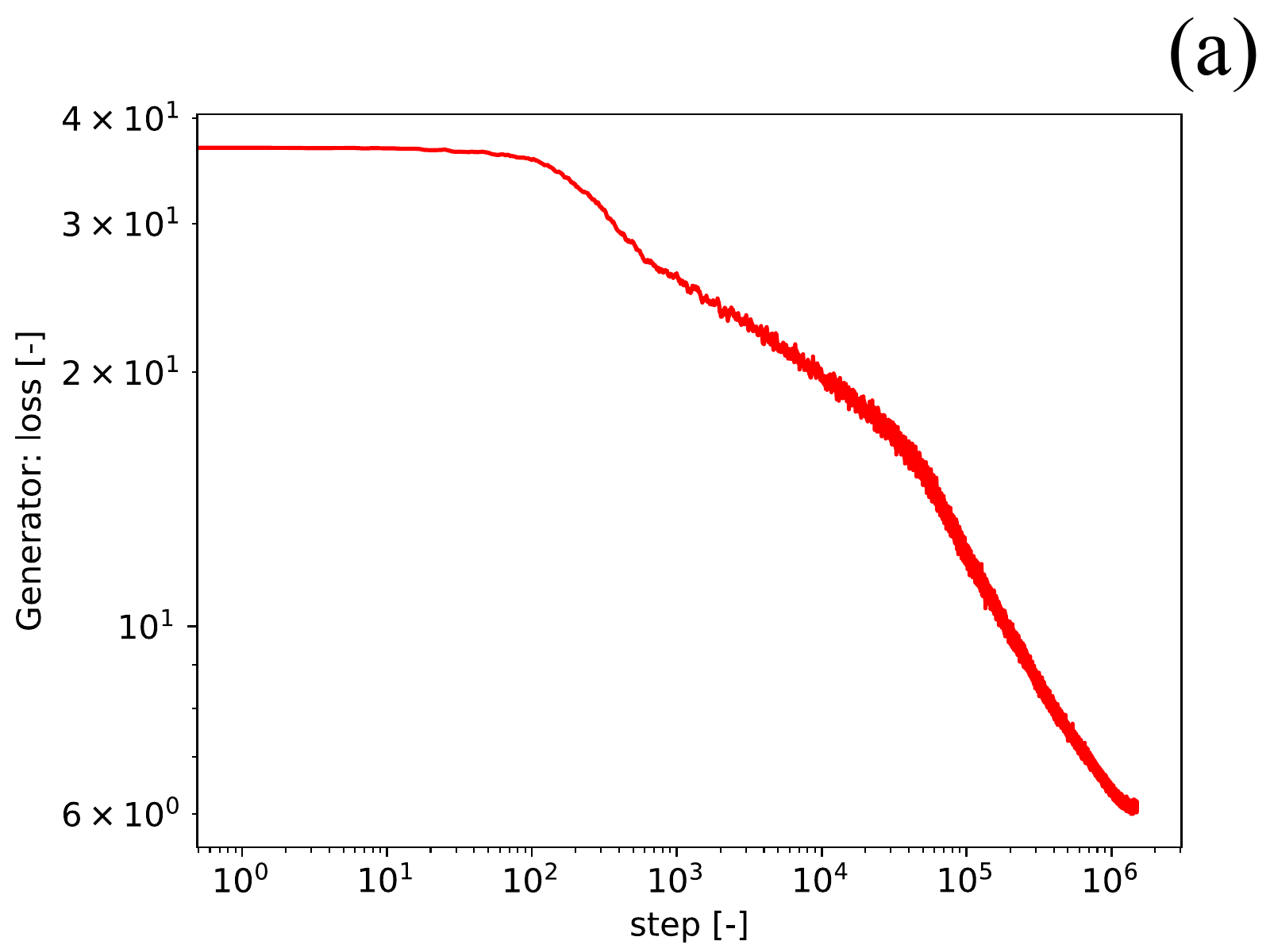}
         \includegraphics[keepaspectratio, height=5.5cm]{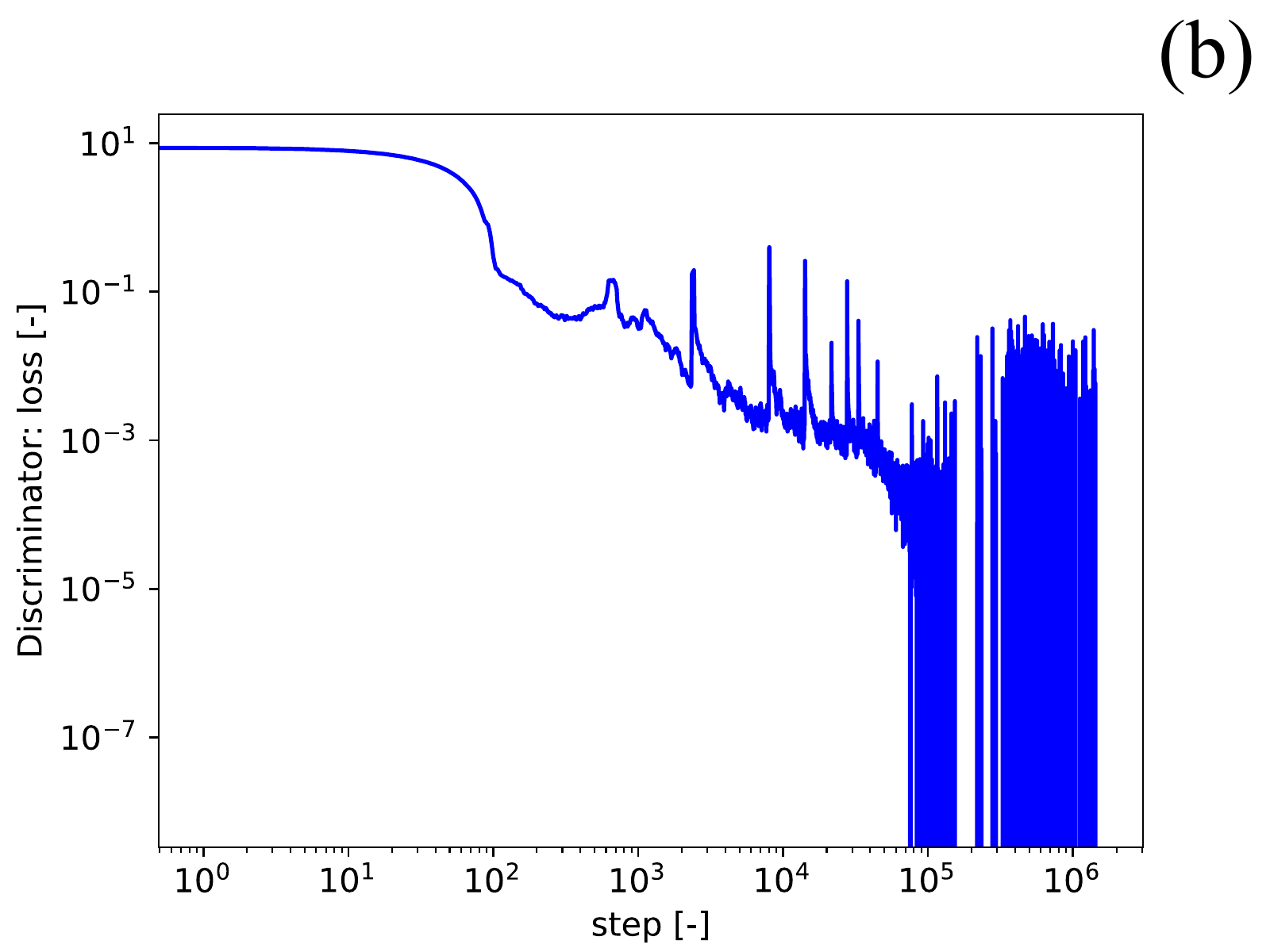}
   \caption{Example 4: 100$^{th}$ moving average of Generator and Discriminator losses of W model}
   \label{fig:ex5_gen_disc_loss}
\end{figure}

Throughout this example, the ROM's input and output fields have the same resolution (i.e., $128 \times 128$). The generator and discriminator's training losses are presented in Fig. \ref{fig:ex5_gen_disc_loss}. Both of losses are converging to zero, which show the training stability of the W model. \par 

\begin{figure}[!ht]
   \centering
         \includegraphics[keepaspectratio, height=4.8cm]{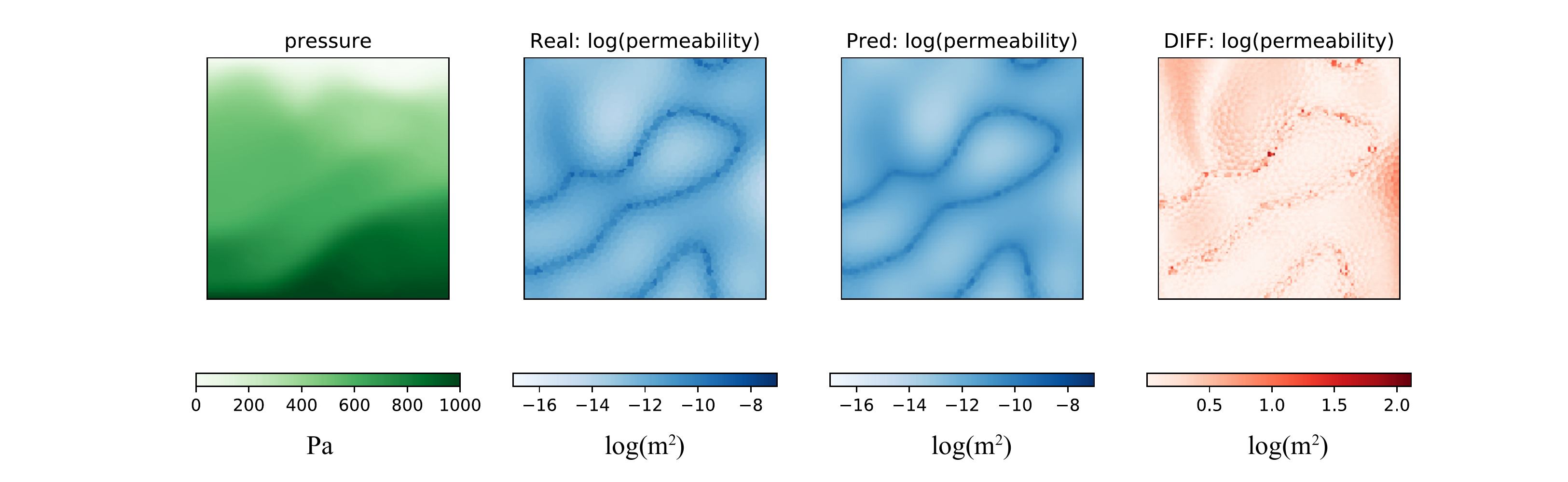}
   \caption{Example 4: test cases' result of the W model. Note that we train our model using 10000 training examples and test them using 1000 test examples. This case shown here is randomly picked from 1000 test examples.}
   \label{fig:ex5_pic}
\end{figure}

The test case example is shown in Fig. \ref{fig:ex5_pic}. From this figure, we observe that with a given pressure field (the results with given pressure, displacement in the x-direction, and displacement in y-direction fields yield approximately similar results), the ROM framework could approximate the permeability field with decent accuracy (see DIFF value). \par

\begin{figure}[!ht]
   \centering
         \includegraphics[keepaspectratio, height=3.5cm]{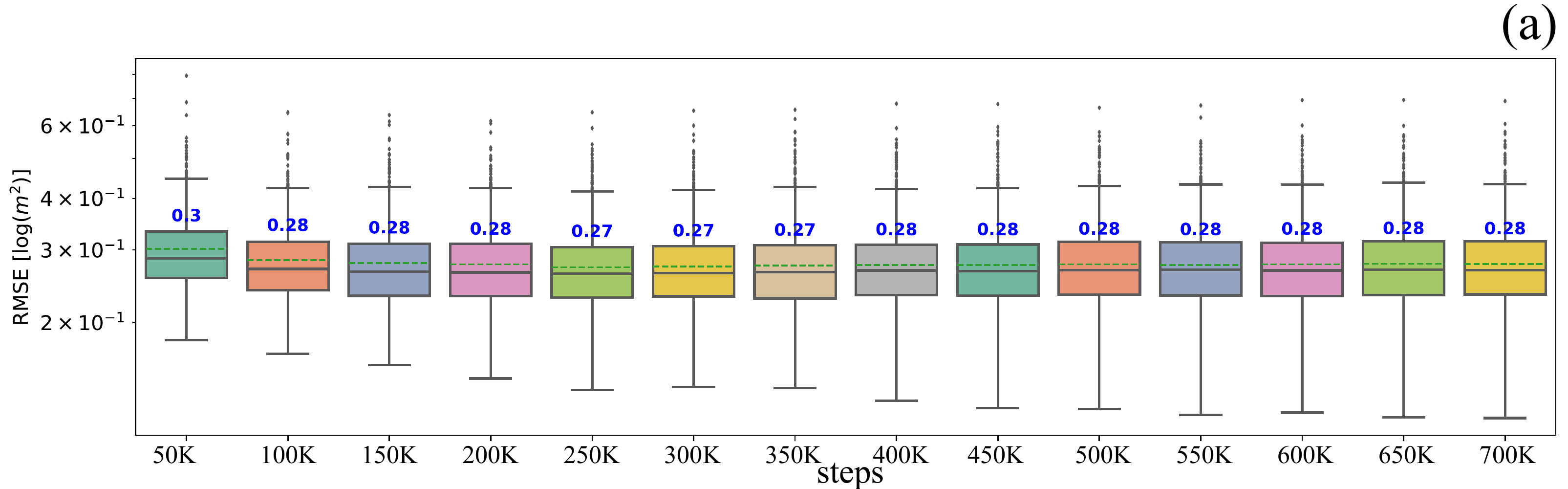}
         \includegraphics[keepaspectratio, height=3.5cm]{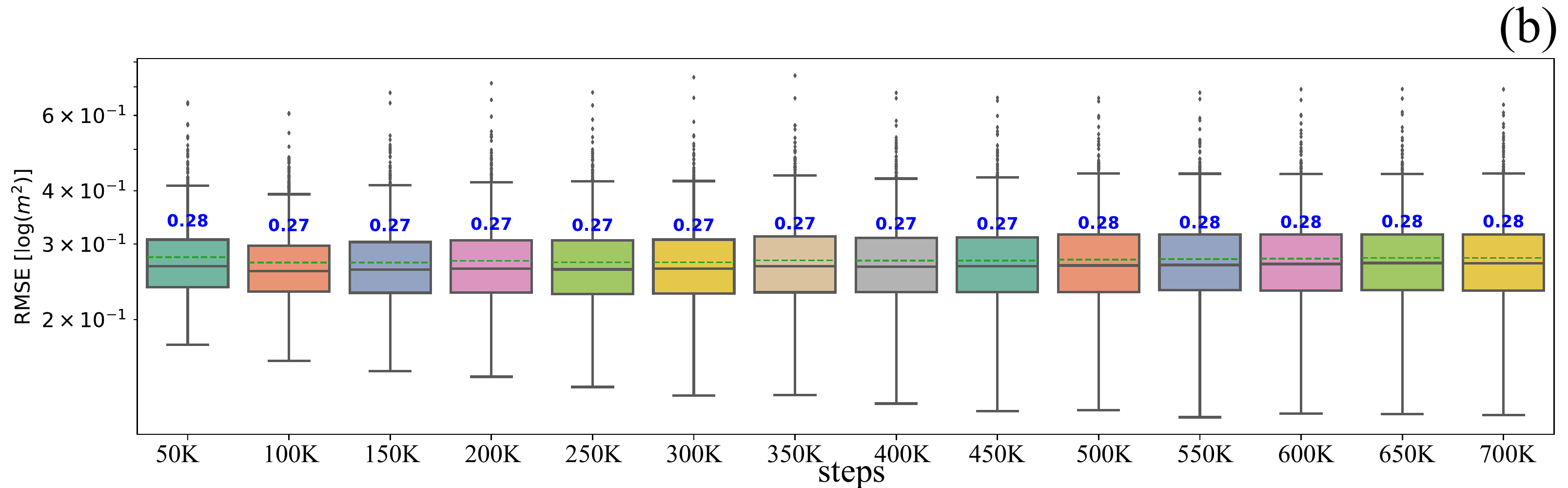}
   \caption{Example 4: Root Mean Square Error (RMSE) of W model using (a) pressure as input and (b) pressure and displacement as input. The number of training examples is 10000. Each step refers to each time we perform back-propagation including updating both generator and discriminator's parameters.}
   \label{fig:ex5_test}

\end{figure}

The RMSE results with different W model steps are presented in Fig. \ref{fig:ex5_test}. Fig. \ref{fig:ex5_test}a represents the ROM with a pressure field as input, and Fig. \ref{fig:ex5_test}b represents the ROM with pressure and displacement fields as input. We could observe that the case where we use pressure and displacement fields as input has slightly better results. The maximum RMSE values are 0.3 and 0.28 log(\si{m^2}) for the first and the second cases, respectively (the absolute value of the magnitude of log(permeability) is in the range of [8, 12]). The minimum RMSE values of the two cases are similar. This reflects the fact that as the ROM framework receives more information, its accuracy improves. However, there is not much difference between these two cases. \par 

\subsubsection{Example 5: using 75\% of input fields - uniformly vs. randomly removed }\label{sec:inverse_random_vs_uniform}

From Example 4, we observe that the ROM framework could approximate the permeability field with given pressure or pressure and displacement fields with decent accuracy (i.e., the maximum RMSE values are one- to two-ordered magnitude less than the log of the field values). This example aims to illustrate cases where the input fields have incomplete data (i.e., the input size is still $128 \times 128$; however, 25\% of these values have no measurements.). Note that we represent no measurements data using a flag of -1 (-1000 in a real value); as mentioned in Section \ref{sec:cgan}, we normalize our data for pressure, displacement, and permeability to [0, 1]. \par

\begin{figure}[!ht]
   \centering
         \includegraphics[keepaspectratio, height=3.5cm]{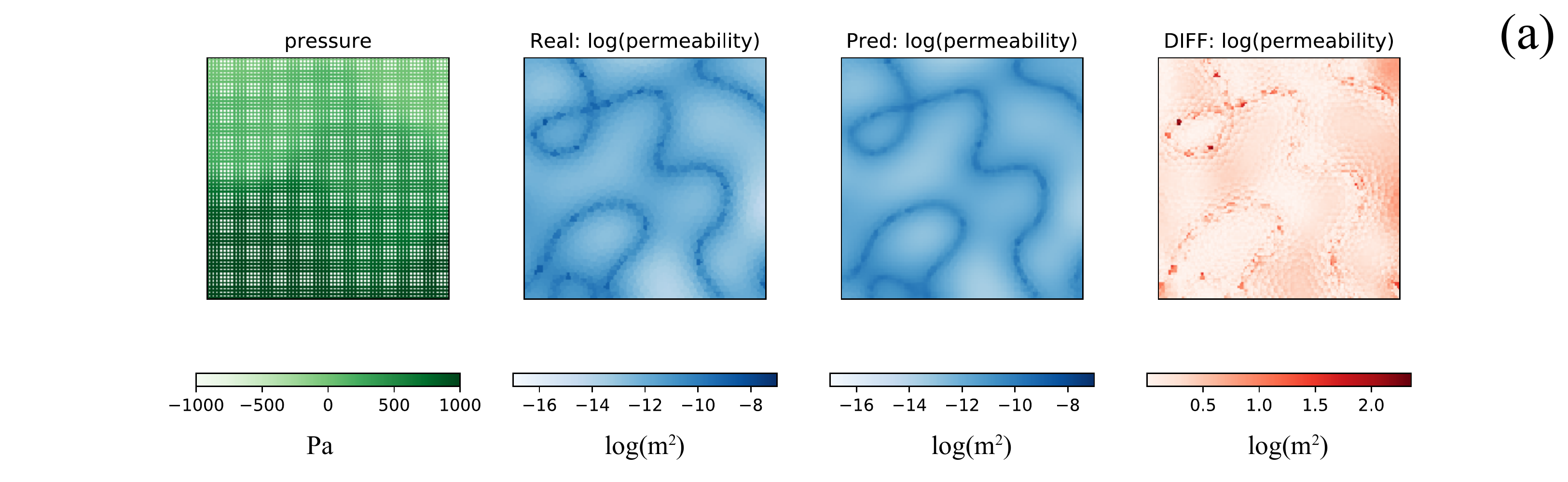}
         \includegraphics[keepaspectratio, height=3.5cm]{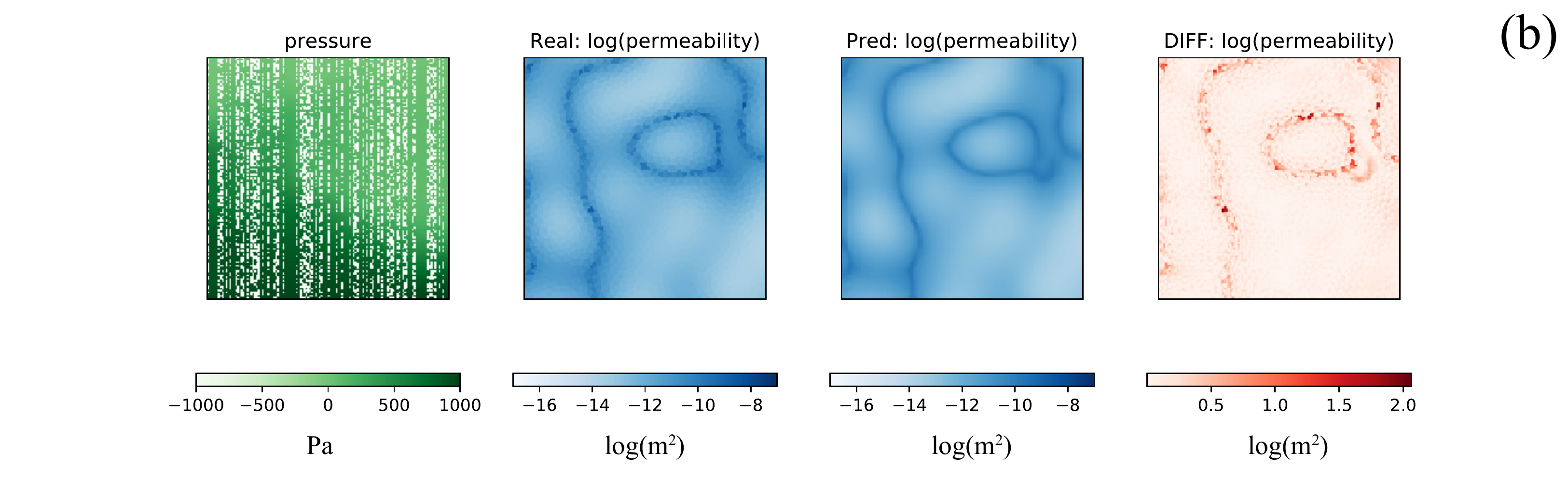}
   \caption{Example 5: test cases' results of the W model of (a) Example 5.1 - using 75\% of input fields - uniformly removed and (b) Example 5.2 - using 75\% of input fields - randomly removed. Note that we train our model using 10000 training examples and test them using 1000 test examples. Each case shown here is randomly picked from 1000 test examples.}
   \label{fig:ex6_pic}
\end{figure}

The examples of test cases are presented in Figs. \ref{fig:ex6_pic}a-b for 25\% of the input data is uniformly removed, and 25\% of the input data is randomly removed, respectively. We see that the second case's DIFF values, Fig. \ref{fig:ex6_pic}b, are generally higher than those of the first case, Fig. \ref{fig:ex6_pic}a. Still, the maximum RMSE values of these two cases are two-ordered magnitude less than the log of the permeability field values. \par

\noindent
\emph{Example 5.1: using 75\% of input fields - uniformly removed }\label{sec:inverse_uniform}

\begin{figure}[!ht]
   \centering
         \includegraphics[keepaspectratio, height=3.5cm]{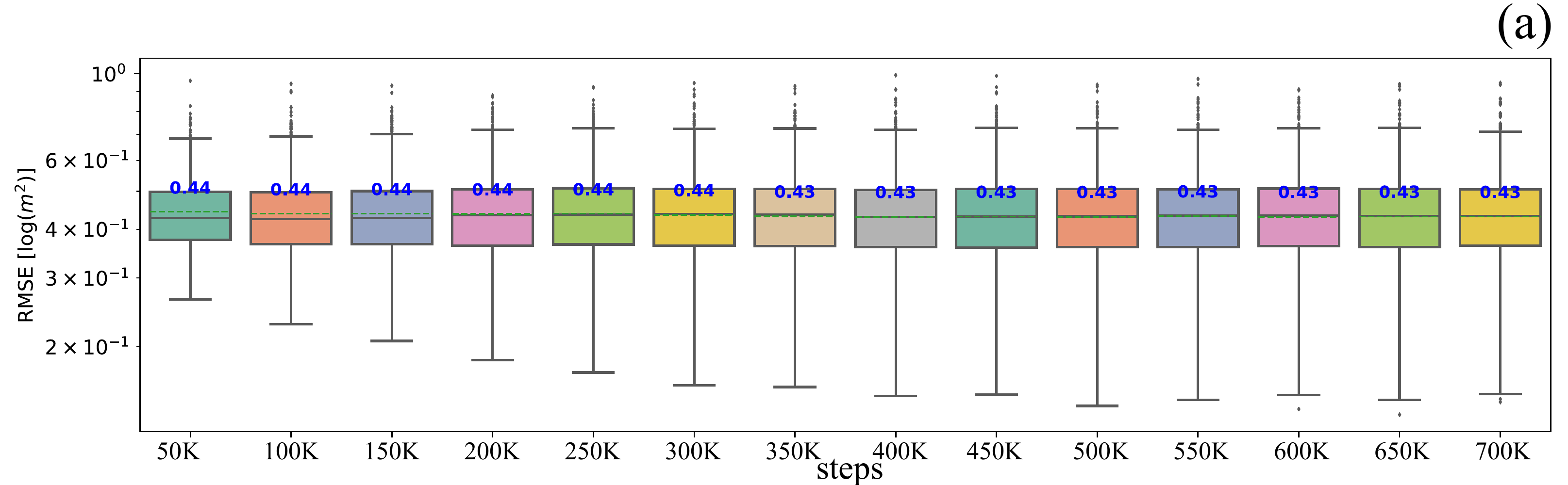}
         \includegraphics[keepaspectratio, height=3.5cm]{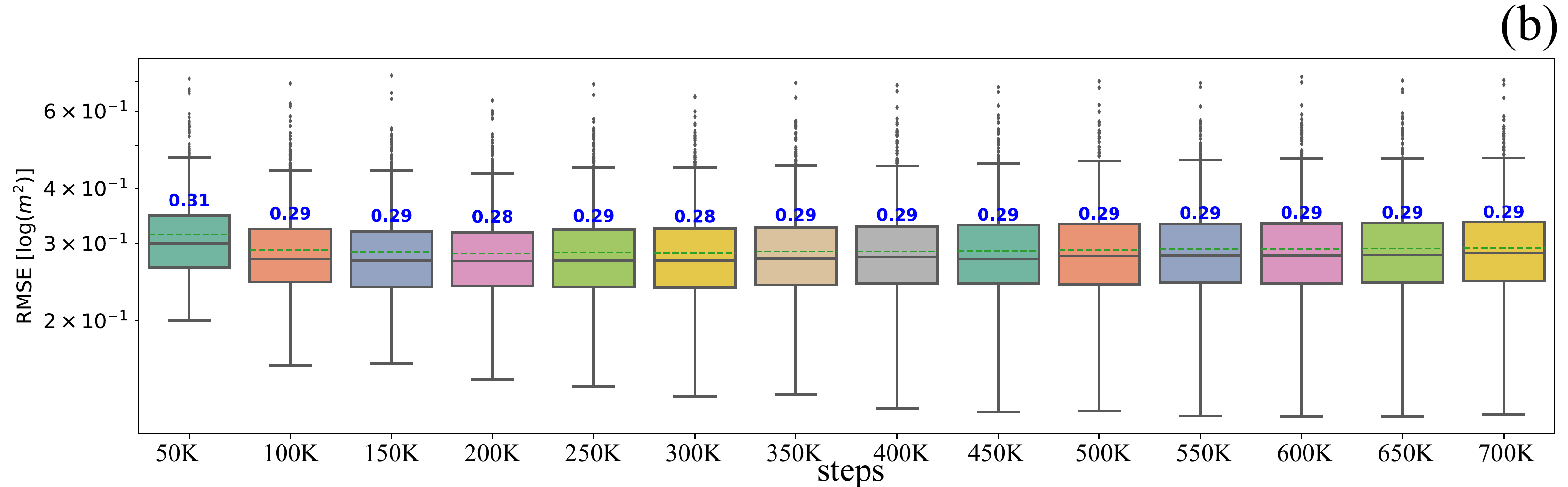}
   \caption{Example 5.1: Root Mean Square Error (RMSE) of W model using (a) pressure as input and (b) pressure and displacement as input. The number of training examples is 10000. Each step refers to each time we perform back-propagation including updating both generator and discriminator's parameters.}
   \label{fig:ex6_1_test}

\end{figure}

The RMSE results of the uniformly removed case are presented in Fig. \ref{fig:ex6_1_test}. Compared to the previous example, Example 4, the RMSE values of this example are generally higher. Besides, there are differences between using only pressure as an input (case 1) and using pressure and displacement as input (case 2), see Figs. \ref{fig:ex6_1_test}a-b. To elaborate, the maximum of RMSE average values are 0.44 and 0.31 log(\si{m^2}), see blue texts in the Fig. \ref{fig:ex6_1_test}, for case 1 and case 2, respectively. the minimum of RMSE average values are 0.43 and 0.29 log(\si{m^2}) for case 1 and case 2, respectively. Both models seem to be improved as the number of steps are increased.  \par

\noindent
\emph{Example 5.2: using 75\% of input fields - randomly removed }\label{sec:inverse_random}

\begin{figure}[!ht]
   \centering
         \includegraphics[keepaspectratio, height=3.5cm]{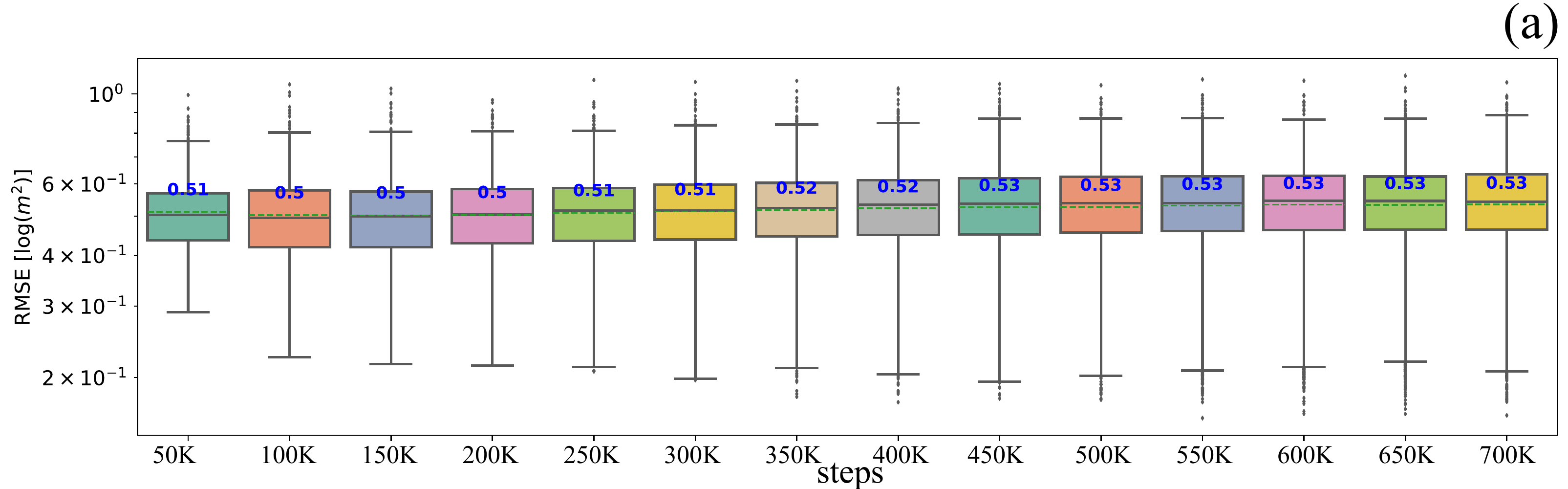}
         \includegraphics[keepaspectratio, height=3.5cm]{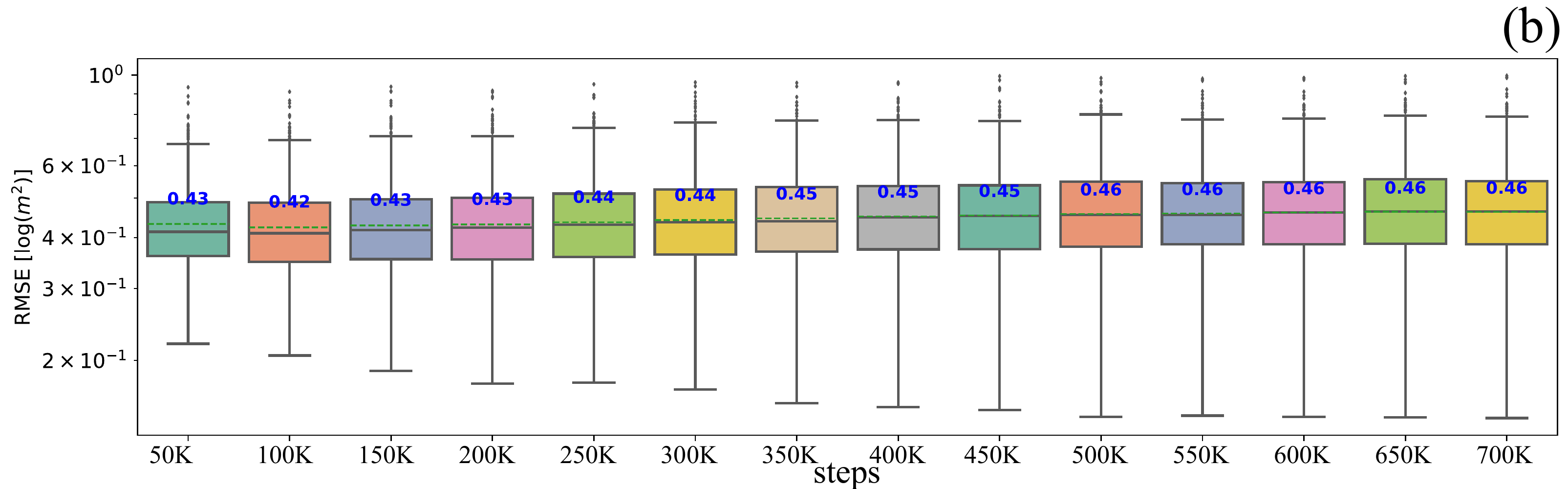}
   \caption{Example 5.2: Root Mean Square Error (RMSE) of W model using (a) pressure as input and (b) pressure and displacement as input. The number of training examples is 10000. Each step refers to each time we perform back-propagation including updating both generator and discriminator's parameters.}
   \label{fig:ex6_2_test}

\end{figure}

The RMSE behaviors of the randomly removed data points of using only pressure as an input (case 1) and using pressure and displacement as input (case 2) are illustrated in Fig. \ref{fig:ex6_2_test}. The maximum of RMSE average values are 0.53 and 0.46 log(\si{m^2}) for case 1 and case 2, respectively, see blue texts in the Fig. \ref{fig:ex6_2_test}. The minimum of RMSE average values are 0.50 and 0.42 log(\si{m^2}) for case 1 and case 2, respectively. Again, we observe that by using more input (i.e., case 2), the model's accuracy is improved. \par 

Even though the number of available data is similar to Example 5.1 (i.e., 75\% of the input fields), the RMSE values of this example are significantly higher than Example 5.1, see Figs. \ref{fig:ex6_1_test} and \ref{fig:ex6_2_test}. This behavior reflects the fact that the available data's location could impact the model's accuracy substantially. Furthermore, we observe that the model exhibits overfitting behavior (i.e., the test cases' accuracy reduces as the number of steps increases.).  \par

\subsubsection{Example 6: resolution of input fields' effect on model dynamics - randomly removed }\label{sec:inverse_random_input_size}

\begin{figure}[!ht]
   \centering
         \includegraphics[keepaspectratio, height=3.5cm]{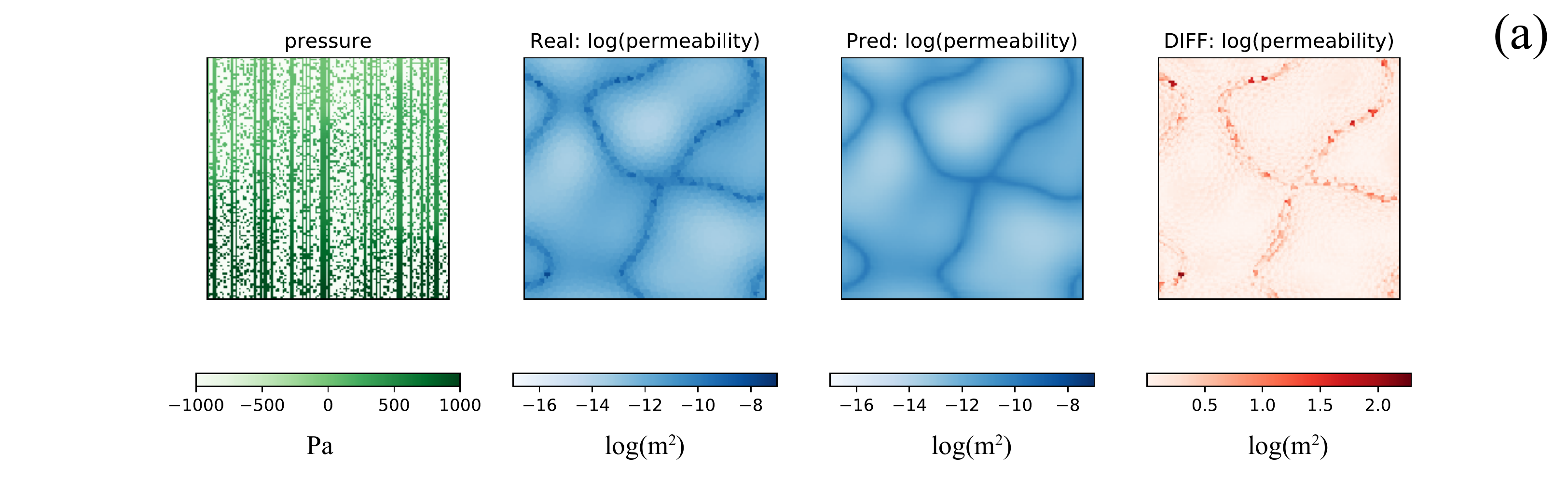}
         \includegraphics[keepaspectratio, height=3.5cm]{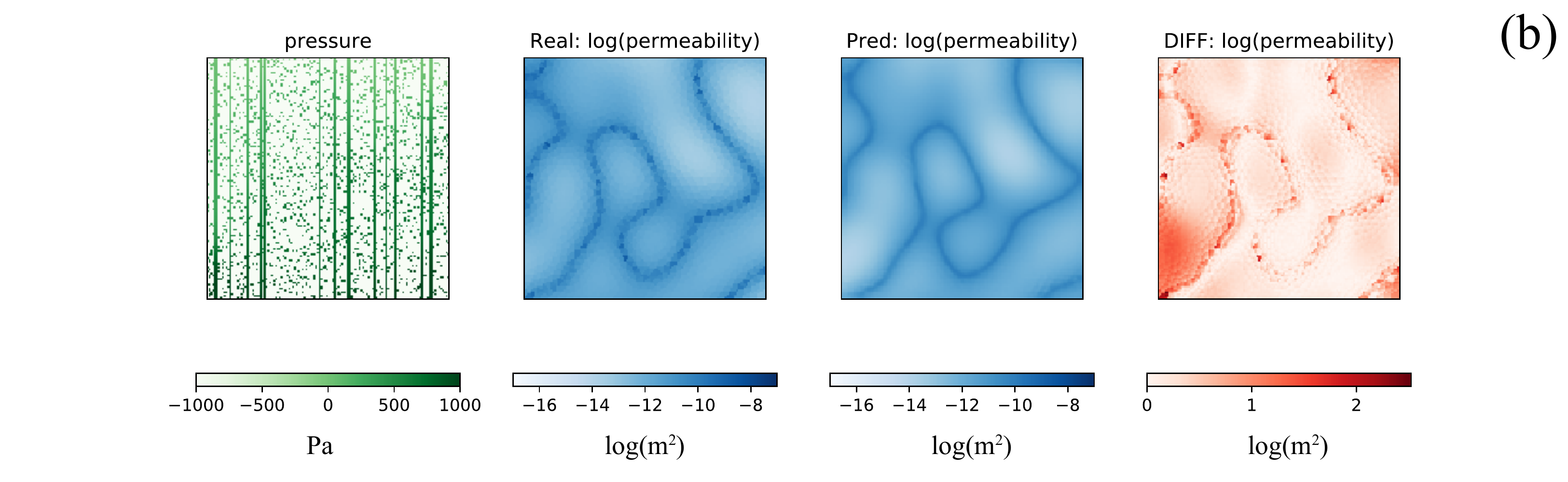}
         \includegraphics[keepaspectratio, height=3.5cm]{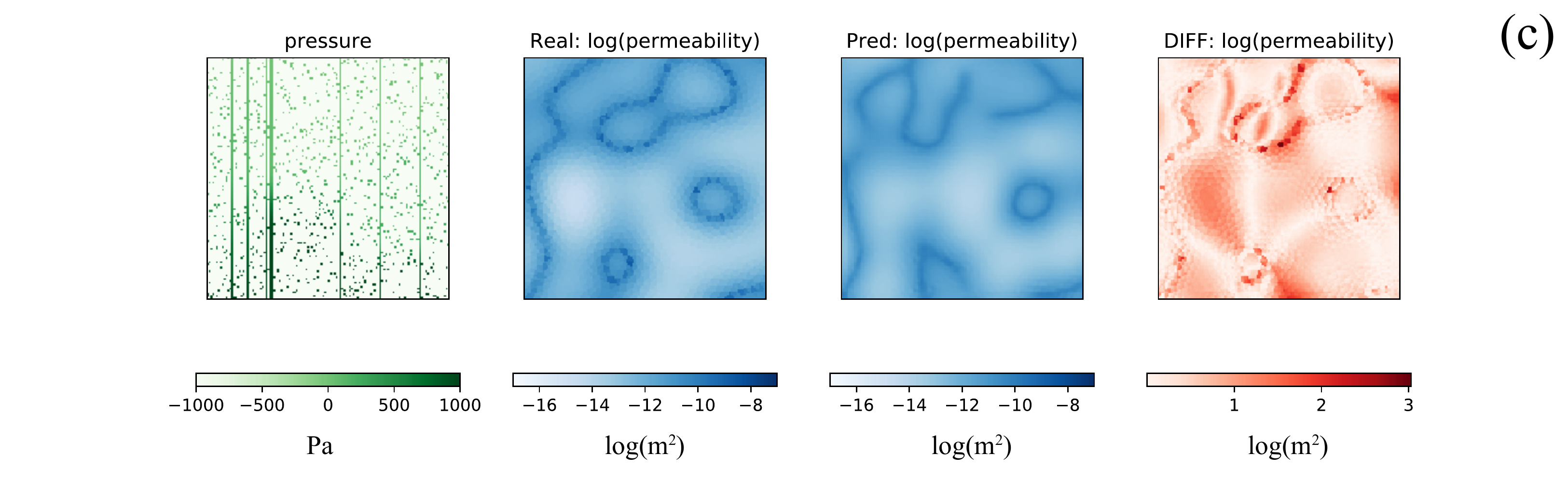}
         \includegraphics[keepaspectratio, height=3.5cm]{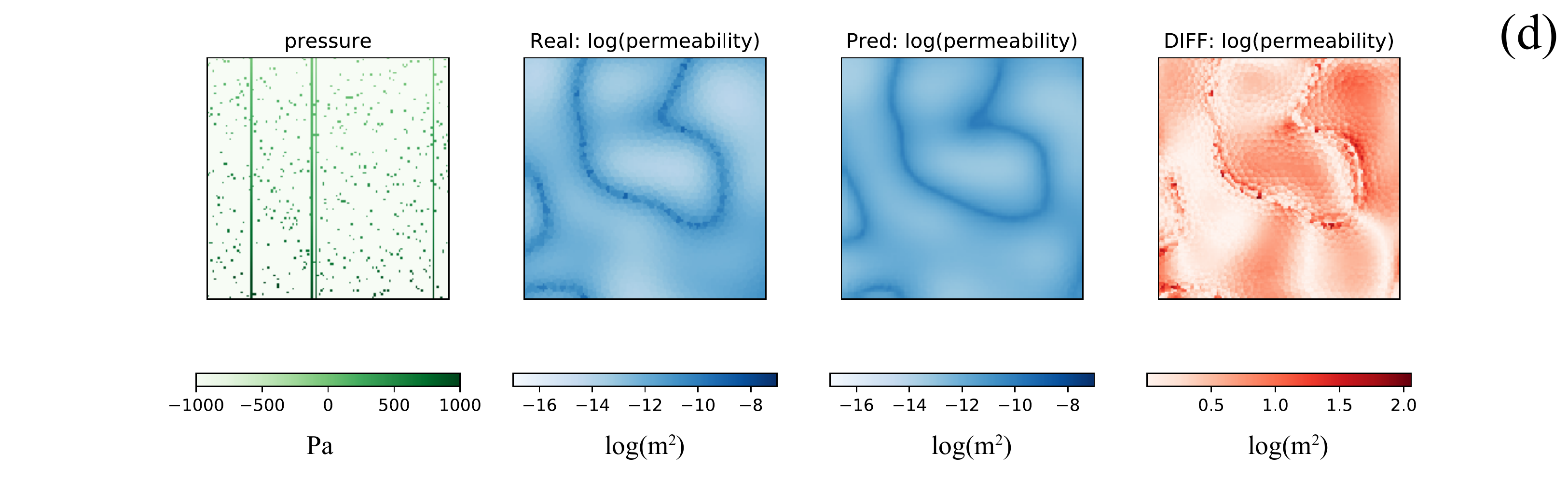}
   \caption{Example 6: test cases' results of the W model of (a) Example 6.1 - using 44\% of input fields,(b) Example 6.2 - using 24\% of input fields, (c) Example 6.3 - using 12\% of input fields, and (d) Example 6.4 - 6\% of input fields. Note that we train our model using 10000 training examples and test them using 1000 test examples. Each case shown here is randomly picked from 1000 test examples.}
   \label{fig:ex7_pic}
\end{figure}

In this example, we study the effects of available data on the model training dynamics as well as test cases' accuracy. We only randomly remove the data points from the input fields, (1) only pressure field or (2) pressure, displacement in the x-direction, and displacement in the y-direction fields. We present five examples of test cases in Fig. \ref{fig:ex7_pic} for (a) Example 6.1 - using 44\% of input fields,(b) Example 6.2 - using 24\% of input fields, (c) Example 6.3 - using 12\% of input fields, and (d) Example 6.4 - 6\% of input fields. As expected, as the number of available data reduces, the model's accuracy deteriorates. However, the DIFF values are still one- to two-ordered magnitude less than the log of the permeability field values. \par

\noindent
\emph{Example 6.1: using 44\% of input fields} \label{sec:inverse_44}

\begin{figure}[!ht]
   \centering
         \includegraphics[keepaspectratio, height=3.5cm]{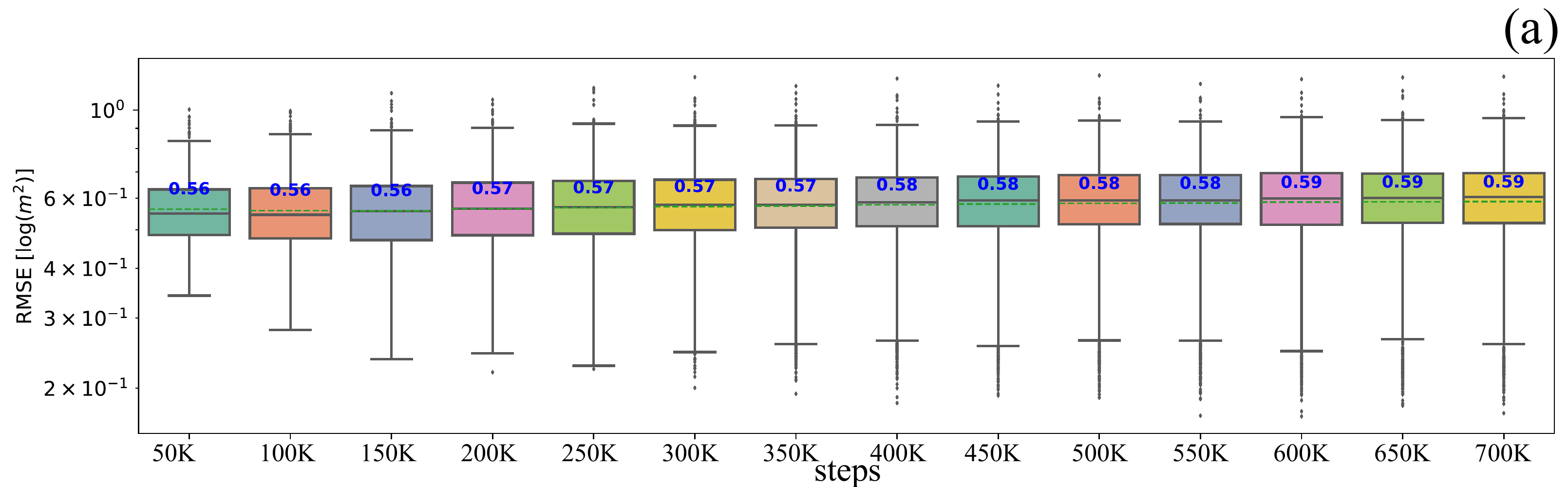}
         \includegraphics[keepaspectratio, height=3.5cm]{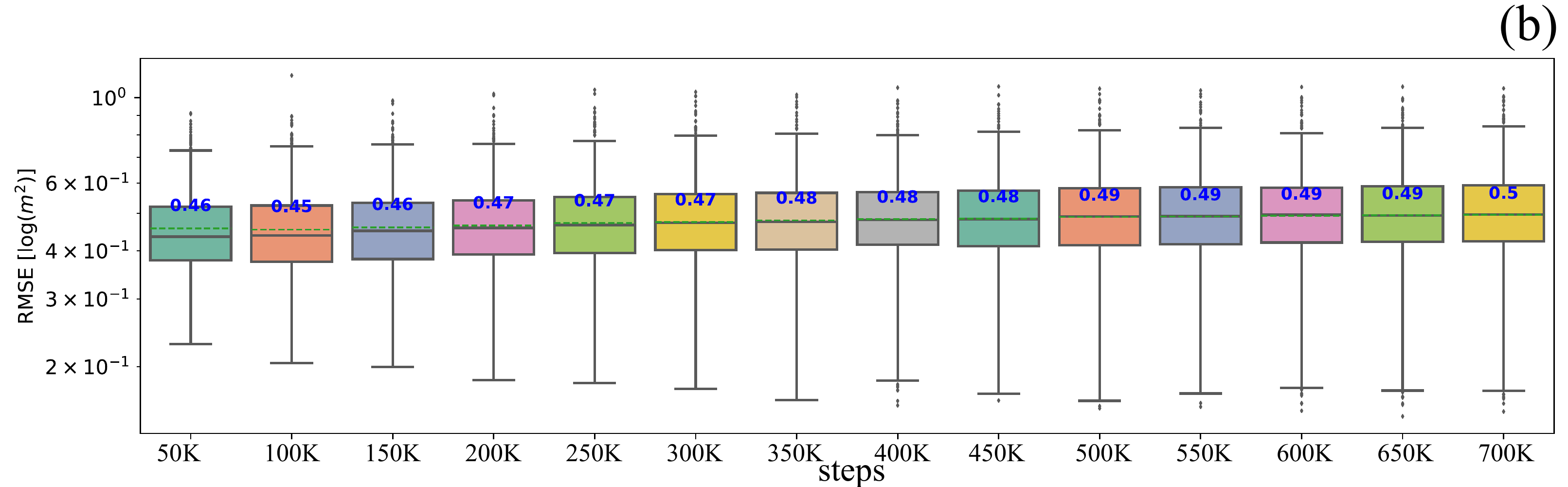}
   \caption{Example 6.1: Root Mean Square Error (RMSE) of W model using (a) pressure as input and (b) pressure and displacement as input. The number of training examples is 10000. Each step refers to each time we perform back-propagation including updating both generator and discriminator's parameters.}
   \label{fig:ex7_1_test}

\end{figure}

Example 6.1 uses the only 44\% of original input fields, and its RSME results are shown in Fig. \ref{fig:ex7_1_test}a using only pressure field as an input - case 1 and Fig. \ref{fig:ex7_1_test}b using pressure and displacement fields as input - case 2. Again, case 2 (more information for the input) has better accuracy than case 1. The maximum of RMSE average values are 0.59 and 0.50 log(\si{m^2}) for case 1 and case 2, respectively, see blue texts in the Fig. \ref{fig:ex7_1_test}. The minimum of RMSE average values are 0.56 and 0.45 log(\si{m^2}) for case 1 and case 2, respectively. As expected, the accuracy deteriorates as we reduce the number of available data (i.e., from 75\% of original input fields to 44\% of original input fields). Similar to Example 5.2, the model exhibits the overfitting behavior. \par

\noindent
\emph{Example 6.2: using 24\% of input fields} \label{sec:inverse_24}

\begin{figure}[!ht]
   \centering
         \includegraphics[keepaspectratio, height=3.5cm]{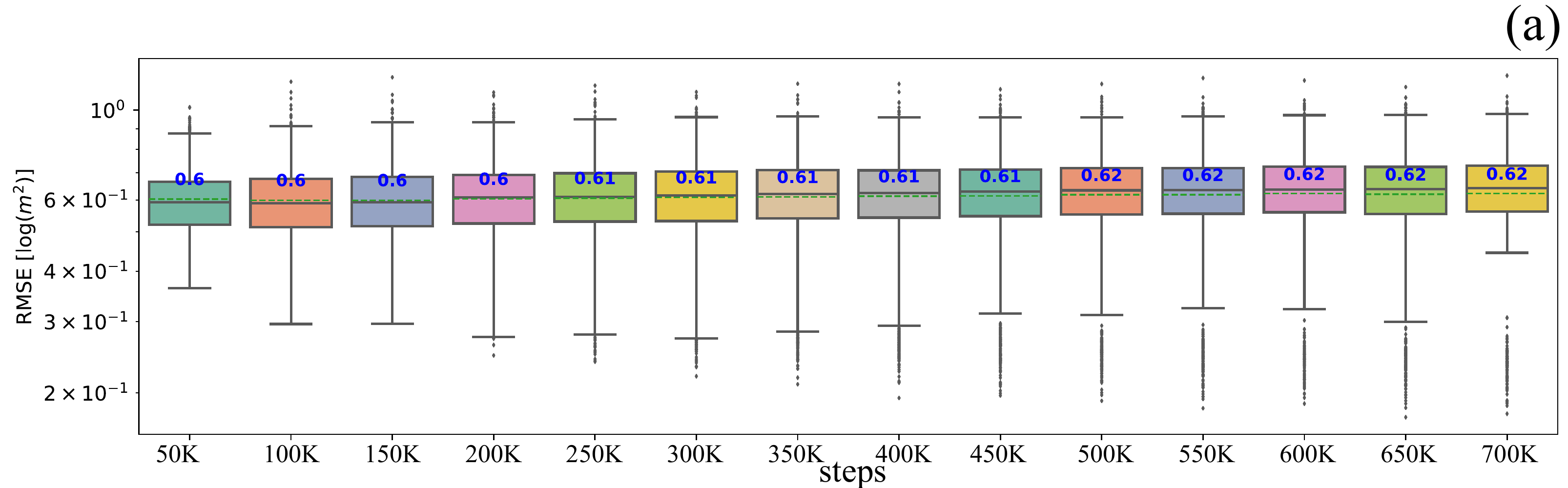}
         \includegraphics[keepaspectratio, height=3.5cm]{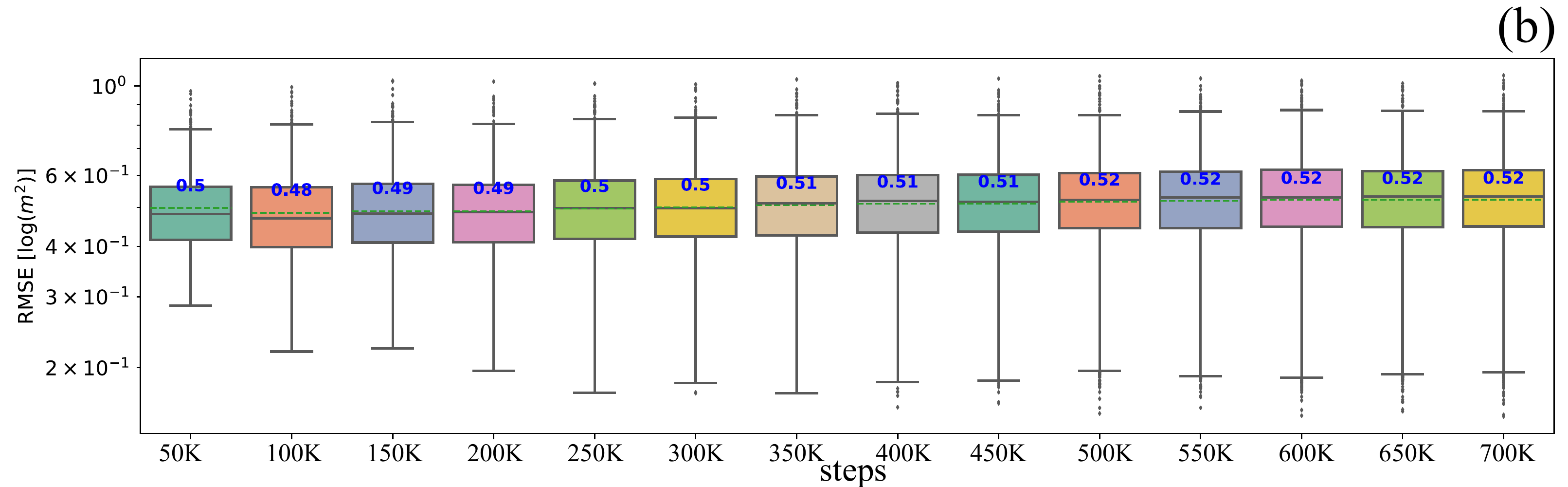}
   \caption{Example 6.2: Root Mean Square Error (RMSE) of W model using (a) pressure as input and (b) pressure and displacement as input. The number of training examples is 10000. Each step refers to each time we perform back-propagation including updating both generator and discriminator's parameters.}
   \label{fig:ex7_2_test}

\end{figure}

We now reduce the available data to 24\% of original input fields. The RMSE values are presented in Fig. \ref{fig:ex7_2_test}a using only pressure field as an input - case 1 and Fig. \ref{fig:ex7_2_test}b using pressure and displacement fields as input - case 2. The maximum of RMSE average values are 0.62 and 0.52 log(\si{m^2}) for case 1 and case 2, respectively. The minimum of RMSE average values are 0.60 and 0.48 log(\si{m^2}) for case 1 and case 2, respectively. Similar to Example 6.1, the model with more input information (case 2) has better accuracy. The ROM framework shows the overfitting behavior. The model's accuracy is decreased as we reduce the available information from 44\% of original input fields to 24\% of original input fields. \par

\noindent
\emph{Example 6.3: using 12\% of input fields} \label{sec:inverse_12}

\begin{figure}[!ht]
   \centering
         \includegraphics[keepaspectratio, height=3.5cm]{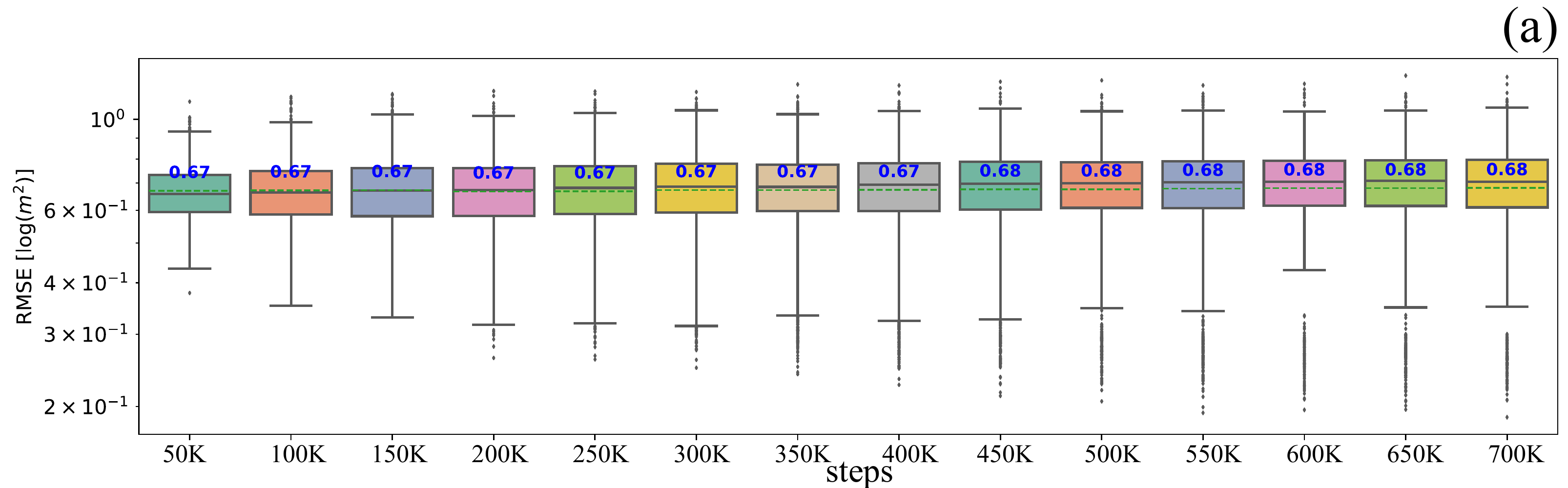}
         \includegraphics[keepaspectratio, height=3.5cm]{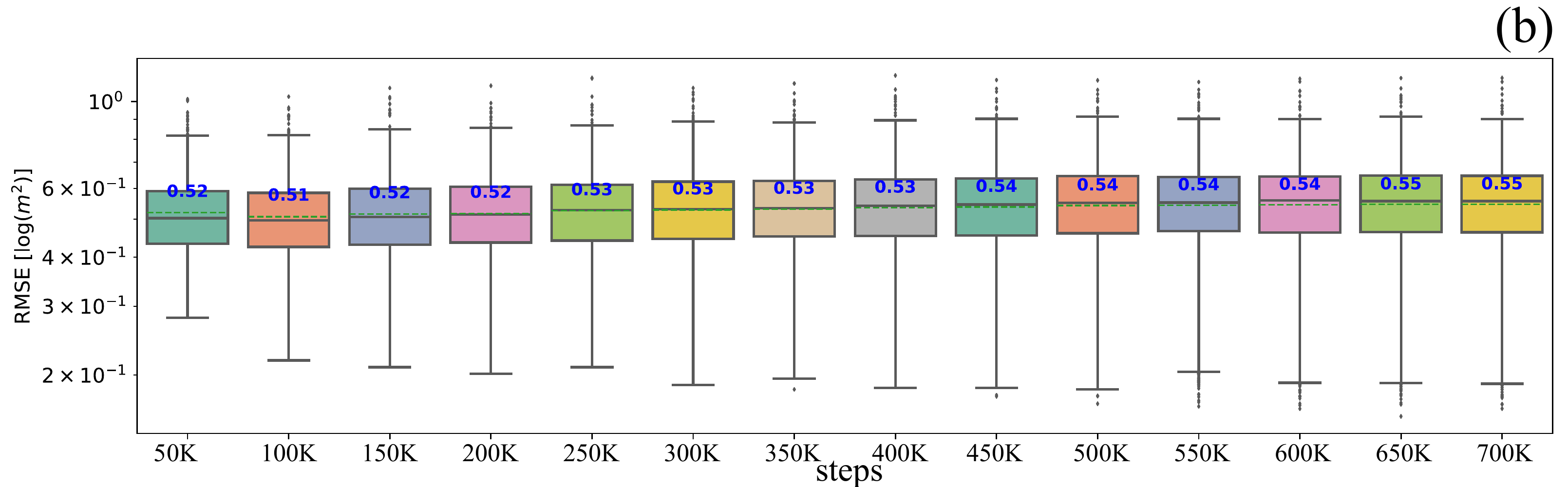}
   \caption{Example 6.3: Root Mean Square Error (RMSE) of W model using (a) pressure as input and (b) pressure and displacement as input. The number of training examples is 10000. Each step refers to each time we perform back-propagation including updating both generator and discriminator's parameters.}
   \label{fig:ex7_3_test}
\end{figure}

We then decrease the available information from 24\% of original input fields to 12\% of original input fields. The RMSE behaviors are illustrated in Fig. \ref{fig:ex7_3_test}a using only pressure field as an input - case 1 and Fig. \ref{fig:ex7_3_test}b using pressure and displacement fields as input - case 2. The maximum of RMSE average values are 0.68 and 0.55 log(\si{m^2}) for case 1 and case 2, respectively. The minimum of RMSE average values are 0.67 and 0.51 log(\si{m^2}) for case 1 and case 2, respectively. Again, case 2 has better test cases' accuracy than those of case 1. The model still shows the overfitting behavior (i.e., the test cases' accuracy reduces as the number of steps increases.). \par

\noindent
\emph{Example 6.4: using 6\% of input fields} \label{sec:inverse_6}

\begin{figure}[!ht]
   \centering
         \includegraphics[keepaspectratio, height=3.5cm]{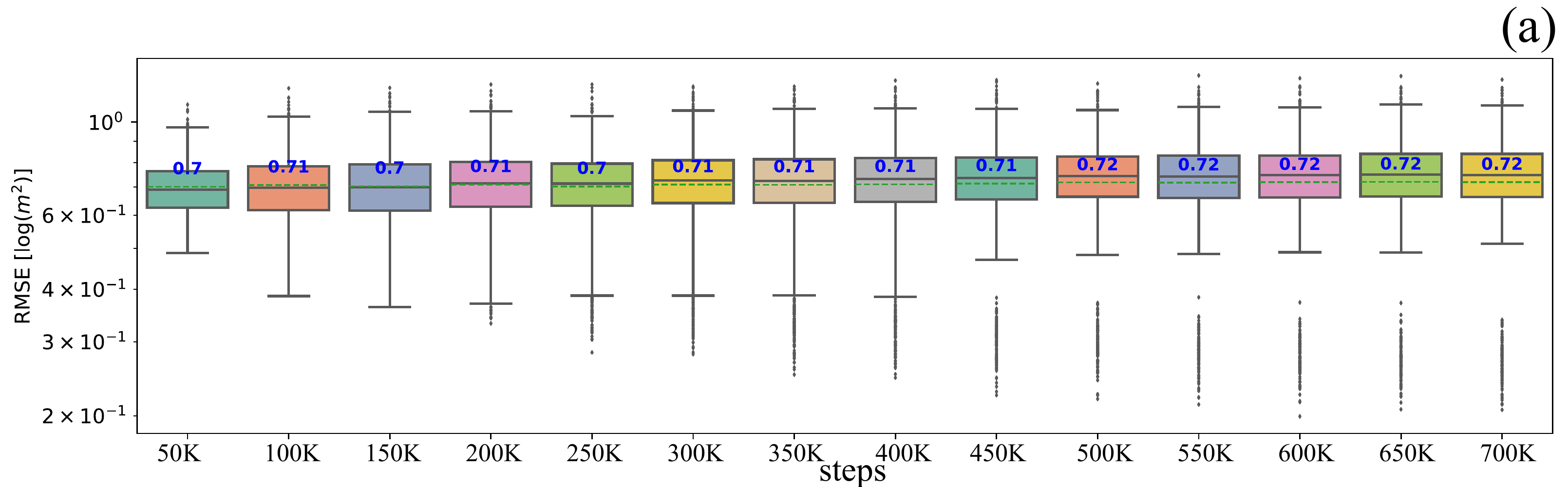}
         \includegraphics[keepaspectratio, height=3.5cm]{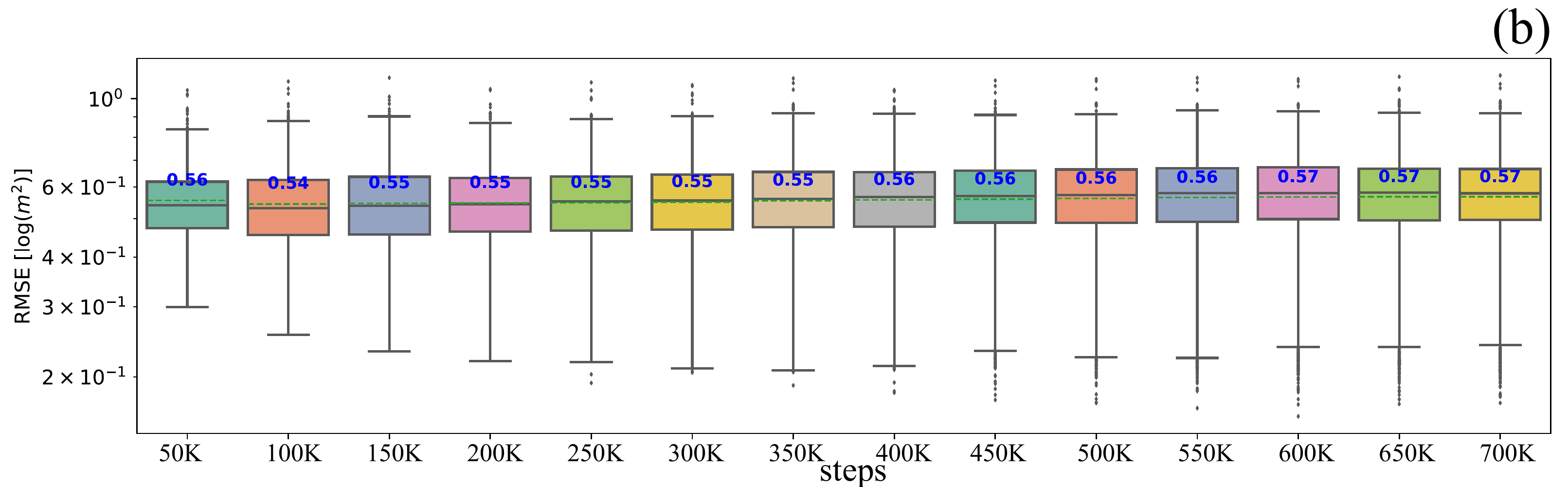}
   \caption{Example 6.4: Root Mean Square Error (RMSE) of W model using (a) pressure as input and (b) pressure and displacement as input. The number of training examples is 10000. Each step refers to each time we perform back-propagation including updating both generator and discriminator's parameters.}
   \label{fig:ex7_4_test}

\end{figure}

Finally, we use only 6\% of original input fields. The RMSE results are shown in Fig. \ref{fig:ex7_4_test}a using only pressure field as an input - case 1 and Fig. \ref{fig:ex7_4_test}b using pressure and displacement fields as input - case 2. The maximum of RMSE average values are 0.72 and 0.57 log(\si{m^2}) for case 1 and case 2, respectively. The minimum of RMSE average values are 0.70 and 0.54 log(\si{m^2}) for case 1 and case 2, respectively. Similar to Examples 5.2, 6.1, 6.2, and 6.3,  case 2 has better test cases' accuracy than those of case 1. The model still shows the overfitting behavior. \par

\subsection{Comparison with Neural Operator approach}\label{sec:com_no}

We compare our W model with the Neural Operator approach \cite{li2020neural}. The methods are tested on the second order elliptic PDE of the form

\begin{equation}
    \begin{aligned}
    - \nabla \cdot a\left(\bm{X}\right) \nabla u\left(\bm{X}\right) = 0 &\text { \: in \: } \Omega, \\
    u(\bm{X}) = 0, &\text { \: on \: } \partial \Omega_{D},
    \end{aligned}
\end{equation}

\noindent
where $\Omega = [0,1] \times [0,1]$. We utilize the training and test data sets as provided by the authors \cite{li2020neural}. The investigated training data consists of 500 training samples. Furthermore, we employ the original Neural Operator code provided by the authors to offer a consistent comparison. \par

We train the W model on a single Quadro RTX 6000, and it takes approximately 15 minutes (resolution of $32 \times 32$) for the model at the 20000 steps checkpoint. The training of the Neural Operator model takes around 3 hours for 500 epochs with a resolution of $35 \times 35$. Consistent with the parameter settings proposed by the authors \cite{li2020neural}, a ReLU activation function and a learning rate of $10^{-4}$ are used. The kernel network consists of 5 layers with a width of 256 each. Both the test and training radii are set to $0.1$. We found that an increase in training resolution (from $35 \times 35$ to $85 \times 85$) significantly increases the training time. This expensive training behavior may limit the utility of the Neural Operator approach. We project that using our available computational infrastructure and employing a training resolution of $241\times 241$ would take more than one week to train. We have not experienced the same issues with the W model. \par

\begin{figure}[!ht]
   \centering
         \includegraphics[keepaspectratio, height=3.5cm]{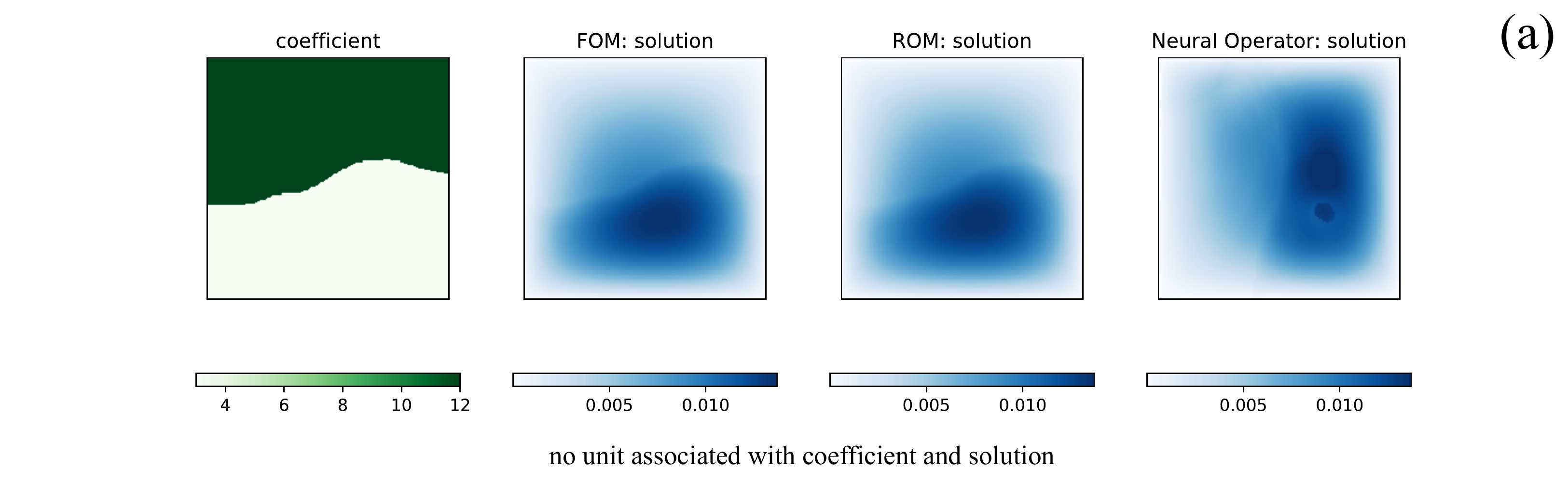}
         \includegraphics[keepaspectratio, height=3.5cm]{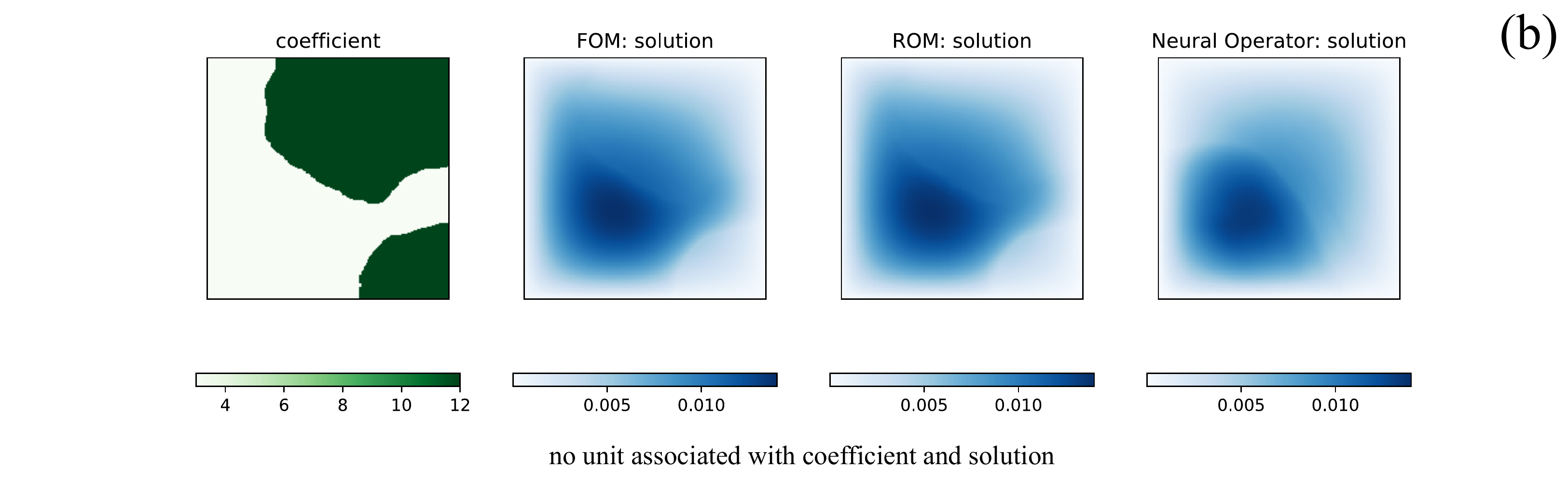}
         \includegraphics[keepaspectratio, height=3.5cm]{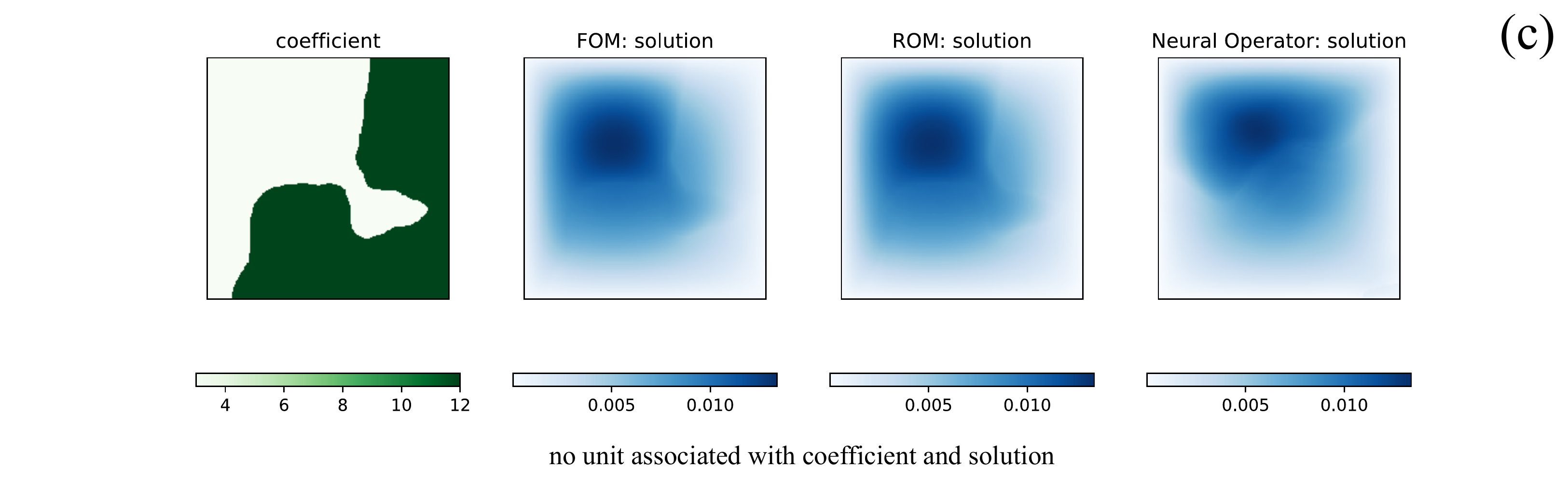}
   \caption{Comparison of test cases' results between the W model and Neural Operator approach. Note that we train the models using 500 training examples and test them using 524 test examples. These three cases shown here are randomly picked from 524 test examples.}
   \label{fig:com_NO_pics}

\end{figure}

The generalization capabilities of the W model and the Neural Operator are compared in Figure \ref{fig:com_NO_pics} for three random test cases, which were not part of the training set. Overall, the RMSE of the W model is $5.69 \times 10^{-10}$, $7.65 \times 10^{-10}$, and $1.50 \times 10^{-9}$ for Figures \ref{fig:com_NO_pics}a, \ref{fig:com_NO_pics}b, and \ref{fig:com_NO_pics}c, respectively. The RMSE of the Neural Operator is $7.68 \times 10^{-5}$, $1.78 \times 10^{-6}$, and $1.89 \times 10^{-6}$ for Figures \ref{fig:com_NO_pics}a, \ref{fig:com_NO_pics}b, and \ref{fig:com_NO_pics}c, respectively. It can be seen that the proposed W model outperforms the Neural Operator approach for the investigated dataset.
\par

\subsection{Comparison with generator that has no skip connections}\label{sec:no_unet}

\begin{figure}[!ht]
   \centering
         \includegraphics[keepaspectratio, height=3.5cm]{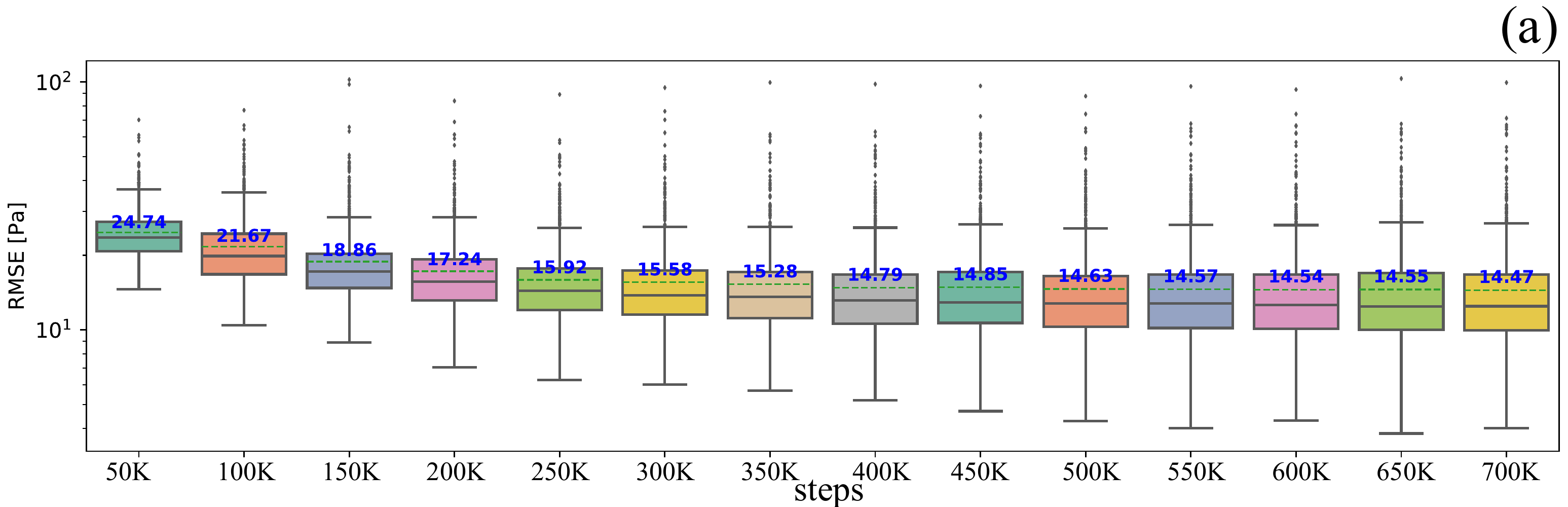}
         \includegraphics[keepaspectratio, height=3.5cm]{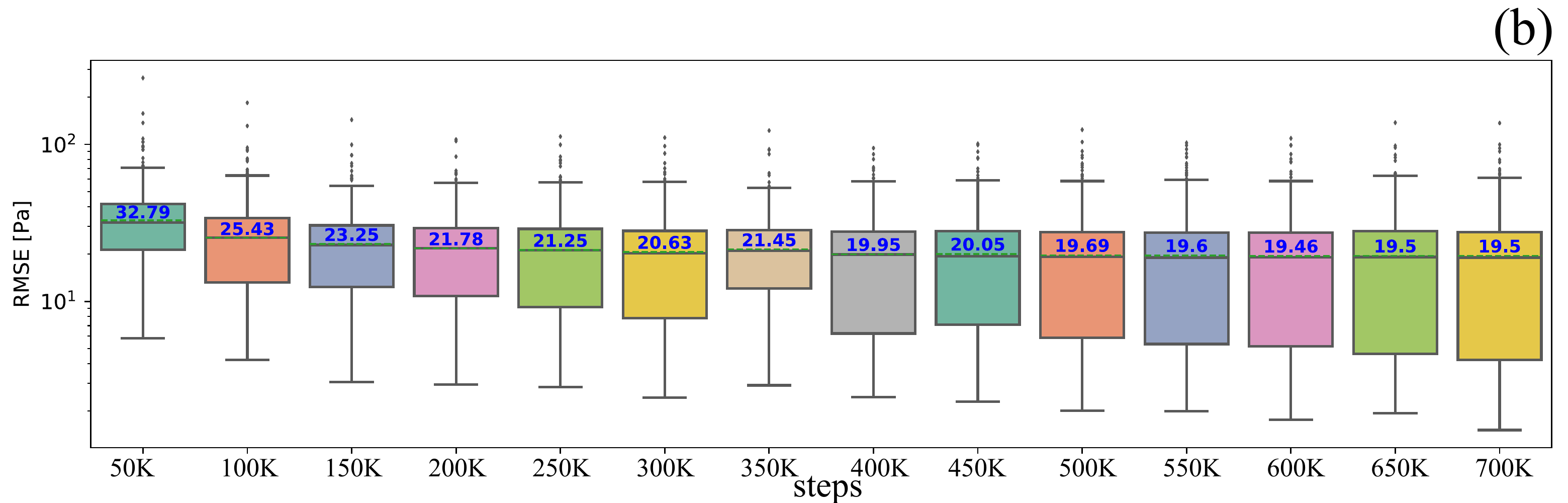}
   \caption{Comparison with generator that has no skip connections: Root Mean Square Error (RMSE) of W model using (a) permeability fields from a Zinn \& Harvey transformation as input and (b)  permeability fields given as a Gaussian distribution, permeability fields defined by a bimodal transformation, and permeability fields from a Zinn \& Harvey transformation as input. Each step refers to each time we perform back-propagation including updating both generator and discriminator's parameters.}
   \label{fig:com_no_unet}
\end{figure}

We compare our W model with two different generators. The first one uses skip connections (U-Net) as described in \ref{sec:cgan}. The second one removes all skip connections and technically is a deep convolutional autoencoder. We illustrate the RMSE results in Figure \ref{fig:com_no_unet} for both using permeability fields from a Zinn \& Harvey transformation as input (Figure \ref{fig:com_no_unet}a) and permeability fields given as a Gaussian distribution, permeability fields defined by a bimodal transformation, and permeability fields from a Zinn \& Harvey transformation as input (Figure \ref{fig:com_no_unet}b). These RMSE results are calculated from a test set. We note that for the first case, we use 10000 training examples and 1000 test examples. The second case uses 7500 training examples (2500 for each field) and 1000 test examples (334 for the Gaussian distribution, 333 for the bimodal transformation, and 333 for the Zinn \& Harvey transformation). \par

Comparing Figures \ref{fig:ex3_test}c and \ref{fig:com_no_unet}a, we observe that the model with a U-Net generator outperforms the model with a deep convolutional autoencoder generator in terms of accuracy. Moreover, the RMSE results of the deep convolutional autoencoder generator become worse as we use permeability fields given as a Gaussian distribution, permeability fields defined by a bimodal transformation, and permeability fields from a Zinn \& Harvey transformation as input (see a comparison between Figures \ref{fig:ex4_test}c and \ref{fig:com_no_unet}b). This behavior reflects that the deep convolutional autoencoder generator could not generalize as well as the U-Net generator when the permeability fields become more complex. \par


